\pdfoutput=1
\PassOptionsToPackage{unicode}{hyperref}
\PassOptionsToPackage{hyphens}{url}
\documentclass[
]{article}
\usepackage{xcolor}
\usepackage{amsmath,amssymb}
\usepackage{iftex}
  \usepackage[T1]{fontenc}
  \usepackage[utf8]{inputenc}
  \usepackage{textcomp} 
\usepackage{lmodern}
\IfFileExists{upquote.sty}{\usepackage{upquote}}{}
\IfFileExists{microtype.sty}{
  \usepackage[]{microtype}
  \UseMicrotypeSet[protrusion]{basicmath} 
}{}
\makeatletter
\@ifundefined{KOMAClassName}{
  \IfFileExists{parskip.sty}{%
    \usepackage{parskip}
  }{
    \setlength{\parindent}{0pt}
    \setlength{\parskip}{6pt plus 2pt minus 1pt}}
}{
  \KOMAoptions{parskip=half}}
\makeatother
\usepackage{longtable,booktabs,array}
\usepackage{calc} 
\usepackage{etoolbox}
\makeatletter
\patchcmd\longtable{\par}{\if@noskipsec\mbox{}\fi\par}{}{}
\makeatother
\IfFileExists{footnotehyper.sty}{\usepackage{footnotehyper}}{\usepackage{footnote}}
\makesavenoteenv{longtable}
\usepackage{graphicx}
\makeatletter
\newsavebox\pandoc@box
\newcommand*\pandocbounded[1]{
  \sbox\pandoc@box{#1}%
  \Gscale@div\@tempa{\textheight}{\dimexpr\ht\pandoc@box+\dp\pandoc@box\relax}%
  \Gscale@div\@tempb{\linewidth}{\wd\pandoc@box}%
  \ifdim\@tempb\p@<\@tempa\p@\let\@tempa\@tempb\fi
  \ifdim\@tempa\p@<\p@\scalebox{\@tempa}{\usebox\pandoc@box}%
  \else\usebox{\pandoc@box}%
  \fi%
}
\def\fps@figure{htbp}
\makeatother
\providecommand{\tightlist}{%
  \setlength{\itemsep}{0pt}\setlength{\parskip}{0pt}}
\usepackage{bookmark}
\IfFileExists{xurl.sty}{\usepackage{xurl}}{} 
\hypersetup{
  hidelinks,
  pdfcreator={LaTeX via pandoc}}

\author{}
\date{}

\usepackage{newunicodechar}
\usepackage{pifont}
\usepackage{textcomp}
\usepackage{etoolbox}
\newcommand\blfootnote[1]{\begingroup\renewcommand\thefootnote{}\footnote{#1}\addtocounter{footnote}{-1}\endgroup}
\newunicodechar{−}{\ensuremath{-}}
\newunicodechar{→}{\ensuremath{\rightarrow}}
\newunicodechar{↔}{\ensuremath{\leftrightarrow}}
\newunicodechar{⟵}{\ensuremath{\longleftarrow}}
\newunicodechar{↑}{\ensuremath{\uparrow}}
\newunicodechar{⇒}{\ensuremath{\Rightarrow}}
\newunicodechar{≈}{\ensuremath{\approx}}
\newunicodechar{≤}{\ensuremath{\leq}}
\newunicodechar{≥}{\ensuremath{\geq}}
\newunicodechar{≠}{\ensuremath{\neq}}
\newunicodechar{∈}{\ensuremath{\in}}
\newunicodechar{∉}{\ensuremath{\notin}}
\newunicodechar{Δ}{\ensuremath{\Delta}}
\newunicodechar{τ}{\ensuremath{\tau}}
\newunicodechar{ρ}{\ensuremath{\rho}}
\newunicodechar{σ}{\ensuremath{\sigma}}
\newunicodechar{α}{\ensuremath{\alpha}}
\newunicodechar{δ}{\ensuremath{\delta}}
\newunicodechar{½}{\ensuremath{\tfrac12}}
\newunicodechar{✓}{\ding{51}}
\newunicodechar{✦}{\ding{72}}
\newunicodechar{†}{\textdagger}
\newunicodechar{⁺}{\textsuperscript{+}}
\newunicodechar{①}{\ding{172}}
\newunicodechar{②}{\ding{173}}
\newunicodechar{③}{\ding{174}}
\newunicodechar{④}{\ding{175}}
\newunicodechar{⑤}{\ding{176}}
\AtBeginEnvironment{longtable}{\scriptsize}
\begin{document}

\begin{center}{\LARGE\bfseries CLQT: A Closed-Loop, Cost-Aware, Strategy-Consistent Benchmark for Diagnostic Evaluation of LLM Portfolio-Management Agents\par}\end{center}\vspace{0.4em}

\begin{center}Bo Qu¹* · Mingguang Chen²*\blfootnote{*~Corresponding authors.\quad ¹~Independent Researcher. boqu.sh2019@gmail.com.\quad ²~DeepGrounding. deepgroundingai@gmail.com.}\end{center}

\vspace{0.9em}\begin{abstract}

Large language model (LLM) agents are increasingly cast as autonomous portfolio managers, yet the dominant evaluation idiom --- a leaderboard of returns over a fixed window --- certifies neither the soundness of an agent's investment process nor the durability of its edge: a single period's return is dominated by the realized market path, and apparent alpha can dissolve once look-ahead bias and realistic trading costs are controlled. We introduce CLQT, a closed-loop benchmark that reframes LLM trading evaluation as diagnosis rather than ranking. CLQT enforces point-in-time data access through a hard TimeGate, models multi-component institutional transaction and financing costs, scores strategy consistency across rounds, and seals every gather--analyze--decide--execute--reflect cycle into a recompute-verifiable audit chain; the same model runs under two process scaffolds --- a constrained multi-agent investment committee or a single autonomous orchestrator --- making harness design an experimental variable. From the resulting audit trail CLQT computes a five-axis capability scorecard --- Coherence, Acuity, Composure, Discipline, Reliability --- with coherence scored partly by a held-out LLM judge to curb self-preference bias. We validate the instrument on a contamination-controlled, year-long multi-model backtest campaign with a 13-configuration ablation grid and a four-week live broker paper-trading track on post-cutoff data. The diagnosis-first read shows that the capability leader is not the Sharpe leader; that agents' allocations systematically fail to follow their own stated analysis --- a stating-versus-doing gap stable across both tracks (+0.30 backtest, +0.23 live); and that module value registers on the capability and behavioral axes when returns alone cannot separate it. Net of realistic costs, agents clear defensive baselines but do not cleanly beat the index. Credible evaluation of LLM investment agents must therefore diagnose the process rather than rank a period's return --- the standard CLQT operationalizes and makes auditable.

\textbf{Keywords:} large language models, autonomous agents, portfolio management, algorithmic trading, transaction costs, backtesting, agent evaluation.

\end{abstract}

\subsection{1. Introduction}\label{introduction}

Generative AI in finance is shifting from analytical tooling toward delegated decision systems --- a transition whose central open question is when, and on what evidence, an AI agent can be trusted with discretion over a portfolio (Kou et al.~2026). Benchmarks for LLM-driven trading have accordingly advanced from static financial QA toward sequential decision-making (Chen et al.~2025; Yu et al.~2025; Fan et al.~2025). Yet two gaps persist. The first is methodological coverage: to our knowledge, no existing benchmark simultaneously enforces \textbf{point-in-time data semantics}, models realistic \textbf{transaction and financing costs} across asset classes, evaluates \textbf{portfolio-level} (not per-asset) decisions, measures \textbf{strategy consistency} across rounds, accumulates \textbf{structured multi-tier memory}, and treats \textbf{tool orchestration} as a benchmarkable capability. An agent that achieves high returns by inadvertently accessing future data, ignoring execution costs, or drifting from its mandate provides no credible signal about LLM capability.

The second gap is interpretive. The dominant evaluation idiom in LLM trading is the \emph{leaderboard}: rank agents by Sharpe ratio or cumulative return over a fixed period. A single period is dominated by the realized market path and by which risk profile happened to suit it; a model can top the table for reasons unrelated to the quality of its decisions. A credible model ranking would also require many models, regimes, and repeated runs, beyond the scope of one study. We therefore argue for \textbf{diagnosis before ranking}: a closed-loop benchmark should first use the decision trail to localize where and why the gather→analyze→decide→execute→reflect process breaks down.

CLQT is built to be that instrument. Its design thesis is that \textbf{a model should be called ``superior'' only when it dominates the relevant capability axes consistently across sub-periods}; until then, the benchmark's product is a reusable map of limitations. We compute that map as a \textbf{five-axis diagnostic capability scorecard} from the \textbf{audit trail} (§6.4). A recompute-verifiable hash chain enables independent checks of the audit trail's integrity (§4.4). Figure 1 summarizes the design.

The contributions of this paper are:

\begin{enumerate}
\def\labelenumi{\arabic{enumi}.}
\tightlist
\item
  A \textbf{diagnosis-before-ranking reframing} of closed-loop trading evaluation: an instrument that maps agent limitations, with an explicit standard (``consistent dominance across axes and sub-periods'') that resists premature ranking from a single outcome metric.
\item
  A \textbf{five-axis capability scorecard (APM-CS)} --- Coherence, Acuity, Composure, Discipline, Reliability --- synthesized from existing audit fields with no new instrumentation. Each axis is a finance instantiation of a general agent-evaluation capability.
\item
  A \textbf{held-out LLM-as-a-judge} for the Coherence axis (non-cohort, reasoning-enabled, cached), guarding against self-preference bias. The headline coherence gap is corroborated by a second, different-family judge and a full judge swap (§7.9), so the finding is not an artifact of one judge.
\item
  A \textbf{recompute-verifiable audit hash chain} that exposes inconsistent modification of the provided store under an explicit threat model (§4.4).
\item
  A \textbf{closed-loop benchmark substrate} --- temporal integrity, multi-component costs, strategy-consistency measurement, three-tier memory, 19 MCP tools, mandate-aware synthesis, and verified live broker execution (§3--§5).
\item
  A \textbf{grounded demonstration}: a chain-verified year-long bi-weekly campaign (5 models × 2 skill modes, five seeds per reliable configuration), a 13-configuration ablation grid, a held-out Coherence judge, and a live broker track. This demonstration shows that the instrument separates outcome from capability, surfaces model-specific limitations, and explicitly bounds what the evidence can claim.
\end{enumerate}

\begin{figure}[tbp]\centering
\includegraphics[width=0.9\linewidth,keepaspectratio]{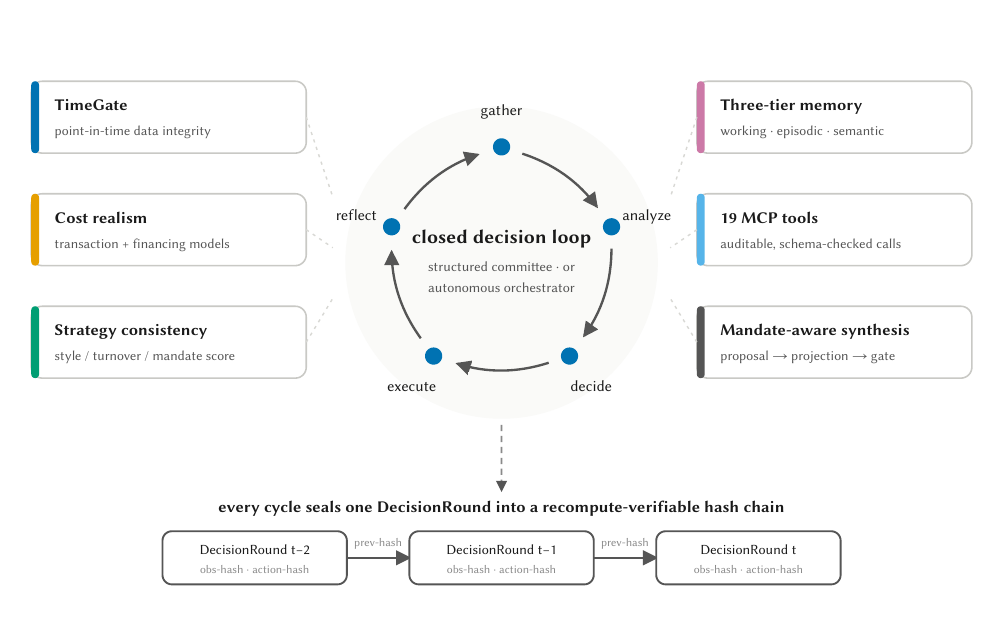}

\vspace{7pt}
\begin{minipage}{0.86\textwidth}\small \textbf{Figure 1.}\textit{ CLQT graphical abstract --- six design pillars surround the five-stage closed decision loop; every cycle seals one \emph{DecisionRound} into a hash-linked audit chain.}\end{minipage}
\end{figure}

\textbf{Roadmap.} §2 locates CLQT in the benchmark landscape and names the agent-infrastructure and cognitive foundations its design borrows. §3--§5 describe the substrate that \emph{produces} the evidence: the closed-loop system (§3), the temporal-integrity and audit machinery that make its numbers trustworthy (§4), and the backtest and live protocols (§5). §6 defines what is measured from that substrate: the five-axis scorecard, its held-out judge protocol, and the behavioral method. §7 then presents the evidence in the order the concepts were built: the year-long backtest campaign (§7.3--§7.6), the ablation grid and passive-baseline read (§7.7--§7.8), and the live validation and behavioral findings (§7.9--§7.10). The research questions and their answers are mapped up front in §7.1. §8 discusses what the findings mean beyond finance, and §8.5 bounds what they may claim. Reproducibility and artifacts are collected in Appendix G. Throughout, we reserve unqualified language for what the campaign directly measures; regime-, cohort-, and judge-conditional findings carry their scope where they are stated, and §8.5 collects every bound in one place.

\subsection{2. Literature Review and Positioning}\label{literature-review-and-positioning}

This section positions CLQT in three contexts: the closed-loop trading benchmarks used for comparison (§2.1--§2.2), the general LLM-agent infrastructure its machinery adapts (§2.3), and the decision-theoretic and cognitive foundations its capability axes operationalize (§2.4).

\subsubsection{2.1 Benchmark Landscape}\label{benchmark-landscape}

We survey twelve representative closed-loop trading benchmarks and map them against CLQT on seven \emph{core} features (Table 1).

\begin{center}\begin{minipage}{0.86\linewidth}\small
\textbf{Table 1.}\textit{ Core-feature coverage across closed-loop LLM trading benchmarks, ordered by year. Features: PiT = point-in-time data enforcement; Cost = transaction-cost modeling; Port = portfolio-level (multi-holding) evaluation; Multi = multi-asset; Mem = cross-round memory; Tools = tool-orchestration evaluation; Live = live-track support. ✓ = supported, --- = absent; a partially-supported feature is credited as ✓ for prior benchmarks. \emph{Covered} is a pure feature-credit tally (of 7), not a quality score. CLQT's further contributions --- strategy-consistency measurement, a verifiable audit trail, and an ablation framework --- are omitted from this shared-feature view. CLQT's \emph{Multi} mark reflects substrate support; the campaign in §7 exercises US equities only.}
\end{minipage}\end{center}

{\def\LTcaptype{none} 
\begin{longtable}[]{@{}
  >{\raggedright\arraybackslash}p{(\linewidth - 18\tabcolsep) * \real{0.4419}}
  >{\raggedright\arraybackslash}p{(\linewidth - 18\tabcolsep) * \real{0.0581}}
  >{\raggedright\arraybackslash}p{(\linewidth - 18\tabcolsep) * \real{0.0465}}
  >{\raggedright\arraybackslash}p{(\linewidth - 18\tabcolsep) * \real{0.0581}}
  >{\raggedright\arraybackslash}p{(\linewidth - 18\tabcolsep) * \real{0.0581}}
  >{\raggedright\arraybackslash}p{(\linewidth - 18\tabcolsep) * \real{0.0698}}
  >{\raggedright\arraybackslash}p{(\linewidth - 18\tabcolsep) * \real{0.0465}}
  >{\raggedright\arraybackslash}p{(\linewidth - 18\tabcolsep) * \real{0.0698}}
  >{\raggedright\arraybackslash}p{(\linewidth - 18\tabcolsep) * \real{0.0581}}
  >{\raggedright\arraybackslash}p{(\linewidth - 18\tabcolsep) * \real{0.0930}}@{}}
\toprule\noalign{}
\begin{minipage}[b]{\linewidth}\raggedright
Benchmark
\end{minipage} & \begin{minipage}[b]{\linewidth}\raggedright
Year
\end{minipage} & \begin{minipage}[b]{\linewidth}\raggedright
PiT
\end{minipage} & \begin{minipage}[b]{\linewidth}\raggedright
Cost
\end{minipage} & \begin{minipage}[b]{\linewidth}\raggedright
Port
\end{minipage} & \begin{minipage}[b]{\linewidth}\raggedright
Multi
\end{minipage} & \begin{minipage}[b]{\linewidth}\raggedright
Mem
\end{minipage} & \begin{minipage}[b]{\linewidth}\raggedright
Tools
\end{minipage} & \begin{minipage}[b]{\linewidth}\raggedright
Live
\end{minipage} & \begin{minipage}[b]{\linewidth}\raggedright
Covered
\end{minipage} \\
\midrule\noalign{}
\endhead
\bottomrule\noalign{}
\endlastfoot
InvestorBench (Li et al.~2024) & 2024 & --- & --- & --- & ✓ & ✓ & --- & --- & 2/\allowbreak{}7 \\
StockBench (Chen et al.~2025) & 2025 & ✓ & --- & --- & --- & --- & --- & --- & 1/\allowbreak{}7 \\
FINSABER (Li et al.~2025b) & 2025 & ✓ & ✓ & --- & --- & --- & --- & --- & 2/\allowbreak{}7 \\
DeepFund (Li et al.~2025a) & 2025 & ✓ & --- & ✓ & --- & --- & --- & ✓ & 3/\allowbreak{}7 \\
AI-Trader (Fan et al.~2025) & 2025 & ✓ & --- & --- & ✓ & --- & ✓ & ✓ & 4/\allowbreak{}7 \\
LiveTradeBench (Yu et al.~2025) & 2025 & ✓ & --- & ✓ & ✓ & --- & --- & ✓ & 4/\allowbreak{}7 \\
When Agents Trade (Qian et al.~2025) & 2025 & ✓ & --- & --- & ✓ & ✓ & --- & ✓ & 4/\allowbreak{}7 \\
PortBench (Zhao et al.~2026) & 2026 & ✓ & --- & ✓ & ✓ & --- & --- & --- & 3/\allowbreak{}7 \\
TraderBench (Yuan et al.~2026) & 2026 & ✓ & ✓ & --- & ✓ & --- & ✓ & --- & 4/\allowbreak{}7 \\
OpenFinGym (Zhang et al.~2026a) & 2026 & ✓ & ✓ & ✓ & ✓ & --- & --- & ✓ & 5/\allowbreak{}7 \\
NextFund (Li et al.~2026) & 2026 & ✓ & ✓ & ✓ & --- & ✓ & --- & ✓ & 5/\allowbreak{}7 \\
OpenPM (Cai et al.~2026) & 2026 & ✓ & ✓ & ✓ & ✓ & --- & ✓ & --- & 5/\allowbreak{}7 \\
\textbf{CLQT (ours)} & 2026 & ✓ & ✓ & ✓ & ✓ & ✓ & ✓ & ✓ & \textbf{7/\allowbreak{}7} \\
\end{longtable}
}

Each prior benchmark goes deep on the facet it was built for --- leakage control (StockBench, DeepFund), live realism (LiveTradeBench, AI-Trader, When Agents Trade), multi-asset pipeline scoring (PortBench, LiveTradeBench), verifiable evaluation infrastructure (OpenFinGym). On any single facet a specialist may go deeper; Table 1 credits that depth in full. The tally is the prerequisite, not the contribution: scoring a decision \emph{process} requires the whole stack in one loop, and only the full set, closed into that loop, supports the capability profile of §6.

\textbf{Recent work (2025--2026), read closely.} Six strands sharpen one facet each.

\emph{Full-pipeline scoring.} PortBench (Zhao et al.~2026) shares CLQT's diagnostic stance --- decomposing \emph{where in the pipeline} error originates --- but scores that pipeline as a stage-by-stage backtest without persistent memory or tool use. CLQT extends the stance to a \emph{closed loop} with memory, MCP tools, realistic costs, and a live track (§3, §5), replacing stage-accuracy with a five-axis capability profile.

\emph{Returns-centric autonomous trading.} StockBench (Chen et al.~2025) and AI-Trader (Fan et al.~2025) report returns-centric outcomes for end-to-end autonomous agents, not a decision-\emph{process} profile. The finding that capable models can trade profitably --- alongside LLM text-processing evidence in exchange-rate market prediction (Yang et al.~2025) --- motivates CLQT's complementary question: \emph{is the profit coherent with the agent's stated reasoning?} (§7 shows it frequently is not). AlphaForgeBench (Zhang et al.~2026b) shows extreme run-to-run variance and action flipping in closed-loop LLM traders even under deterministic decoding --- motivating CLQT's prespecified repeated-run noise floor (§7.7) and consistency-as-capability scoring. Non-LLM learning-based portfolio management advances in parallel (e.g., factor-informed reinforcement-learning portfolio optimization; Kim et al.~2026).

\emph{Live arenas and architecture comparison.} When Agents Trade (Qian et al.~2025) runs four agent frameworks continuously on live stock and crypto markets, finding that agent \emph{architecture}, not the LLM backbone, dominates outcome variation --- echoing the scaffold sensitivity CLQT measures \emph{within} one framework (§7.6, §7.9). NextFund (Li et al.~2026) tracks a fixed multi-agent stack over eight backbones across three equity markets, logging the full decision path; TraderBench (Yuan et al.~2026) probes agents under adversarial market scenarios. All three score outcomes --- signal-level returns without transaction costs (When Agents Trade), pipeline returns (NextFund), realized risk metrics (TraderBench). None verifies its decision log's integrity or scores the process.

\emph{Integrity and leakage.} Profit Mirage (Li et al.~2025c) shows apparent LLM-agent alpha largely collapses once information leakage is removed. Memory-controlled masking (Zhu et al.~2026) reports that returns under leakage-controlled evaluation are substantially explained by passive market and style exposure rather than stock-selection alpha. These are the failure modes CLQT's TimeGate (§4) and per-run integrity ledger exist to prevent. A protocol-coded audit of 77 LLM-trading studies finds that only one of its nineteen primary closed-loop studies reports an explicit transaction-cost model (Xia et al.~2026). This contrasts with cost-aware backtesting practice outside the LLM-agent line (e.g., Grądzki et al.~2025). A coded matrix over thirty trading studies finds evaluation assumptions (timing, turnover, costs) reported far less clearly than architectures (Yao and Zheng 2026). A position paper argues that reported alpha should not count as deployment evidence before temporal-integrity and friction tests (Ye et al.~2026).

\emph{Memory as a measured variable.} The same memory-controlled line (Zhu et al.~2026) (``from knowing to doing'') isolates whether recalled experience improves the next action. CLQT promotes this variable to a first-class axis through the three-tier memory (§3.6) and its ablation (§7.7).

\emph{Verifiable environments and task suites.} OpenFinGym (Zhang et al.~2026a) packages forecasting, simulation, trading, and fraud tasks behind a containerized verifier that withholds outcomes from the agent runtime; its paper-trading engine offers configurable slippage and per-trade costs. FinWorld (Zhang et al.~2025) is the platform sibling, spanning data through deployment. Both evaluate single-pass submissions rather than the memory- and tool-bearing closed-loop agents CLQT instruments. Episodic task suites (e.g., Herculean's four MCP-skill workflows spanning trading, hedging, and auditing (Peng et al.~2026)) catalogue skills without a portfolio carried across rounds. Closest to CLQT's own framing, OpenPM (Cai et al.~2026) couples point-in-time enforcement with typed constraints compiled from natural-language risk mandates and per-run audit artifacts --- a contamination certificate, a cost-sensitivity curve, a constraint-adherence report --- over a long-only S\&P 500 book. It evaluates a single frozen window without market impact, so neither cross-round strategy consistency nor carried memory is scored, and there is no live execution track.

\emph{Synthesis.} \textbf{None of the surveyed benchmarks combines these facets.} Even the closest --- OpenFinGym, NextFund and OpenPM, at five of seven features each --- evaluate single-pass submissions, a single fixed pipeline, or a single frozen window. The audit artifacts one of them emits certify the run's inputs rather than making every recorded decision recomputable, and none scores the decision process itself. CLQT fills the remaining position: a closed-loop, cost-aware, point-in-time, memory- and tool-bearing system scoring the decision \emph{process} as a five-axis capability profile across both backtest and live (Table 1).

\subsubsection{2.2 Key Gaps Addressed}\label{key-gaps-addressed}

Read against Table 1, CLQT closes six gaps: four sharpen features prior work supports only partially, and two are absent from prior benchmarks entirely.

\textbf{Costs beyond a flat commission.} Most prior benchmarks ignore transaction costs (Chen et al.~2025; Yu et al.~2025; Fan et al.~2025; Li et al.~2025a; Qian et al.~2025) or apply a single flat commission (Li et al.~2025b). Recent environments add configurable slippage and per-trade fees (Yuan et al.~2026; Zhang et al.~2026a), a cost-sensitivity sweep (Cai et al.~2026), or order-level microstructure simulation (Papadakis et al.~2025). Cost neglect is independently documented as a pervasive evaluation failure in this literature (Xia et al.~2026; Nguyen and Pham 2026). CLQT implements four parameterized cost models (§3.4) across asset classes and four cost tiers, plus daily financing-cost accrual, with the default tier's commission and spread legs validated against live broker fills (§5.2).

\textbf{Measured cross-round learning.} Memory-bearing agents appear across prior work --- InvestorBench (Li et al.~2024) mentions memory without formalizing or measuring it, and the live arenas field memory-based agents (Qian et al.~2025; Li et al.~2026) without isolating memory's effect. CLQT both formalizes the mechanism (three-tier, decaying, consolidating; §3.6) and measures its value directly by ablation against a repeated-run noise floor (§7.7).

\textbf{Auditable tool orchestration.} AI-Trader (Fan et al.~2025) mandates tool-only execution but does not record orchestration; CLQT logs every MCP tool call with hashes and latency, making orchestration failures measurable (§7.5).

\textbf{Intertemporal information accumulation.} Prior benchmarks treat news as single-round context; CLQT accumulates news across cycles in a persistent store (§3.8) and surfaces sentiment \emph{dynamics} to the agent.

\textbf{Multi-dimensional candidate scouting.} Where prior work screens candidates by price momentum, if at all, CLQT's 3-D SCOUT stage (§3.3) adds earnings surprise and macro-correlation.

\textbf{Live execution as instrument validation.} Benchmarks with live tracks report live outcomes; CLQT pairs live broker execution (§5.2) with the same decision-level audit trail as the backtest, so the live track validates the \emph{instrument}, not only the agents.

\subsubsection{2.3 The LLM-Agent Infrastructure CLQT Builds On}\label{the-llm-agent-infrastructure-clqt-builds-on}

Almost none of CLQT's machinery originates in financial AI: it adapts the general agent-infrastructure literature of 2022--2025, and its contribution is making each borrowed mechanism \emph{measurable} in a cost-bearing, temporally-gated decision process.

The five-stage loop follows ReAct (Yao et al.~2023) and Reflexion's self-critique loop (Shinn et al.~2023), within the unifying frame of CoALA (Sumers et al.~2024; cf.~the ACT* grounding of §2.4). The three-tier memory adapts Generative Agents' memory-stream design (Park et al.~2023), MemGPT's paging (Packer et al.~2023), and CoALA's memory tiers to a portfolio ledger; the novelty is treating \emph{``does recalled experience improve the next decision?''} as a measured axis (§7.7) --- the general question of memory-controlled benchmarks (Zhu et al.~2026). Tool-mediated action follows ToolLLM (Qin et al.~2024), operationalized as a Model-Context-Protocol (Anthropic 2024) layer of 19 quantitative tools, every call hashed, timed, and auditable. The role-decomposed committee and structured skill mode draw on MetaGPT (Hong et al.~2024) and, within finance, on TradingAgents (Xiao et al.~2024) and FinMem (Yu et al.~2023). MetaGPT's encoded standard operating procedures (SOPs) reduce cascading error, motivating the SOP-vs-autonomy trade-off as an experimental variable (§7.6).

Reporting a \emph{profile} rather than a scalar follows HELM (Liang et al.~2022) and trajectory-scoring benchmarks such as AgentBench (Liu et al.~2024) and τ-bench (Yao et al.~2024). Cautions from the LLM-as-a-judge literature about self-preference bias (Zheng et al.~2023) shape the Coherence judge: it is held out of the evaluated cohort and cached. Its scores are reported alongside the mechanical signal--action agreement from which they systematically diverge (§7.9), in the spirit of procedure-aware evaluation (Cao et al.~2026). Judge-scored finance evaluations are emerging (LATTICE scores crypto decision-support agents on six dimensions with LLM judges (Chan et al.~2026)); CLQT adds a blinded human anchoring of the judge itself (Appendix F.5).

\subsubsection{2.4 Theoretical and Cognitive Foundations}\label{theoretical-and-cognitive-foundations}

The decision loop and three of the five capability axes instantiate established models of decision-making or cognition (the remaining two --- Discipline and Reliability --- are operational constructs; §8.2 maps all five to their general-agent readings)

\emph{The closed loop ⟵ decision-cycle theory.} The gather→analyze→decide→execute→reflect loop instantiates Boyd's OODA (Observe--Orient--Decide--Act) cycle (Boyd 1987) and Endsley's three-level situation-awareness model (Endsley 1995): Intelligence Acquisition is \emph{perception}, Signal Synthesis is \emph{comprehension}, Portfolio Construction is \emph{projection}-and-decision. This grounding also \emph{defines} Acuity (§6.4): a Level-1 (perception) failure --- attending to salient-but-irrelevant cues --- is what the axis measures by contrasting responses to informative signals against a volatility \emph{distractor}.

\emph{Reasoning vs.~acting ⟵ Anderson's declarative/procedural distinction.} Anderson's ACT* theory (Anderson 1983) separates \emph{declarative} knowledge from \emph{procedural} condition--action rules compiled only through repeated practice. This predicts \textbf{a specific deficit in text-trained models: they encode the \emph{propositions} of a strategy but not the \emph{production rules} that let those propositions govern an action}. Compilation requires practice a text corpus cannot supply. This is the theoretical basis for the Coherence axis and, we argue in §8, the most parsimonious explanation of the headline stating-versus-doing gap.

\emph{Reflect-and-learn ⟵ experiential-learning theory.} The Attribution \& Lesson Extraction stage and three-tier memory operationalize Kolb's experiential-learning cycle (Kolb 1984). In Anderson's terms, reflection is an attempt to \emph{supply the missing practice}; the memory ablation (§7.7) answers whether it helps.

\emph{Composure ⟵ behavioral finance.} Composure measures overreaction to volatility --- the LLM-agent analogue of the classic overreaction hypothesis (De Bondt and Thaler 1985) and of noise-driven trading: a disciplined agent should not churn its book on transient volatility. These groundings are what make the axes portable beyond finance (§8.2).

\subsection{3. System Design}\label{system-design}

CLQT's design goal is that every mechanism an agent runs on is also one the evaluation can \emph{measure}. We walk the substrate in execution order --- loop and agents (§3.1--§3.2), information gathering and scouting (§3.3), cost-aware execution (§3.4), consistency measurement (§3.5), memory (§3.6), decision pipeline (§3.7), sentiment trajectory (§3.8) --- and note the audit fields each component emits for the §6.4 scorecard.

\subsubsection{3.1 Architecture Overview}\label{architecture-overview}

CLQT operates as a \textbf{five-stage closed loop} executed by \textbf{TradingEngine}, which orchestrates \textbf{Protocol modules} through a \textbf{six-agent architecture}. This makes the OODA / situation-awareness cycle of §2.4 executable: perception (IA), comprehension (MSS), projection-and-decision (MAPC), action (COE), reflection (ALE). The role decomposition follows the multi-agent financial-trading line (Xiao et al.~2024); reflect-and-learn instantiates ReAct (Yao et al.~2023) / Reflexion (Shinn et al.~2023). Each stage is a Python \textbf{Protocol interface} --- implementations swap without inheritance. All quantitative analysis runs through \textbf{19 typed tools} in the \textbf{Model-Context-Protocol (MCP)} standard, grouped into four namespaces (market-data, quant-tools, news-sentiment, execution). The tools execute \emph{in-process} against the underlying quant and data libraries, using only the tool-definition interface and schemas, with no networked transport. Every analytical step is an \textbf{auditable, schema-checked tool call}, not free-form text (inventory: Appendix B, Table B1). The five stages:

\begin{enumerate}
\def\labelenumi{\arabic{enumi}.}
\tightlist
\item
  \textbf{Intelligence Acquisition (IA)} --- gather market, news, macro, and asset-class data; scout universe candidates.
\item
  \textbf{Multi-Source Signal Synthesis (MSS)} --- signals, risk assessments, factor decompositions.
\item
  \textbf{Mandate-Aware Portfolio Construction (MAPC)} --- propose and validate target allocations against the mandate.
\item
  \textbf{Cost-Optimal Order Execution (COE)} --- cost-minimizing order generation and execution.
\item
  \textbf{Attribution \& Lesson Extraction (ALE)} --- attribute performance, extract structured lessons, update memory.
\end{enumerate}

The atomic unit of evaluation is the \textbf{DecisionRound}:

{\scriptsize\begin{verbatim}
DecisionRound:
  round_id, as_of_ts, run_id, model_id
  |- IA:      market_snapshot, news_items, sentiment_trajectory, macro_indicators,
  |           asset-class-data, candidate_signals
  |- MSS:     signals, risk_report_pre, factor_attribution
  |- MAPC:    portfolio_before, target_allocation, mapc_delta_proposed/projected,
  |           mapc_adjustments_clipped, universe_changes, target_adherence
  |- COE:     orders, execution_report, portfolio_after, financing_cost
  |- ALE:     risk_report_post, reflection_notes, lessons_generated
  |- Memory:  consistency_score, realized_pnl_this_round, memory_provenance
  |- Cost/IO: agent_costs[] (in/out/cache/reasoning tokens, elapsed_s, cost_usd)
  |- Reliab:  decision_attempts, parse_failures
  `- Audit:   observation_hash, action_hash, prev_round_hash, tool_calls_log, signed
\end{verbatim}}

\emph{sentiment\_trajectory}, \emph{candidate\_signals}, the \emph{mapc\_delta\_*} fields, \emph{agent\_costs{[}{]}}, \emph{parse\_failures}, and \emph{prev\_round\_hash} feed the \textbf{diagnostic scorecard} (§6.4) and \textbf{hash chain} (§4.4); \textbf{every metric in this paper is recoverable from this record} --- no added instrumentation.

\begin{figure}[tbp]\centering
\includegraphics[width=0.9\linewidth,keepaspectratio]{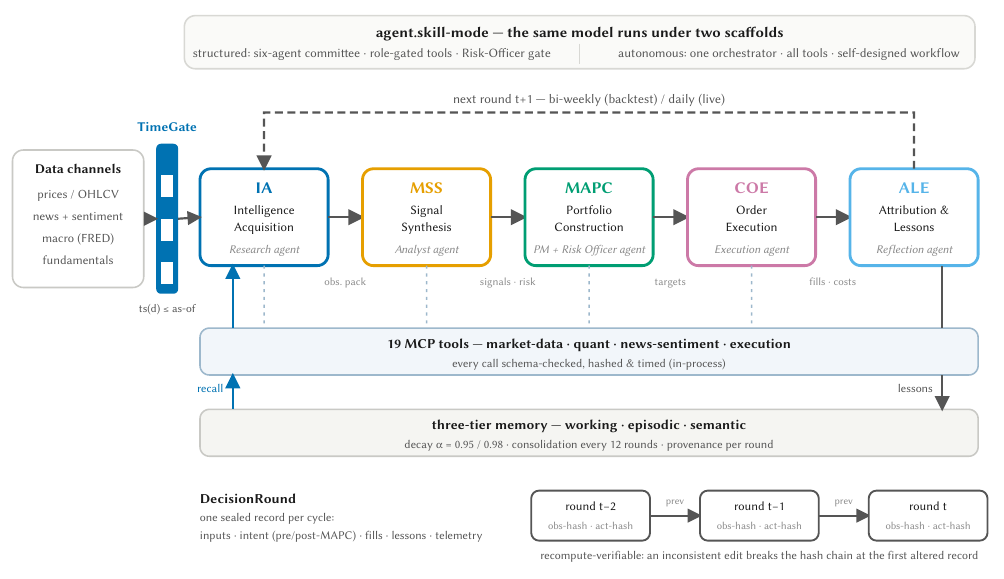}

\vspace{7pt}
\begin{minipage}{0.86\textwidth}\small \textbf{Figure 2.}\textit{ The closed decision loop: five stages (IA → MSS → MAPC → COE → ALE) run by six specialized agents (structured mode; a strict investment-committee process) or one full-access agent (autonomous mode); each cycle reads/writes three-tier memory and emits one sealed, hash-linked \emph{DecisionRound}.}\end{minipage}
\end{figure}

\subsubsection{3.2 Six-Agent Architecture}\label{six-agent-architecture}

Six specialized agents receive \textbf{agent skills}: structured workflow protocols prescribing MCP tool-call sequence and reasoning steps. Each (Research, Analyst, PM, Risk Officer, Execution, Reflection) maps onto a loop stage with only the tool subset its role needs (\textbf{stage-to-agent-to-tool mapping}, Figure 2); exact skill files and tool assignments are in the artifact set (Appendix G).

\paragraph{Structured vs.~Autonomous Skill Modes}\label{structured-vs.-autonomous-skill-modes}

CLQT's \textbf{process-scaffolding variable} --- the agent's \emph{harness}, in agent-evaluation terms --- runs the same model under two skill modes, set by \emph{agent.skill-mode}:

\begin{itemize}
\tightlist
\item
  \textbf{structured}: six agents, role-restricted tool subsets, prescribed skill protocols; agents reason \emph{within} the workflow (views, signal-weight adjustments) but cannot change its topology, and a Risk Officer gate may reject and request revision.
\item
  \textbf{autonomous}: one \textbf{AutonomousOrchestratorAgent} owns SCOUT through ALE with all tools and only a mandate document, constraint set, and required output schema --- choosing which tools to call, in what order, how often, and whether to skip stages.
\end{itemize}

Both modes use \textbf{real agentic tool calling}: tool-use blocks execute, and their results are fed back over multiple turns --- \textbf{not simulated tool use or pre-assembled prompt data}. \emph{(The mode switch confounds Discipline and Acuity; §6.4/§7 compare within mode.)}

\subsubsection{3.3 Multi-Source Information Gathering and 3-Dimensional SCOUT Stage}\label{multi-source-information-gathering-and-3-dimensional-scout-stage}

\paragraph{Standard Data Channels}\label{standard-data-channels}

The \textbf{IA stage} gathers from complementary sources, each TimeGate-enforced (§4.1) and mapped to backtest and live providers so both tracks see equivalent inputs (Table 2):

{\def\LTcaptype{none} 
\begin{longtable}[]{@{}
  >{\raggedright\arraybackslash}p{(\linewidth - 6\tabcolsep) * \real{0.1724}}
  >{\raggedright\arraybackslash}p{(\linewidth - 6\tabcolsep) * \real{0.3563}}
  >{\raggedright\arraybackslash}p{(\linewidth - 6\tabcolsep) * \real{0.2644}}
  >{\raggedright\arraybackslash}p{(\linewidth - 6\tabcolsep) * \real{0.2069}}@{}}
\toprule\noalign{}
\begin{minipage}[b]{\linewidth}\raggedright
Data Channel
\end{minipage} & \begin{minipage}[b]{\linewidth}\raggedright
Fields
\end{minipage} & \begin{minipage}[b]{\linewidth}\raggedright
Backtest source
\end{minipage} & \begin{minipage}[b]{\linewidth}\raggedright
Live source
\end{minipage} \\
\midrule\noalign{}
\endhead
\bottomrule\noalign{}
\endlastfoot
Price/\allowbreak{}Volume & OHLCV + adjusted close (252-day lookback) & Yahoo Finance & Yahoo Finance /\allowbreak{} broker snapshot \\
Fundamentals & P/\allowbreak{}E,\allowbreak{} EPS,\allowbreak{} revenue,\allowbreak{} earnings surprise,\allowbreak{} analyst targets & Finnhub & Finnhub /\allowbreak{} SEC EDGAR \\
News/\allowbreak{}Sentiment & Headlines,\allowbreak{} sentiment score & Alpha Vantage /\allowbreak{} Polygon /\allowbreak{} RSS /\allowbreak{} NewsAPI & same \\
Macro & 12 series (GDP,\allowbreak{} CPI,\allowbreak{} unemployment,\allowbreak{} fed funds,\allowbreak{} VIX,\allowbreak{} \ldots) & FRED & FRED \\
Benchmark & SPY,\allowbreak{} IEF buy-and-hold NAV & Yahoo Finance & Yahoo Finance \\
\end{longtable}
}

\begin{center}\begin{minipage}{0.86\linewidth}\small
\textbf{Table 2.}\textit{ Data sources and channel registry (backtest and live providers per channel).}
\end{minipage}\end{center}

\paragraph{3-Dimensional SCOUT Stage}\label{dimensional-scout-stage}

A \textbf{three-dimensional (3-D) SCOUT stage} between IA and MSS computes the following as of \emph{as\_of\_ts} for each candidate-pool symbol not currently held:

\begin{itemize}
\tightlist
\item
  \textbf{Dimension 1 --- Technical Momentum} (\(m_i\)): 12-month minus 1-month price return.
\item
  \textbf{Dimension 2 --- Core Fundamentals} (\(f_i\)): most recent earnings-surprise percentage (Finnhub/EDGAR), analyst-consensus deviation as fallback.
\item
  \textbf{Dimension 3 --- Macro Correlation} (\(\rho_i^{\text{mkt}}\)): 60-day rolling Pearson correlation of daily log returns with SPY; companion \(\rho_i^{\text{rate}}\) (vs.~TLT) recorded for rate sensitivity, not in the composite.
\end{itemize}

All three are z-scored cross-sectionally and combined:

\[z_i^{\text{composite}} = 0.50 \cdot z_i^{\text{mom}} + 0.30 \cdot z_i^{\text{eps}} + 0.20 \cdot z_i^{\text{mkt\_corr}}\]

with a degenerate-window guard (Appendix B). The top-\(K\) candidates (default \(K=5\)) go to MAPC and are recorded in \emph{candidate\_signals} for the information-foraging analysis in §6.5 (Appendix D). The 50/30/20 split is an ablation default, not a derived optimum (\emph{ablation\_scout\_weights} reserved). Figure 3 shows a real campaign example (JNJ vs.~FCX).

\begin{figure}[tbp]\centering
\includegraphics[width=0.9\linewidth,keepaspectratio]{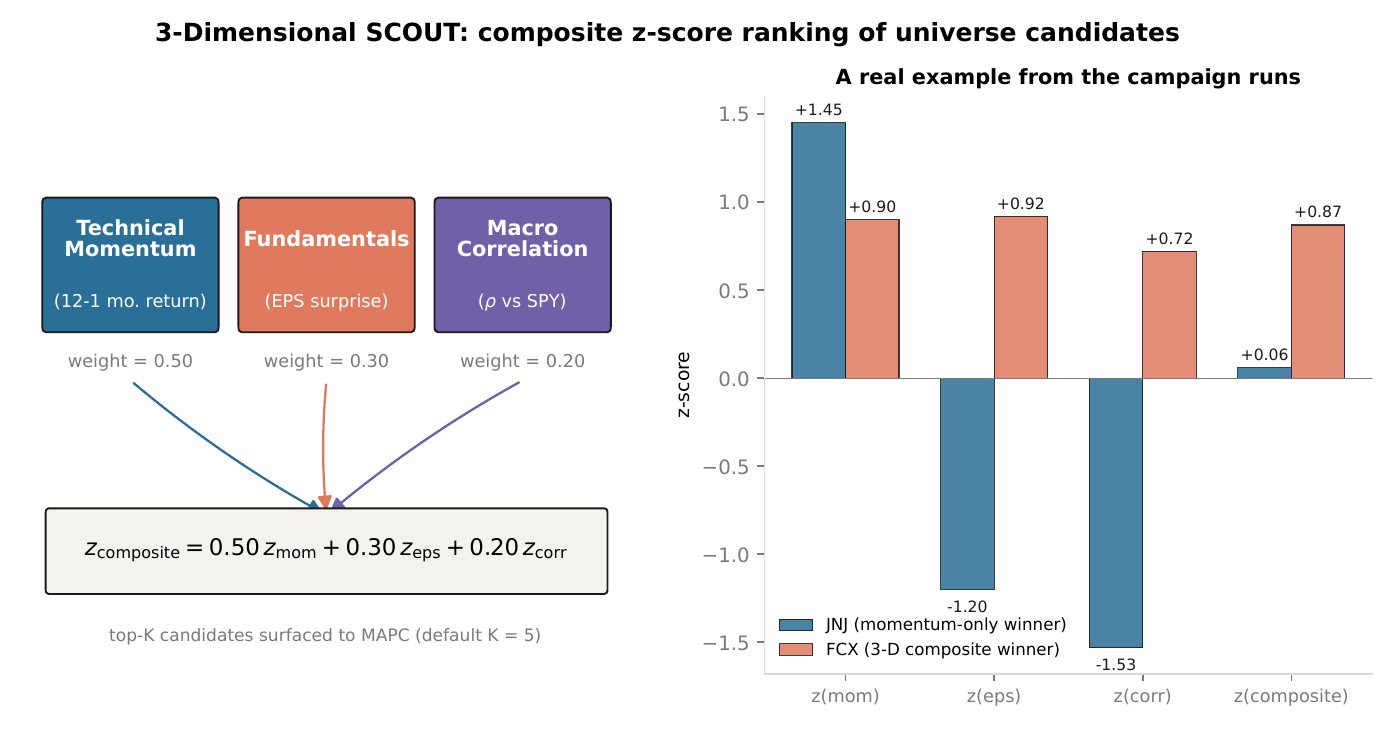}

\vspace{7pt}
\begin{minipage}{0.86\textwidth}\small \textbf{Figure 3.}\textit{ SCOUT's composite z-score ranking (momentum, earnings surprise, macro correlation). Right panel, real campaign example: JNJ leads on momentum alone, but a stronger earnings surprise and higher market participation make FCX the composite winner.}\end{minipage}
\end{figure}

\subsubsection{3.4 Cost-Aware Execution}\label{cost-aware-execution}

Cost modeling is explicit and experiment-configurable --- before costs, apparent skill may be mere over-trading (§1). \textbf{Four market-impact models} (fixed BPS, square-root, Almgren--Chriss, structural) are selectable per \textbf{cost tier} (Appendix A, Table A1), enabling the \textbf{cost-tier sensitivity ablation} (§7.7). The default LOW tier's commission and spread legs are validated against live broker paper fills (§5.2); its impact leg is parameterized (square-root), not fitted, since paper fills embed no market impact.

Position-level PnL is tracked via \textbf{weighted-average cost basis}. On a buy of \(q_{\text{fill}}\) shares at \(p_{\text{fill}}\) into a position of \(q_{\text{old}}\) at \(\bar{c}_{\text{old}}\):

\[\bar{c}_{\text{new}} = \frac{q_{\text{old}} \cdot \bar{c}_{\text{old}} + q_{\text{fill}} \cdot p_{\text{fill}}}{q_{\text{old}} + q_{\text{fill}}}\]

On a sell, \(\text{PnL}_{\text{realized}} = (p_{\text{fill}} - \bar{c}) \cdot q_{\text{fill}} - c_{\text{tx}}\). \textbf{Daily financing costs} accrue for leveraged/short positions, priced off the round's FRED Fed-Funds series plus a broker spread --- structurally zero under the default long-only, leverage-≤1 mandate. The realized cost decomposition feeds the cost-awareness metrics (§6.3) and the execution-cluster ablation (§7.7).

\subsubsection{3.5 Strategy Consistency Measurement}\label{strategy-consistency-measurement}

Mandate drift voids the capability signal even when returns are strong (§1), so consistency is measured, not assumed, via a per-round \textbf{ConsistencyScore} \(\mathcal{C}_t\):

\[\mathcal{C}_t = 0.25 \cdot (1 - d_{\text{style}}) + 0.25 \cdot \tau_{\text{consist}} + 0.25 \cdot s_{\text{adhere}} + 0.25 \cdot a_{\text{target}}\]

with components \textbf{style drift} (\(d_{\text{style}}\)), \textbf{turnover consistency} (\(\tau_{\text{consist}}\)), \textbf{signal adherence} (\(s_{\text{adhere}}\)), and \textbf{mandate alignment} (\(a_{\text{target}}\)), each in {[}0,1{]}, recoverable from the audit trail (Appendix D). Equal weights and the 0.7 threshold are fixed defaults, not tuned against returns; when \(\mathcal{C}_t < 0.7\) a \textbf{drift warning} is injected with per-asset delta cap \(|\delta_i| \leq \bar{\delta}/2\). Components feed the Composure axis (§6.4); Figure 4 shows the decomposition.

\begin{figure}[tbp]\centering
\includegraphics[width=0.9\linewidth,keepaspectratio]{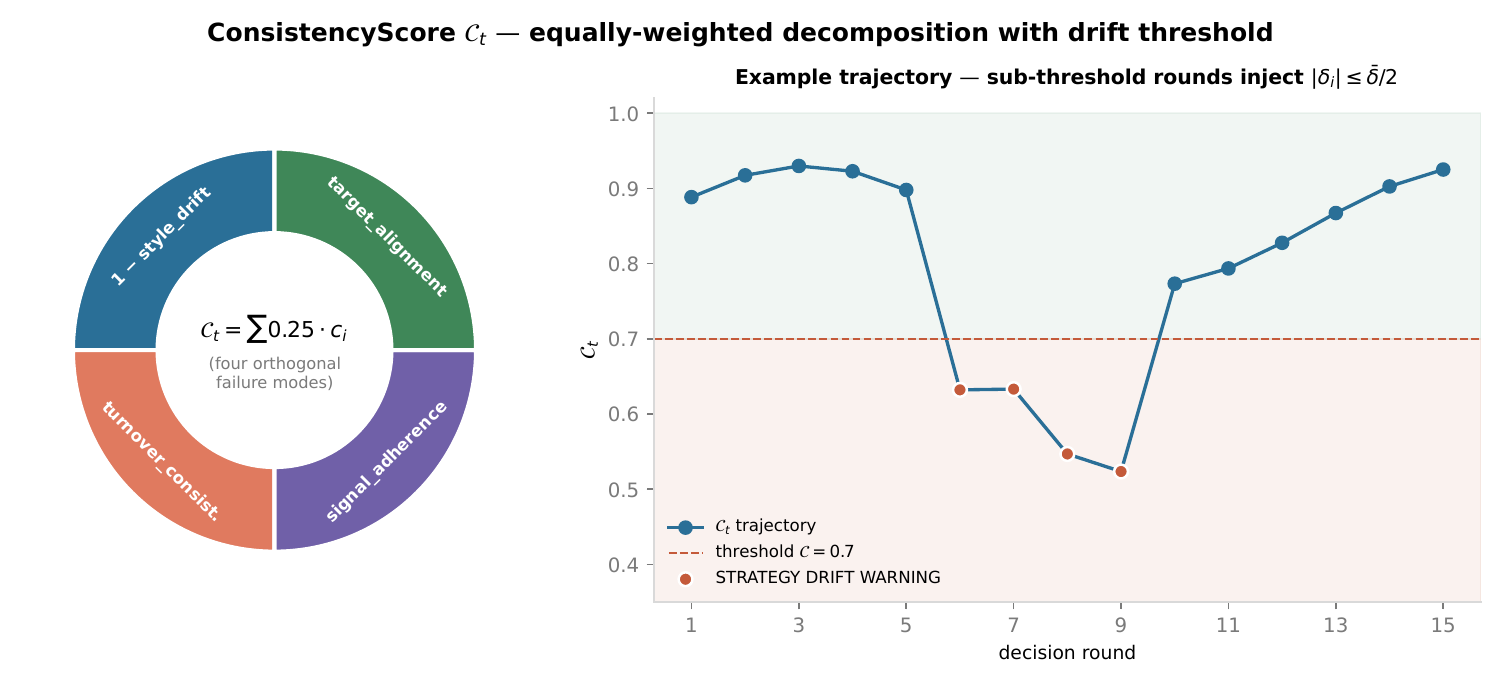}

\vspace{7pt}
\begin{minipage}{0.86\textwidth}\small \textbf{Figure 4.}\textit{ ConsistencyScore decomposition into its four components and the drift-warning threshold.}\end{minipage}
\end{figure}

\subsubsection{3.6 Structured Multi-Modal Memory System}\label{structured-multi-modal-memory-system}

A \textbf{three-tier memory architecture} (Figure 5) accumulates and generalizes behavioral patterns across sessions:

\begin{itemize}
\tightlist
\item
  \textbf{Working memory}: current observation pack plus the most recent \(N=3\) rounds.
\item
  \textbf{Episodic memory}: structured event-action-outcome records, decaying at \(\alpha_e = 0.95\)/round.
\item
  \textbf{Semantic memory}: generalized cross-asset patterns consolidated from episodic records, \(\alpha_s = 0.98\).
\end{itemize}

Every \(k=12\) rounds, a consolidation pass merges episodic records sharing asset/event/outcome patterns. \emph{memory\_provenance} logs which memories influenced each round, enabling the memory ablation (§7.7). The consistency leg that memory produces feeds the Composure axis (§6.4). At retrieval, decay enters the retrieval-relevance score as the recency term \(\alpha^{\Delta\text{round}}\), alongside asset/event/regime overlap. Both rates are \textbf{fixed defaults, not tuned against returns}, encoding only the prior that episodic records fade faster than semantic patterns (\(\alpha_e=0.95\) and \(\alpha_s=0.98\) per round correspond to half-lives of ≈13.5 and ≈34 rounds).

\begin{figure}[tbp]\centering
\includegraphics[width=0.9\linewidth,keepaspectratio]{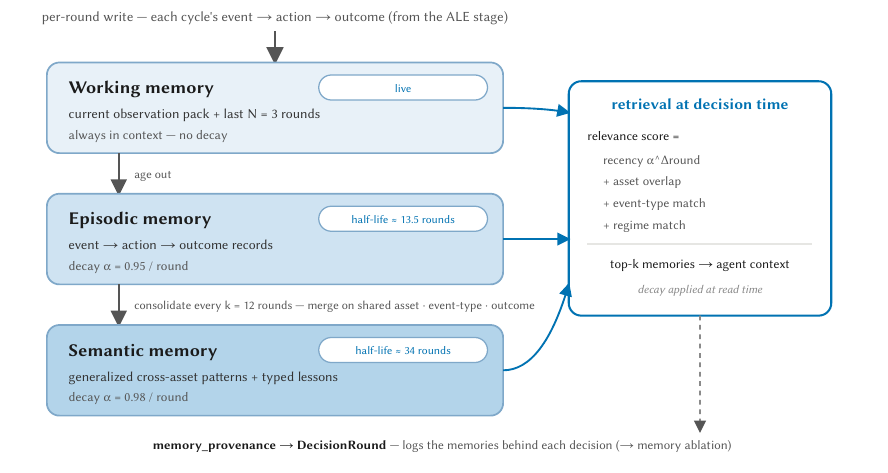}

\vspace{7pt}
\begin{minipage}{0.86\textwidth}\small \textbf{Figure 5.}\textit{ Working / episodic / semantic memory tiers with decay rates and the periodic consolidation pass.}\end{minipage}
\end{figure}

\subsubsection{3.7 Knowledge Synthesis and Decision Pipeline}\label{knowledge-synthesis-and-decision-pipeline}

The \textbf{MAPC decision} reconciles a \textbf{quantitative optimizer proposal} with the \textbf{PM agent's tactical overlay} through four layers: \textbf{quantitative proposal} (\(w_{\text{opt}}\)), \textbf{budget-neutral agent overlay} (\(\delta\) with conviction scores), \textbf{constrained projection} onto the feasible set \(\mathcal{F}\), and \textbf{Risk Officer validation} with an execution gate. The overlay \(\delta\) is recorded pre-projection as \emph{mapc\_delta\_proposed} and after projection as \emph{mapc\_delta\_projected}; \emph{mapc\_adjustments\_clipped} counts clips. These fields let the scorecard measure \emph{intent} (Acuity, Coherence on proposed deltas) separately from the optimizer/MAPC machinery that may overwrite it.

\paragraph{Dynamic Universe Management}\label{dynamic-universe-management}

The top-\(K\) \textbf{SCOUT candidates} are surfaced to the PM alongside current holdings, enabling \textbf{agent-driven rotation} under a per-round addition cap, a minimum holding-period guard, and the turnover budget. A \textbf{post-parse cross-validation pass} resolves logical contradictions.

\subsubsection{3.8 Intertemporal Sentiment Trajectory and News Persistence}\label{intertemporal-sentiment-trajectory-and-news-persistence}

Prior benchmarks treat news as point-in-time context; CLQT surfaces \emph{how} sentiment evolves. \emph{gather\_news()} uses a \textbf{dynamic lookback} \(L_t = \max(7,\, t_{\text{as\_of}} - t_{\text{last\_rebalance}} + 1\text{ day})\) and a persistent \textbf{NewsStore}. The store deduplicates and supplements articles across runs, keeping weekend and prior-round articles available. Per-(symbol, day) \textbf{SentimentDay} objects (mean sentiment, count, most-extreme headline) form a \textbf{sentiment trajectory}. The trajectory is classified as IMPROVING/DETERIORATING/STABLE by comparing the last-3-day mean with the first-3-day mean, using the fixed default threshold \(\delta_{\text{thresh}}=0.1\). The PM receives the trajectory table (Figure 6), which is stored in \emph{DecisionRound.sentiment\_trajectory} for replay. Its signal enters the informative composite used by the Acuity axis (§6.4), and the \emph{news-sentiment} ablation (§7.7) toggles the channel.

\begin{figure}[tbp]\centering
\includegraphics[width=0.9\linewidth,keepaspectratio]{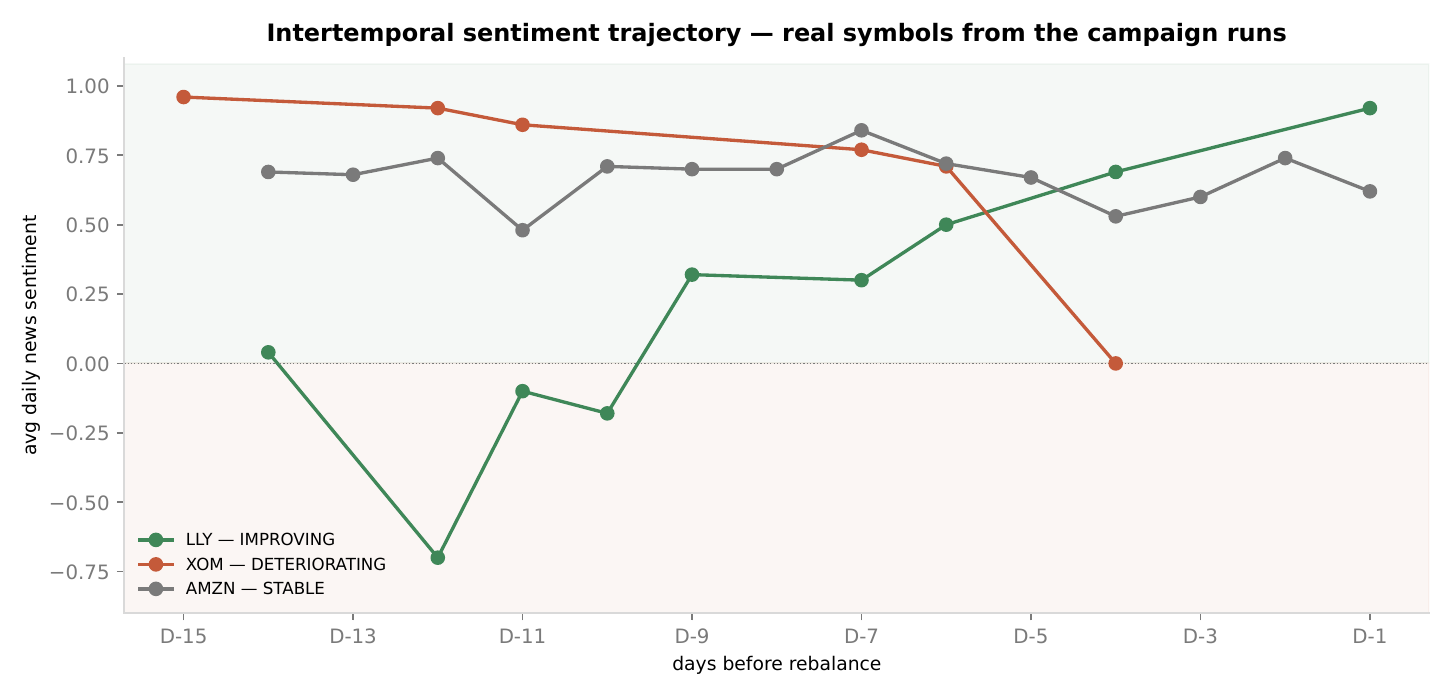}

\vspace{7pt}
\begin{minipage}{0.86\textwidth}\small \textbf{Figure 6.}\textit{ Per-symbol sentiment trajectories across the inter-rebalance window with improving / deteriorating / stable classification, as surfaced to the PM. Real campaign symbols: LLY (improving), XOM (deteriorating), AMZN (stable).}\end{minipage}
\end{figure}

Every component thus emits audit fields a later section reads --- scorecard axes (§6.4), behavioral method (§6.5), ablation grid (§7.7). None is trustworthy unless no decision ever saw the future; §4 supplies that guarantee.

\subsection{4. Data Integrity}\label{data-integrity}

Apparent agent
alpha can largely dissolve once look-ahead bias is controlled (Li et al.~2025c; Zhu et al.~2026). CLQT therefore treats two
guarantees as preconditions: no decision may see the future (§4.1--§4.3), and the
record of what each decision saw and did must be independently checkable (§4.4).

\subsubsection{4.1 TimeGate: Point-in-Time Enforcement}\label{timegate-point-in-time-enforcement}

All data access passes through \textbf{TimeGate}, enforcing \(\forall d \in \mathcal{D}_t:\ \text{timestamp}(d) \leq t_{\text{as\_of}}\). For live mode an \emph{intraday\_cutoff} freezes access \(N\) minutes before the decision timestamp. The TimeGate applies uniformly to all modalities, including memory retrieval and the \textbf{NewsStore} supplement pass.

\subsubsection{4.2 Contamination Prevention}\label{contamination-prevention}

\textbf{Post-cutoff windows.} Per-model eligibility filtering ensures backtest data postdates each model's training cutoff. \textbf{Universe integrity.} Point-in-time universe membership avoids survivorship bias.

\subsubsection{4.3 Contamination Detection}\label{contamination-detection}

Two complementary probes are implemented: canary documents and future-shift leakage probes. Canary documents are synthetic disclosures with unique identifiers detected via substring and fuzzy matching. Future-shift leakage probes pair baseline and probe decisions and measure their normalized L1 weight divergence. These detection probes are part of the substrate but were not exercised in this single, benign campaign. Contamination resistance here rests on the point-in-time gate and per-model cutoff filtering (§4.2), the live post-cutoff track (§7.9), and the verified hash chain (§4.4). Active adversarial/canary probing to stress the capability axes directly is a hardening direction (§8.6).

\subsubsection{4.4 Recompute-Verifiable Audit Trail}\label{recompute-verifiable-audit-trail}

Each \textbf{DecisionRound} is sealed into a \textbf{recompute-verifiable hash chain} using a content hash over the inputs
the agent saw and a content hash over the decision it took. Each round's hash is folded into the next.
Verification re-derives every hash from the persisted record and checks both that it matches the stored
value and that the chain linkage holds. An inconsistent edit breaks the chain at the first altered record. This
claim is deliberately scoped: the chain checks the integrity of a provided store; it is not an externally
anchored cryptographic signature and does not provide non-repudiation against wholesale replacement and
recomputation of the store. Every campaign run passes this end-to-end integrity verification.

\subsection{5. Backtest and Live Evaluation Protocols}\label{backtest-and-live-evaluation-protocols}

With the substrate (§3) and its integrity guarantees (§4) in place, this section defines the two tracks
that generate the paper's evidence --- a contamination-aware backtest (§5.1) and a broker paper-trading
live track (§5.2) --- together with the rebalancing semantics (§5.3) and the protocol for comparing across
tracks (§5.4).

\subsubsection{5.1 Backtest Track: Contamination-Aware Period Design}\label{backtest-track-contamination-aware-period-design}

Rather than a single fixed period, CLQT assigns models to evaluation groups by training cutoff: Group B (cutoff more than 2 years ago) → a 2-year backtest from the cutoff; Group A (cutoff 1--2 years ago) → a 1-year backtest. Backtest parameters: 252-day warmup, \$1M initial capital (configurable), bi-weekly rebalance (the 26-round cadence of §7.2), LOW cost tier, deterministic given the configuration. In this paper's campaign every evaluated model's cutoff precedes the single 1-year window (§7.2), so the longer 2-year Group B path, though supported, is not exercised by this cohort.

\textbf{Universe and starting book.} The investable universe is derived from the S\&P 100 --- large, liquid, survivorship-controlled names with point-in-time membership. Each run opens from a core book of roughly 14 equal-weighted holdings, against a candidate pool of \textasciitilde58 additional names (≈72 symbols in total) that the SCOUT stage (§3.3) ranks for rotation. The universe is deliberately broad enough to permit meaningful selection and turnover, yet bounded so the cohort stays comparable across models and the news/fundamentals channels remain tractable within free-tier API limits.

\subsubsection{5.2 Live Track: Broker Paper-Trading Architecture}\label{live-track-broker-paper-trading-architecture}

A live track serves two purposes: (1) genuine contamination resistance --- decisions execute on data that
postdates every model's training cutoff --- and (2) a check that the benchmark's execution abstractions map
onto a real broker's order lifecycle. CLQT integrates a commercial broker's paper-trading API. Each
session gathers and decides with the decision timestamp fixed before that day's market open. A Monday
session's gather window covers the weekend. Using the backtest's pre-open as-of convention keeps
the two tracks comparable. The session then previews and submits orders at the open, polls fills, reconciles
the resulting book from the broker's positions and cash ledger, and locks the portfolio between rebalances.
A dedicated execution agent optimizes only cost and fill quality --- the target weights are fixed by the
upstream decision. The broker integration is part of the artifact set available on request (Appendix G).

\subsubsection{5.3 Rebalancing vs.~Re-Optimization}\label{rebalancing-vs.-re-optimization}

CLQT distinguishes rebalancing (mechanical weight restoration) from re-optimization (full pipeline producing a new target). It does so through three rebalance modes --- calendar, threshold, and agent-discretion --- which provide a substrate hook for evaluating \emph{when}-to-trade judgment. This paper's campaigns run calendar mode throughout (bi-weekly backtest, daily live); the other two modes are supported but not exercised here.

\subsubsection{5.4 Cross-Track Comparison Protocol}\label{cross-track-comparison-protocol}

In this paper's campaign the two tracks deliberately differ: a year of bi-weekly backtest rounds versus
four weeks of daily live rounds on a newer model generation. The cross-track comparison therefore runs at the
\emph{decision} level on track-invariant axes rather than on returns. The general protocol is broader: when the
two tracks share a window and configuration, they are reported on the same metric suite so discrepancies
are attributable. A backtest Sharpe that exceeds live Sharpe by more than \textasciitilde0.2 (a conservative
heuristic) flags possible contamination. Fill-price divergence between simulated and live execution
gauges cost-model realism; this comparison was used to validate the LOW cost tier (§3.4). The live
campaign itself --- four trading weeks, eighteen valid NYSE days, now complete --- is reported in §7.9.
With both tracks defined, §6 specifies what is measured from them.

\subsection{6. Evaluation Framework}\label{evaluation-framework}

This section defines what is measured: the ablation dimensions (§6.1), the passive baseline suite (§6.2), the metric dictionary (§6.3), the five-axis capability scorecard that is the central read-out (§6.4) with its held-out judge protocol (§6.4.1), and the behavioral-interpretability companion (§6.5).

\subsubsection{6.1 Ablation Dimensions}\label{ablation-dimensions}

CLQT supports 12 ablation dimensions --- nine module flags, two config overrides (\emph{cost-tier} ∈ \{zero, low, medium, high\}; \emph{rebalance-mode} ∈ \{calendar, threshold, agent-discretion\}), and the \emph{skill-mode} autonomy switch (structured vs.~autonomous) --- grouped in Table 3 by what each isolates.

{\def\LTcaptype{none} 
\begin{longtable}[]{@{}
  >{\raggedright\arraybackslash}p{(\linewidth - 4\tabcolsep) * \real{0.1977}}
  >{\raggedright\arraybackslash}p{(\linewidth - 4\tabcolsep) * \real{0.3837}}
  >{\raggedright\arraybackslash}p{(\linewidth - 4\tabcolsep) * \real{0.4186}}@{}}
\toprule\noalign{}
\begin{minipage}[b]{\linewidth}\raggedright
Group
\end{minipage} & \begin{minipage}[b]{\linewidth}\raggedright
Dimensions
\end{minipage} & \begin{minipage}[b]{\linewidth}\raggedright
What it isolates
\end{minipage} \\
\midrule\noalign{}
\endhead
\bottomrule\noalign{}
\endlastfoot
Information & \emph{news-sentiment},\allowbreak{} \emph{world-macro},\allowbreak{} \emph{asset-class-data} & what data the agent can access \\
Decision quality & \emph{advanced-quant},\allowbreak{} \emph{investment-target} & optimization and mandate enforcement \\
Execution & \emph{cost-modeling},\allowbreak{} \emph{systematic-execution} & trade costing and execution strategy \\
Learning & \emph{reflection},\allowbreak{} \emph{memory} & whether the agent learns across rounds \\
Regime & \emph{cost-tier},\allowbreak{} \emph{rebalance-mode} & environmental conditions \\
Autonomy & \emph{skill-mode} & process prescription vs.~agent-designed workflow \\
\end{longtable}
}

\begin{center}\begin{minipage}{0.86\linewidth}\small
\textbf{Table 3.}\textit{ Ablation dimension taxonomy.}
\end{minipage}\end{center}

\subsubsection{6.2 Baselines and Benchmark Comparison}\label{baselines-and-benchmark-comparison}

Every run is scored against a passive suite computed automatically: the index ETFs SPY (S\&P 500 proxy) and IEF (7--10y Treasuries); a synthetic 60/40 (0.6·SPY + 0.4·IEF); and an equal-weight portfolio of the agent's own investable universe (EW-Universe). The suite also includes four algorithmic baselines --- buy-and-hold, equal-weight (rebalanced), risk-parity (inverse-vol), and 12-1 momentum. For each of the eight, the run reports cumulative/excess return, information ratio, and tracking error. The headline question --- \emph{does any agent beat the cap-weighted index net of costs?} --- is answered against this suite, not a single benchmark.

\subsubsection{6.3 Metric Dictionary}\label{metric-dictionary}

Table 4 lists the recorded metric suite, all recoverable from the audit trail.

{\def\LTcaptype{none} 
\begin{longtable}[]{@{}
  >{\raggedright\arraybackslash}p{(\linewidth - 2\tabcolsep) * \real{0.2442}}
  >{\raggedright\arraybackslash}p{(\linewidth - 2\tabcolsep) * \real{0.7558}}@{}}
\toprule\noalign{}
\begin{minipage}[b]{\linewidth}\raggedright
Category
\end{minipage} & \begin{minipage}[b]{\linewidth}\raggedright
Metrics
\end{minipage} \\
\midrule\noalign{}
\endhead
\bottomrule\noalign{}
\endlastfoot
Performance & CumRet,\allowbreak{} AnnRet,\allowbreak{} Vol,\allowbreak{} Sharpe,\allowbreak{} Sortino,\allowbreak{} Calmar,\allowbreak{} MaxDD \\
Benchmark-relative & excess,\allowbreak{} IR,\allowbreak{} TE vs SPY/\allowbreak{}IEF/\allowbreak{}60-40/\allowbreak{}EW-Universe + 4 baselines; ``beat /\allowbreak{}8'' \\
Cost-awareness & turnover,\allowbreak{} bps decomposition (spread/\allowbreak{}commission/\allowbreak{}slippage/\allowbreak{}impact),\allowbreak{} net-vs-gross,\allowbreak{} financing drag \\
Governance & RO-reject \%,\allowbreak{} no-trade \%,\allowbreak{} MAPC clips/\allowbreak{}round,\allowbreak{} constraint violations \\
Consistency & composite \(\mathcal{C}\),\allowbreak{} style drift,\allowbreak{} turnover consistency,\allowbreak{} signal adherence \\
Closed-loop learning & lessons/\allowbreak{}round,\allowbreak{} intra-run learning curve \\
Efficiency & \$/\allowbreak{}round,\allowbreak{} in/\allowbreak{}out/\allowbreak{}reasoning tokens,\allowbreak{} latency,\allowbreak{} \$ per 1\% excess \\
Reliability & completion,\allowbreak{} parse-fail,\allowbreak{} timeout,\allowbreak{} output-cap-hit \\
Integrity & chain-verify pass,\allowbreak{} leakage divergence,\allowbreak{} canary recall,\allowbreak{} cutoff compliance \\
\end{longtable}
}

\begin{center}\begin{minipage}{0.86\linewidth}\small
\textbf{Table 4.}\textit{ Metric dictionary (all recoverable from the audit trail; cf.~Appendix D). Sharpe and Sortino use a zero risk-free rate --- the convention of the agent-trading benchmarks we compare against (Chen et al.~2025; Yu et al.~2025; Li et al.~2025b); a risk-free leg would shift all agents' ratios down similarly without changing rankings.}
\end{minipage}\end{center}

\subsubsection{6.4 The Five-Axis Diagnostic Capability Scorecard (APM-CS)}\label{the-five-axis-diagnostic-capability-scorecard-apm-cs}

The \textbf{A}utonomous-\textbf{P}ortfolio-\textbf{M}anagement \textbf{C}apability \textbf{S}corecard (APM-CS) is the instrument's read-out: five capability axes scored from the audit trail, ordered by how fundamental they are to competent autonomous portfolio management. Each axis is a finance instantiation of a general agent-evaluation capability, so trading-agent findings generalize to statements about the model's reasoning. Outcome (returns) is \emph{not} an axis --- it is a result the axes explain.

\textbf{D1 · Coherence} --- \emph{general capability:} global, systems-level reasoning vs.~local greedy pattern-matching. At the level of reasoning, this is analogous to systematic alignment vs.~turn-by-turn instruction-following: does the conclusion follow from the agent's own analysis? \emph{Finance:} the chosen allocation follows from the agent's stated research and the signals it cited. \emph{Synthesis:}
\[D_1 = 0.5\,A_{\mathrm{sig}} + 0.5\,J_{\mathrm{judge}}\]
where \(A_{\mathrm{sig}}\) is signal-action agreement: the fraction of proposed trades (\emph{mapc\_delta\_proposed}, pre-MAPC) whose direction matches the sign of that name's cross-sectional informative composite (below); \(J_{\mathrm{judge}}\) is the held-out LLM judge's score (§6.4.1), normalized to {[}0,1{]}.

\textbf{D2 · Acuity} --- \emph{general:} salience attention vs.~distraction, i.e.~endurance under polluted / misleading context. \emph{Finance:} attend to informative signals amid a large noisy universe and a flood of news. \emph{Synthesis}, on proposed adjustments (not final weights, which the optimizer/MAPC confound):
\[D_2 = \rho_s(\Delta_{\mathrm{prop}}, I) - \rho_s(\Delta_{\mathrm{prop}}, N)\]
where \(\rho_s\) is Spearman correlation and \(I=z(m_{12-1})+z(e_{\mathrm{surp}})+z(n_{\mathrm{sent}})\) is the informative composite (12-1 momentum, earnings surprise, news sentiment). The two-leg noise composite, \(N=\operatorname{mean}\big(z(\sigma_{\mathrm{realized}}),\,z(\mathrm{RSI}_{14})\big)\), combines volatility level, which carries no directional alpha at this horizon, and the 14-day RSI oscillator, which is salient in the data feed but horizon-mismatched to a bi-weekly rebalance. Short-term \emph{reversal} is deliberately excluded: it is a documented (if weak) alpha factor, so penalizing it would mislabel signal as noise. A single-leg (realized-vol-only) variant is a robustness column (Appendix F.1b). Each term is z-scored cross-sectionally per round and averaged over rounds with ≥3 proposed trades. Positive ⇒ intent tracks alpha-bearing signals more than distractors; near-zero or negative ⇒ chasing noise.

\textbf{D3 · Composure} --- \emph{general:} robustness / consistency under input perturbation vs.~reactive overcorrection. \emph{Finance:} respond to volatility and misleading data with discipline, not churn. \emph{Synthesis}:
\[D_3 = \operatorname{mean}\left(1 - s_{\mathrm{react}},\ c_{\mathrm{cons}}\right)\]
where \(s_{\mathrm{react}}\) is the non-negative regression coefficient of per-round one-way turnover on a market-volatility proxy (cross-sectional mean \emph{realized\_vol}). The coefficient is squashed to \([0,1]\) as \(\mathrm{slope}/(\mathrm{slope}+\text{median turnover})\). The term \(c_{\mathrm{cons}}\) is the strategy-consistency leg when available. We use the mean rather than a product because a product lets one weak leg collapse the axis. The consistency leg is memory-produced, so a product would also be undefined under the memory ablation (§7.7). The mean falls back to the memory-free \emph{calm} leg when \emph{consistency} is absent, so Composure is defined for \emph{every} configuration. (Style-drift is ≈0 cohort-wide, so it is reported separately.)

\textbf{D4 · Discipline} --- \emph{general:} systematic alignment to a standing mandate vs.~turn-by-turn instruction-following --- does the agent self-regulate to a policy even absent a hard guardrail? \emph{Finance:} respect caps/turnover/risk budgets \emph{without} relying on the MAPC clip. \emph{Synthesis} (bounded to \([0,1]\)):
\[D_4 = \tfrac{1}{2}\,r_{\mathrm{self}} + \tfrac{1}{2}\,(1-d_{\mathrm{clip}})\]
where \(r_{\mathrm{self}}\) is the self-constrained rate: the fraction of rounds whose target satisfies constraints pre-MAPC. \(d_{\mathrm{clip}}=\min(1,\bar{a}_{\mathrm{clip}}/\kappa)\) is clip dependence, measuring reliance on the reconciler to \emph{force} the book inside the mandate (\(\kappa=10\), a fully-saturated structured round; \(\bar{a}_{\mathrm{clip}}\) is the mean number of clipped MAPC adjustments). Both legs lie in \([0,1]\), so Discipline does too. (A turnover-cap leg is excluded: campaign turnover \textasciitilde0.15 sits far below any real cap --- a non-discriminating constant.) \emph{Mode confound:} RO-approval is meaningful only in \emph{structured}; \emph{autonomous} sets \emph{risk\_approved=True} by construction --- so Discipline is compared within mode.

\textbf{D5 · Reliability} --- \emph{general:} protocol/format adherence + operational robustness + efficiency --- the analogue of an agent completing a tool-use task in valid form, every time, at reasonable cost. \emph{Finance:} close the loop in valid schema at acceptable compute. \emph{Synthesis:}
\[D_5 = r_{\mathrm{complete}}\,(1-r_{\mathrm{parse}})\,(1-r_{\mathrm{trunc}})\]
where \(r_{\mathrm{complete}}\) is completion rate, \(r_{\mathrm{parse}}\) is parse-fail rate, and \(r_{\mathrm{trunc}}\) is truncation rate. Efficiency is reported beside it (\$/round, reasoning-tokens/round, latency). Truncation is detected against the \emph{role-level} output cap recorded on each call (e.g.~the PM's 2,048-token budget), not the global 8,192 budget alone. The completion leg is clamped at 1 so an under-specified round plan can never yield Reliability \textgreater{} 1.

\textbf{Scoring \& aggregation.} Each axis maps to a cohort percentile; the five form a radar fingerprint (Figure 7), and Table 5 tabulates the profile plus a composite APM-CS (mean of the five percentiles). The thesis bar requires winning the large majority of axes against all rivals across regime sub-periods and runs. D2 and D4 are mode-confounded: raw values are compared only within mode, and cross-mode APM-CS contrasts are descriptive, not clean scaffold-effect estimates. APM-CS is also cohort-relative --- adding or removing configurations rescales it --- so the raw D1--D5 values are the portable read-out.

\paragraph{6.4.1 Held-out LLM-as-Judge for Coherence}\label{held-out-llm-as-judge-for-coherence}

The judge component of Coherence is scored by a held-out, non-cohort model --- \emph{not} in the evaluated set --- to avoid the documented LLM-as-a-judge self-preference bias. Per round, the judge receives a compact, audit-derived prompt containing the agent's stated analysis (research brief and/or optimizer rationale), the top names by informative-composite strength with their signals, the chosen target weights, and the proposed changes. It returns a 1--5 score for the single question \emph{``does the chosen allocation follow coherently from the stated analysis and signals?''} and is told to ignore profitability and penalize self-contradiction and untethered allocations. Scores are cached by \emph{round\_id} (Appendix G); reasoning is enabled on the judge deliberately.

\textbf{Two-judge protocol (per-track primaries + cross-corroboration).} \emph{Held-out} means held out of the cohort being scored. The scorecard builder enforces this by refusing to run when the judge matches any scored run. Since minimax-m3 sits in the \emph{live} panel, the tracks use different primaries. \textbf{minimax-m3 judges the backtest cohort} (which does not include it), while \textbf{GLM-5.2, a different-family reasoning model outside both cohorts, judges the live track}. The two are calibrated on shared backtest rounds (Spearman ρ = 0.65, 97 \% within one point on the 1--5 rubric). Their scores are not blended --- each track reports its own primary. Robustness of the headline coherence gap to a full judge swap is established in §7.9. A blinded human-anchoring study corroborates directly: on 50 blind-rated rounds, the judge's ordering tracks a human rater as closely as the two judges track each other (Appendix F.5).

\subsubsection{6.5 Behavioral Interpretability: Method}\label{behavioral-interpretability-method}

The same audit trail that powers the quantitative scorecard (§6.4) also supports a qualitative, behavioral read: each model's decision behaviors, stated reasoning, and recurring patterns are rendered as a human-legible ``decision personality''. This probes the LLM black box through \emph{observable actions and self-reports} rather than weights. Every round is a complete, hash-chained, verifiable record (§4.4). Captured per round, per model × mode (audit-field mapping: Appendix D, Table D1):

\begin{itemize}
\tightlist
\item
  \textbf{Stated reasoning} --- the agent's narrative of \emph{why}: research and analyst briefs, allocation rationale, risk-officer notes, post-round reflection.
\item
  \textbf{Metacognition and learning} --- how a model abstracts experience into typed, reusable ``lessons'', and whether they are later retrieved back into context.
\item
  \textbf{Intent vs.~guardrails} --- the gap between what the model \emph{proposed} and what the reconciler \emph{allowed}.
\item
  \textbf{Inter-agent dynamics} --- how the PM and Risk Officer negotiate, and when the committee overrides itself.
\item
  \textbf{Information foraging} --- which tools each model calls, in what order, and how widely it explores the candidate universe.
\item
  \textbf{Cognitive effort} --- where a model ``thinks hardest'' (reasoning-token and latency budget per role) and whether effort tracks decision quality.
\item
  \textbf{Struggle signatures} --- where and why valid decisions fail: retry loops, parse failures, constraint violations.
\end{itemize}

Method. Thematic coding of the reasoning corpus (optionally LLM-assisted via a \emph{held-out} summarizer mirroring §6.4.1), frequency analysis of intent-vs-clip and committee overrides, per-model tool-use fingerprints, lesson formation and reuse, and illustrative anecdotes --- each tied to a hash-verified round, hence reproducible.

Scope (honest caveat). We analyze stated and observable reasoning artifacts and behavioral metadata, not raw hidden chain-of-thought: providers generally do not return the thinking trace, and we persist reasoning-token \emph{counts}, not trace text. The interpretability is thus of \emph{behavior + self-report} --- auditable, one layer above hidden cognition.

With the read-out defined, §7 applies it to the campaigns.

\subsection{7. Experiments}\label{experiments}

The experiments comprise the year-long backtest campaign on a 5-model cohort, the 13-configuration
ablation grid, and the completed live track --- a \emph{map of limitations}, not a model ranking. The purpose is to show that the five-axis scorecard discriminates decision \emph{quality} where outcome metrics conflate it and surfaces model-specific failures invisible to returns. By the instrument's own standard (\emph{consistent dominance across axes and sub-periods}), it also resists premature ranking even with repeated-run evidence. Rows
report means over repeated runs (seeds); bootstrapped CIs and Kendall-τ rank-stability are a planned
broader-run pass (§8.6).

\subsubsection{7.1 Research Questions, Answers, and Where the Evidence Lives}\label{research-questions-answers-and-where-the-evidence-lives}

Each research question is stated
here with its short answer and the section that carries the evidence (findings F1--F5 in §7.3 unpack
the headline table). Each question is falsifiable in the campaign's own terms: RQ1 fails if the
scorecard merely re-ranks Sharpe; RQ2 if any configuration dominates all five axes; RQ4 if capability
profiles are mode-invariant; and RQ5 if any module knockout clears the repeated-run noise band. The
answers report where the evidence ran against the instrument's own design (the HIGH cost tier failing
its calibration, RQ6) as plainly as where it confirmed it.

\begin{itemize}
\tightlist
\item
  \textbf{RQ1 (instrument validity):} Does the scorecard discriminate decision \emph{quality} where outcome metrics
  conflate it? --- \textbf{Yes.} The scorecard separates configurations that returns conflate: the nominal Sharpe
  ``winner'' carries the cohort's lowest composite and a Reliability of 0.19 (§7.3-F1), and module removals
  that leave Sharpe flat register plainly on the capability axes (§7.7).
\item
  \textbf{RQ2 (capability ≠ outcome):} Is the Sharpe leader the capability leader, and does any configuration
  dominate all five axes? --- \textbf{No and no.} The capability leader (deepseek·structured, APM-CS 72.9) is
  mid-pack on Sharpe, and no configuration encloses all five axes (§7.3-F1, §7.4) --- which is why, by the
  instrument's own standard, the cohort is not cleanly rankable.
\item
  \textbf{RQ3 (model-specific limitations):} Does the audit trail surface model-specific failures invisible to
  returns? --- \textbf{Yes.} Three distinct backtest reliability deficits with three different mechanisms (§7.5),
  and two distinct live hold mechanisms that a targeted scaffold fix then eliminated (§7.10).
\item
  \textbf{RQ4 (scaffold effect):} How does the structured↔autonomous mode shape the capability profile? ---
  \textbf{Model- and axis-dependently.} The backtest composite descriptively favors structure for the reliable
  configurations while the live cohort uniformly favors autonomy. The mode-comparable sensitivity
  analysis reconciles the two readings (§7.6, §7.9, Appendix F.4).
\item
  \textbf{RQ5 (module importance):} Can module value be ranked from returns? --- \textbf{No, not even with repeated
  runs.} No single-module or cluster knockout's run range clears the ±0.42 repeated-run noise band; what a
  module contributes registers on the capability and behavioral axes instead (§7.7).
\item
  \textbf{RQ6 (cost realism):} Are the cost tiers calibrated? --- \textbf{LOW yes, HIGH no.} LOW's commission and
  spread legs validate against live broker fills, while the HIGH tier's implausible per-round drag exposes
  its own mis-calibration --- the instrument auditing itself (§7.7).
\item
  \textbf{RQ7 (passive comparison):} Net of costs, does any agent beat the passive suite? --- \textbf{The defensive
  baselines yes, the cap-weighted index not cleanly.} Reliable structured configurations clear a median
  \textasciitilde6 of 8 baselines; the index shortfall reads as a deliberate de-risking trade, not absent skill (§7.8).
\end{itemize}

\subsubsection{7.2 Setup}\label{setup}

\textbf{Campaign grid.} Five models spanning families, sizes, and reasoning styles are evaluated in structured and autonomous skill modes: qwen3-235b (large MoE, non-thinking), gemini-2.5-flash-lite (small), deepseek-chat-v3.1 (mid, hybrid→non-thinking), gpt-5-mini (reasoning, effort low), and claude-haiku-4.5 (small, thinking off). Each structurally-reliable configuration (gemini, deepseek, gpt-5-mini, haiku in both modes, and qwen3 autonomous) uses five repeated seeds; results are reported as means over seeds. qwen3·structured is a single showcase seed --- seven seed attempts all failed to complete (§7.5). Period: 26 bi-weekly rounds, 2025-06-16 → 2026-06-12; S\&P-100-derived universe (14 core + 58 candidate = 72), all modules enabled, LOW cost tier, initial NAV \$1M, context capped 32k for cohort comparability, uniform output budget 8192. Coherence judge: held-out, non-cohort minimax-m3, reasoning enabled, cached. A \emph{seed} denotes an independent end-to-end repetition of the configuration: the engine is deterministic given its inputs (§5.1), so run-to-run variation originates in provider-side LLM sampling nondeterminism.

\textbf{Cutoff safety.} Every model's training cutoff precedes the 2025-06-16 start (gemini Jan'25, gpt-5-mini May'24, claude-haiku Feb'25, deepseek Mar'25); qwen3 (Jun'25) sits within a documented ≤1-month grace. The \emph{IntegrityChecker} records cutoff compliance per run.

\textbf{Ablation grid.} The single anchor is gemini-2.5-flash-lite, structured, LOW (cheap; its structured runs complete). The grid includes six module-off variants (\emph{cost-modeling, memory, reflection, news-sentiment, world-macro, advanced-quant}), each with repeated runs, plus two cost-tier variants (\emph{medium, high}, single-run). Every run completed via \emph{run\_ablation.py} (§7.7, Figure 9).

\textbf{Parsing hardening.} Before the campaign, the PM/Risk schema parsers were hardened to drop null list items (and lift bare-string items) in \emph{adjustments}/\emph{checks}. Small models (notably gemini in structured mode) emit \emph{\{``adjustments'': {[}null, \ldots{]}\}}, which previously failed validation and degraded the round. This recovers the valid content and materially improves structured Reliability across the cohort (§7.5).

\textbf{Decision retries.} \emph{max\_decision\_retries} is 3 for all campaign runs. It was raised from 1 after gemini's residual schema-unreliability caused \textasciitilde16\% optimizer-fallback rounds. Setting retries=3 cuts that to \textasciitilde0 without masking the Reliability axis (D5 still records per-attempt failures). The live timeout and \emph{decide} budgets were likewise widened for the reasoning-heavy models.

\subsubsection{7.3 Campaign Results (RQ1, RQ2)}\label{campaign-results-rq1-rq2}

The campaign's central result is one table: \textbf{the capability leader is not the Sharpe leader, and
the nominal Sharpe ``winner'' is an artifact the scorecard exposes} (Table 5); findings F1--F5 unpack
it.

\begin{center}\begin{minipage}{0.86\linewidth}\small
\textbf{Table 5.}\textit{ Per (model × mode), 26 bi-weekly rounds, hash-chains verified, sorted by APM-CS.
Means over \textbf{five seeds} per configuration (qwen3·structured: single showcase seed --- see note). Beat /8
vs \{SPY, IEF, 60/40, EW-Universe, buy-and-hold, equal-weight, risk-parity, momentum\}. D1 Coherence =
0.5·agreement + 0.5·held-out minimax-m3 judge (the backtest primary, §6.4.1); D2 Acuity uses the
two-leg noise composite (§6.4); D5 Reliability = completion·(1−parse-fail)·(1−truncation), role-cap
aware. APM-CS = mean cohort percentile across D1--D5. Axes D2/D4 are mode-confounded --- compare within
mode. Bold marks the values the analysis turns on (cohort bests and the floors/artifacts of F1--F5),
not uniformly the per-column leader.}
\end{minipage}\end{center}

{\def\LTcaptype{none} 
\begin{longtable}[]{@{}
  >{\raggedright\arraybackslash}p{(\linewidth - 18\tabcolsep) * \real{0.3721}}
  >{\raggedright\arraybackslash}p{(\linewidth - 18\tabcolsep) * \real{0.0814}}
  >{\raggedright\arraybackslash}p{(\linewidth - 18\tabcolsep) * \real{0.0814}}
  >{\raggedright\arraybackslash}p{(\linewidth - 18\tabcolsep) * \real{0.0814}}
  >{\raggedright\arraybackslash}p{(\linewidth - 18\tabcolsep) * \real{0.0581}}
  >{\raggedright\arraybackslash}p{(\linewidth - 18\tabcolsep) * \real{0.0698}}
  >{\raggedright\arraybackslash}p{(\linewidth - 18\tabcolsep) * \real{0.0581}}
  >{\raggedright\arraybackslash}p{(\linewidth - 18\tabcolsep) * \real{0.0581}}
  >{\raggedright\arraybackslash}p{(\linewidth - 18\tabcolsep) * \real{0.0581}}
  >{\raggedright\arraybackslash}p{(\linewidth - 18\tabcolsep) * \real{0.0814}}@{}}
\toprule\noalign{}
\begin{minipage}[b]{\linewidth}\raggedright
Model · mode
\end{minipage} & \begin{minipage}[b]{\linewidth}\raggedright
Sharpe
\end{minipage} & \begin{minipage}[b]{\linewidth}\raggedright
CumRet
\end{minipage} & \begin{minipage}[b]{\linewidth}\raggedright
Beat/\allowbreak{}8
\end{minipage} & \begin{minipage}[b]{\linewidth}\raggedright
D1
\end{minipage} & \begin{minipage}[b]{\linewidth}\raggedright
D2
\end{minipage} & \begin{minipage}[b]{\linewidth}\raggedright
D3
\end{minipage} & \begin{minipage}[b]{\linewidth}\raggedright
D4
\end{minipage} & \begin{minipage}[b]{\linewidth}\raggedright
D5
\end{minipage} & \begin{minipage}[b]{\linewidth}\raggedright
\textbf{APM-CS}
\end{minipage} \\
\midrule\noalign{}
\endhead
\bottomrule\noalign{}
\endlastfoot
deepseek-chat-v3.1 · structured & 2.99 & +28.6\% & 6.8 & 0.45 & 0.07 & 0.93 & 0.90 & 0.99 & \textbf{72.9} \\
claude-haiku-4.5 · structured & \textbf{3.21} & +23.4\% & 6.0 & \textbf{0.23} & \textbf{0.44} & 0.95 & 1.00 & 1.00 & 71.0 \\
gpt-5-mini · structured & 2.62 & +22.5\% & 4.6 & 0.43 & 0.08 & 0.92 & 0.79 & 1.00 & 67.4 \\
gpt-5-mini · autonomous & 1.76 & +22.2\% & 4.6 & 0.47 & 0.03 & 0.92 & 0.42 & 0.99 & 50.3 \\
gemini-2.5-flash-lite · structured & 2.31 & +24.8\% & 6.0 & 0.38 & 0.07 & 0.93 & 0.84 & \textbf{0.61} & 47.4 \\
claude-haiku-4.5 · autonomous & 1.92 & +22.5\% & 4.8 & 0.44 & 0.01 & 0.81 & 0.61 & 0.98 & 40.4 \\
qwen3-235b · autonomous & 1.20 & +12.6\% & 2.2 & 0.42 & \textbf{−0.06} & 0.91 & 0.56 & 0.99 & 39.9 \\
deepseek-chat-v3.1 · autonomous & 1.67 & +21.5\% & 4.4 & 0.42 & 0.00 & 0.90 & 0.64 & \textbf{0.62} & 35.1 \\
gemini-2.5-flash-lite · autonomous & 1.75 & +15.5\% & 3.0 & 0.47 & −0.07 & \textbf{0.77} & 0.47 & 0.95 & 32.3 \\
qwen3-235b · structured ✦ & \textbf{9.82} & +13.1\% & 8.0 & 0.33 & \textbf{−0.79} & 0.95 & 0.80 & \textbf{0.19} & 30.7 \\
\end{longtable}
}

✦ \textbf{qwen3-235b structured}: completed rounds only --- a \emph{5-round self-selected window} (mechanism in
§7.5); read its Sharpe and beat count against its Reliability (\textasciitilde0.2), per F1. The autonomous rows
(five seeds) are the trustworthy qwen3 rows.

\textbf{Findings.}

\begin{itemize}
\item
  \textbf{F1 --- Capability leader ≠ outcome leader (the thesis, in one table).} deepseek·structured tops
  APM-CS (72.9; the most balanced profile: D1 0.45, Discipline 0.90, Reliability 0.99) yet is
  \emph{mid-pack} on Sharpe (2.99). The nominal Sharpe ``winner,'' qwen3·structured at an implausible
  9.82 / beat 8-of-8, ran only 5/26 rounds with Reliability 0.19 and the cohort's lowest APM-CS
  (30.7) --- a survivorship artifact of a self-selected window, not a strategy. Among \emph{reliable}
  configs the Sharpe leader is haiku·structured (3.21; split profile in F4). A leaderboard would
  crown qwen3; the scorecard returns a \emph{profile} per configuration instead (Figure 7, Figure 8).
\item
  \textbf{F2 --- Scaffold effects are model- and axis-dependent.} The cohort-relative composite favors
  structure for the four reliable multi-mode backtest models; the live cohort uniformly favors
  autonomy. A sensitivity analysis restricted to the mode-comparable axes reconciles the two (§7.6;
  Appendix F.4).
\item
  \textbf{F3 --- Reliability deficits hidden by returns.} Three configurations carry reliability deficits
  via three mechanisms: gemini·structured's persistent schema parse-fails behind a healthy Sharpe,
  qwen3·structured's tool-loop non-termination, and deepseek·autonomous's mode-flipped reliability.
  Each is invisible to a returns-only view (§7.5, Figure 8).
\item
  \textbf{F4 --- Acuity is weak cohort-wide, with one thinly-sampled exception.} Under the two-distractor
  noise test, D2 sits within −0.07\ldots0.08 for eight of ten configurations; these values are not
  separable from near-zero at this sample size. Both qwen3 modes are negative (−0.06 autonomous
  mean; −0.79 structured showcase seed): their proposed trades track the distractors more than the
  alpha-bearing composite. The exception is haiku·structured at 0.44 (best structured point estimate;
  D2 compares within mode, §6.4). It alone tracks the informative signals over both distractors in
  the noisy 72-name universe --- yet its narrative explains its own trades least well (Coherence 0.23,
  cohort floor; §7.6). Caveat: Acuity needs ≥3 simultaneous proposed tilts to score a round; haiku·structured's
  sparse-proposal style clears that bar in one round across its five seeds --- thinly-sampled, not a
  robust cohort-best.
\item
  \textbf{F5 --- Net of realistic costs, agents clear \emph{some} of the passive suite, not the index cleanly.}
  Reliable structured runs beat a median \textasciitilde6/8 baselines; autonomous runs beat fewer --- composition of
  the beaten set and the risk-adjusted reading in §7.8.
\end{itemize}

\subsubsection{7.4 Radar Fingerprints (RQ2, continued)}\label{radar-fingerprints-rq2-continued}

\textbf{Capability is a \emph{shape}, not a scalar.} The five-axis radar (Figure 7) visualizes the §7.3 finding that no configuration encloses all five axes. The most balanced enclosure (deepseek·structured) does not top Coherence. The high-\emph{Sharpe} configurations are uneven --- strong on one or two axes, weak on another (qwen3 on Reliability, haiku on Coherence).

\begin{figure}[tbp]\centering
\includegraphics[width=0.98\linewidth,height=\textheight,keepaspectratio]{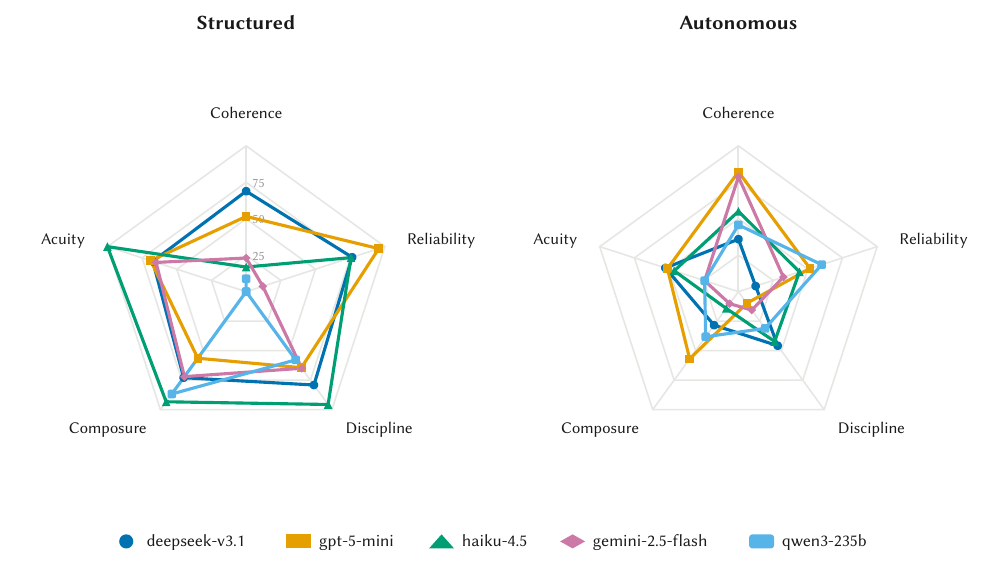}

\vspace{7pt}
\begin{minipage}{0.86\textwidth}\small \textbf{Figure 7.}\textit{ Five-axis diagnostic capability scorecard (within-cohort percentile per axis), all ten
configurations split into structured (left) and autonomous (right) panels, one colour per model. No
configuration encloses all five axes. The panels are descriptive: D2 and D4 are mode-confounded (§6.4)
and should not be interpreted as a clean structured-versus-autonomous effect.}\end{minipage}
\end{figure}

\subsubsection{7.5 RQ3: Model-Specific Reliability Failures and Their Mechanisms}\label{rq3-model-specific-reliability-failures-and-their-mechanisms}

Reliability is the axis that most cleanly separates the cohort.
Three configurations carry reliability deficits: two are genuine decision-schema failures (gemini,
deepseek-autonomous), and one is an agentic tool-orchestration failure (qwen3).
Each surfaces only in the audit trail's \emph{parse\_failures} / \emph{decision\_attempts} / \emph{completion} fields:

\begin{itemize}
\tightlist
\item
  \textbf{qwen3-235b · structured} is the reliability floor, root-caused as an \emph{agentic tool-orchestration} failure. Across \textbf{seven} seed attempts it never completes more than 10 of 26 rounds (2--10 per
  attempt; the showcase seed's Reliability is ≈0.2), after two simpler hypotheses --- output-cap
  truncation and provider throttling --- were eliminated.
  The audit trail shows the mechanism: qwen3 \emph{cannot terminate} within the structured committee's
  bounded tool loop. It re-invokes the same analysis tools with near-empty outputs until the role
  context hits its 32k cap. Forced finalization then returns null, killing the round. Decisively, the
  \emph{same} model at the \emph{same} context cap with the \emph{same} tools completes \textbf{autonomous} runs cleanly
  (five complete seeds, zero nulls), and the other four models clear the structured committee at the
  same cap. \textbf{The scaffold, not the infrastructure, flips qwen3 from unusable to usable.} We therefore
  report qwen3·structured only as a reliability/orchestration case study, not a capability row.
\item
  \textbf{gemini-2.5-flash-lite · structured} posts a healthy +24.8\% / Sharpe 2.31 (five-seed means,
  Table 5) yet Reliability 0.61. Residual parse-failures (\textasciitilde10/26 rounds on at least one attempt)
  persist across all its runs even with retries enabled. Retries recover the round, but the per-attempt
  schema failures are still recorded --- strong outcome, measurable reliability deficit.
\item
  \textbf{deepseek-chat-v3.1 · autonomous} inverts the pattern: its Reliability is excellent under the
  scaffold (0.99 in structured mode) but falls to 0.62 in autonomous mode. The \emph{same model} is reliable
  with the staged hand-off and substantially less reliable without it.
\end{itemize}

By contrast gpt-5-mini and claude-haiku-4.5 are schema-clean in both modes (Reliability ≈ 1.0).
So decision-schema adherence is a real, model × mode-specific capability limit --- the smaller models cannot reliably emit the structured decision --- and one that a returns-only view misses entirely: gemini·structured beats \textasciitilde6/8 baselines \emph{while} failing the schema.

\begin{figure}[tbp]\centering
\includegraphics[width=0.82\linewidth,height=\textheight,keepaspectratio]{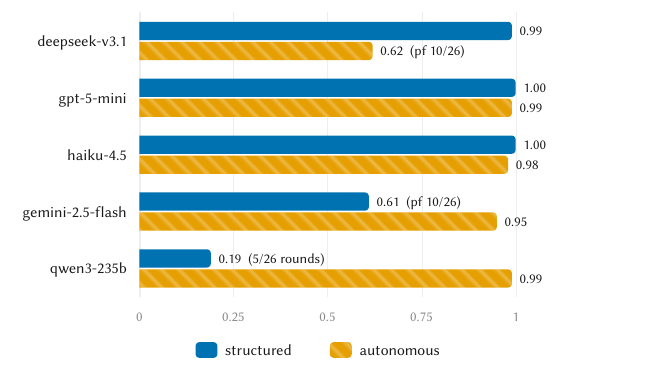}

\vspace{7pt}
\begin{minipage}{0.86\textwidth}\small \textbf{Figure 8.}\textit{ Operational reliability (D5) by configuration; ``pf'' labels mark mean parse-fail rounds,
the mechanical driver of the D5 spread. qwen3·structured and gemini·structured carry the cohort's
reliability deficits, both invisible to their returns; deepseek's reliability flips with mode. qwen3's
incompletion is the separate tool-loop non-termination failure (§7.5), not a parse-fail.}\end{minipage}
\end{figure}

\subsubsection{7.6 RQ4: Scaffold Shapes the Capability Profile}\label{rq4-scaffold-shapes-the-capability-profile}

\textbf{The composite favors structure for the reliable backtest models, but the advantage is partly
definitional.} The full cohort-relative APM-CS is higher for the four reliable structured backtest
configurations, most dramatically for deepseek (72.9 vs 35.1), followed by haiku (71.0 vs 40.4),
gpt-5-mini (67.4 vs 50.3), and gemini (47.4 vs 32.3). These cross-mode contrasts are descriptive:
D2 and D4 are mode-confounded (§6.4), and qwen3's structured row (30.7) rests on only five reliable
rounds. The sensitivity analysis restricted to the mode-comparable axes (the D1/D3/D5 mean,
Appendix F.4) shows that the composite's structured advantage depends substantially
on the confounded axes. On the
comparable axes, structure is favored only for deepseek in the backtest cohort, and all six live pairs
favor autonomy (§7.9). Committee scaffolding is not universally beneficial.

The clearer axis-level result is that structure broadly lifts Composure (D3): structured configs hold
one-way turnover flat in volatility (D3 ≈ 0.92--0.95), and the autonomous configs sit slightly lower
(D3 ≈ 0.77--0.92). The lowest is gemini·autonomous at 0.77 --- the only configuration with a positive
volatility-overreaction slope (the others' slopes are ≈0). In this configuration, full autonomy coincides with
greater volatility-sensitive turnover, a stability difference hidden by headline Sharpe.

\textbf{On Coherence.} Signal-action agreement is moderate (\textasciitilde0.53--0.66) for every configuration but
haiku·structured, yet the reasoning-enabled minimax-m3 judge scores the allocations' coherence \emph{low}.
As a result, D1 sits in a 0.38--0.47 band for most configs. The outlier is claude-haiku·structured
at 0.23: it posts the cohort's \emph{best reliable Sharpe (3.21)} and the best structured-mode Acuity (0.44),
yet its allocations are the least consistent with its own stated analysis.
The two Coherence components disagree: agents trade in broadly the right direction on average, but
their narratives do not justify the \emph{specific} weights. That disagreement is the diagnostic,
visible only because Coherence pairs a mechanical agreement check with a separately constructed,
reasoning-capable judge. Either alone would mislead. §7.9 quantifies the divergence across both tracks.

\subsubsection{7.7 RQ5--RQ6: Ablation Grid (single- and multi-module)}\label{rq5rq6-ablation-grid-single--and-multi-module}

The grid first asks whether removing a module moves Sharpe beyond run-to-run noise and answers no
for every module and cluster (§7.7.1). §7.7.2 turns to the axes that \emph{do} register removal
costs, and §7.7.3 to the regime condition that bounds every returns-based reading.

\paragraph{7.7.1 Single- and Multi-Module Knockouts: Returns Cannot Rank Module Value}\label{single--and-multi-module-knockouts-returns-cannot-rank-module-value}

The separability bar comes from the data itself: a \textbf{±0.42-Sharpe repeated-run noise band} (±1σ of
run-to-run Sharpe across the anchor configuration's repeated runs) bounds what a single ΔSharpe can
claim. Repeated runs move Sharpe by up to \textasciitilde2.3 and the composite by up to \textasciitilde25 APM-CS points across
configurations. A variant is therefore \emph{separable} only if its whole run range clears the band.
Table 6 reads every module-off Δ against that bar.

\begin{center}\begin{minipage}{0.86\linewidth}\small
\textbf{Table 6.}\textit{ Ablation ΔSharpe vs.~the structured-gemini baseline (Sharpe 2.26, the ablation anchor's
baseline; the same configuration's five-seed campaign mean is 2.31, Table 5; bi-weekly, LOW tier).
Six module-off variants as repeated runs, the two cost-tier variants single-run; every run completed
26/26 rounds, chains verified, 0 failures.}
\end{minipage}\end{center}

{\def\LTcaptype{none} 
\begin{longtable}[]{@{}
  >{\raggedright\arraybackslash}p{(\linewidth - 6\tabcolsep) * \real{0.2093}}
  >{\raggedright\arraybackslash}p{(\linewidth - 6\tabcolsep) * \real{0.1744}}
  >{\raggedright\arraybackslash}p{(\linewidth - 6\tabcolsep) * \real{0.1744}}
  >{\raggedright\arraybackslash}p{(\linewidth - 6\tabcolsep) * \real{0.4419}}@{}}
\toprule\noalign{}
\begin{minipage}[b]{\linewidth}\raggedright
Variant
\end{minipage} & \begin{minipage}[b]{\linewidth}\raggedright
ΔSharpe (mean)
\end{minipage} & \begin{minipage}[b]{\linewidth}\raggedright
run range
\end{minipage} & \begin{minipage}[b]{\linewidth}\raggedright
Reading
\end{minipage} \\
\midrule\noalign{}
\endhead
\bottomrule\noalign{}
\endlastfoot
cost-high & \textbf{−3.32} & --- & HIGH tier \textbf{mis-calibrated} (≈527 bps/\allowbreak{}rd → −75\% cum,\allowbreak{} Sharpe −1.06); the one \textbf{separable} signal \\
cost-medium & −0.19 & --- & Δcost tiny → decision noise \\
no-reflection & +0.09 & {[}−0.16,\allowbreak{} +0.39{]} & within the noise band \\
no-news-sentiment & +0.12 & {[}−0.34,\allowbreak{} +0.59{]} & within the noise band \\
no-advanced-quant & +0.16 & {[}−0.06,\allowbreak{} +0.32{]} & within the noise band \\
no-cost-modeling & +0.37 & {[}+0.36,\allowbreak{} +0.39{]} & tight,\allowbreak{} but still inside ±0.42 \\
no-memory & +0.52 & {[}+0.24,\allowbreak{} +0.72{]} & straddles band on Sharpe --- \textbf{but drops Composure's consistency leg} (below) \\
no-world-macro & +0.66 & {[}+0.33,\allowbreak{} +1.06{]} & largest mean,\allowbreak{} but range dips into the band → a favorable draw,\allowbreak{} \textbf{not} ``macro helps'' \\
\end{longtable}
}

\begin{figure}[tbp]\centering
\includegraphics[width=0.9\linewidth,keepaspectratio]{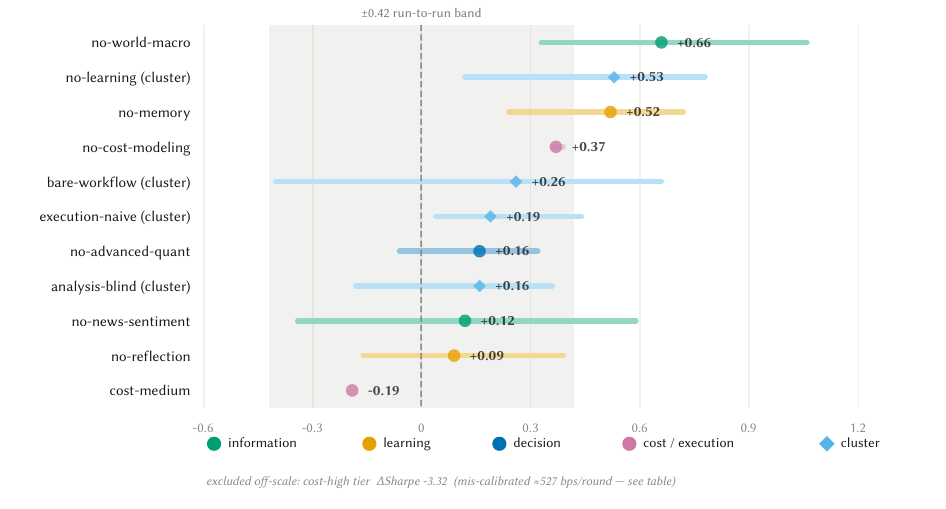}

\vspace{7pt}
\begin{minipage}{0.86\textwidth}\small \textbf{Figure 9.}\textit{ Ablation ΔSharpe (mean ± range across repeated runs) vs.~the structured-gemini
baseline, single-module \emph{and} multi-module (hatched) knockouts. Only the mis-calibrated HIGH cost
tier clears the ±0.42 repeated-run noise band; no single-module or cluster removal --- down to the
bare-workflow extreme --- separates on returns.}\end{minipage}
\end{figure}

\emph{Module importance is not rankable by Sharpe --- even across repeated runs (RQ5).} Every module-off
ΔSharpe is \emph{positive} (means +0.09\ldots+0.66): removal never cleanly \emph{hurts}, which is mechanistically
implausible. Every run range straddles or sits inside the ±0.42 band; extra runs \emph{narrowed}
the estimates without changing the verdict. The largest mover, \emph{no-world-macro}, is a
\emph{favorable draw}, not ``macro helps'' (Table 6). Returns variance dominates on this fixed market
path; the design does not separate smaller effects on returns.

\emph{Cost realism (RQ6).} The HIGH tier (Almgren--Chriss; Almgren and Chriss 2001) is mis-calibrated:
≈527 bps/round (\textasciitilde5.3\%/round → −75\% cumulative, Sharpe −1.06) --- an implausible drag that confirms
keeping LOW and flags a recalibration task.

\emph{The capability scorecard catches what Sharpe cannot.} Removing the information stack leaves Sharpe
in the noise yet collapses Acuity to ≈−0.01 (the analysis-blind cluster, below): no informative
signals remain to attend to --- a capability loss invisible to returns. Removing memory strips
Composure's strategy-\emph{consistency} leg (the \emph{consistency\_score} only the memory module produces).
The axis falls back to its volatility-overreaction leg (§6.4), which sat below the calm leg in this
benign window. The composite therefore does not fall --- the signal is the \emph{absent leg} in the trail.

\emph{Multi-module (interaction) ablations.} Modules may be individually redundant yet collectively
essential, so whole capability clusters are knocked out across repeated runs (Table 7):

{\def\LTcaptype{none} 
\begin{longtable}[]{@{}
  >{\raggedright\arraybackslash}p{(\linewidth - 10\tabcolsep) * \real{0.1744}}
  >{\raggedright\arraybackslash}p{(\linewidth - 10\tabcolsep) * \real{0.1744}}
  >{\raggedright\arraybackslash}p{(\linewidth - 10\tabcolsep) * \real{0.1744}}
  >{\raggedright\arraybackslash}p{(\linewidth - 10\tabcolsep) * \real{0.0814}}
  >{\raggedright\arraybackslash}p{(\linewidth - 10\tabcolsep) * \real{0.1395}}
  >{\raggedright\arraybackslash}p{(\linewidth - 10\tabcolsep) * \real{0.2558}}@{}}
\toprule\noalign{}
\begin{minipage}[b]{\linewidth}\raggedright
Cluster knocked out (modules off)
\end{minipage} & \begin{minipage}[b]{\linewidth}\raggedright
ΔSharpe (mean)
\end{minipage} & \begin{minipage}[b]{\linewidth}\raggedright
run range
\end{minipage} & \begin{minipage}[b]{\linewidth}\raggedright
CumRet
\end{minipage} & \begin{minipage}[b]{\linewidth}\raggedright
Acuity (D2)
\end{minipage} & \begin{minipage}[b]{\linewidth}\raggedright
Reading
\end{minipage} \\
\midrule\noalign{}
\endhead
\bottomrule\noalign{}
\endlastfoot
\textbf{analysis-blind} (news + macro + advanced-quant) & +0.16 & {[}−0.18,\allowbreak{} +0.36{]} & +27.1\% & \textbf{−0.01} & within band on returns; \textbf{Acuity collapses} (no signal sources to attend to) \\
\textbf{no-learning} (memory + reflection) & +0.53 & {[}+0.12,\allowbreak{} +0.78{]} & +29.6\% & −0.01 & within band; \textbf{Composure loses its consistency leg} (memory off) \\
\textbf{execution-naive} (cost-model + systematic-exec) & +0.19 & {[}+0.04,\allowbreak{} +0.44{]} & +23.3\% & 0.12 & within band; turnover unchanged (\textasciitilde0.18) \\
\textbf{bare-workflow} (only mandate + cost-aware execution) & +0.26 & {[}−0.40,\allowbreak{} +0.66{]} & +27.3\% & 0.22 & within band; \textbf{the bare-workflow agent matches the full agent on returns} \\
\end{longtable}
}

\begin{center}\begin{minipage}{0.86\linewidth}\small
\textbf{Table 7.}\textit{ Cluster ablation: ΔSharpe vs.~the structured-gemini baseline (Sharpe 2.26) with an
entire capability cluster removed (12 runs, all 26/26 rounds, same gemini anchor). \emph{Run range} =
min--max across repeated runs; \emph{Reading} names the axis most implicated. All four clusters stay
inside the ±0.42 band on returns while the capability axes register what returns hide
(analysis-blind → Acuity −0.01, zero signal discrimination; two-leg noise composite, §6.4).}
\end{minipage}\end{center}

\emph{Combined removal does not separate on returns either; bare-workflow is the striking case.}
Bare-workflow has \emph{no} information modules, \emph{no} learning, and no quant-optimization engine.
The LLM sets weights itself under the mandate, with a risk-parity execution fallback. This variant
is not separable from the fully-augmented six-module agent (ΔSharpe +0.26, range {[}−0.40, +0.66{]}).
On \emph{this} benchmark and period, the agentic scaffolding contributes little to \emph{Sharpe}, while the
axes register what returns hide (Table 7). Bare-workflow's \emph{higher} Acuity point estimate arises
from how ``proposed'' is measured without the optimizer, not from superior signal attention.
\emph{Module value cannot be read off returns ---
single-module, cluster-level, or against a stripped-to-the-mandate baseline --- but the multi-axis
scorecard records what each removal costs.}

\paragraph{7.7.2 Looking Past Sharpe: Risk, Cost, and Decision Behavior}\label{looking-past-sharpe-risk-cost-and-decision-behavior}

A returns-only read would call all
thirteen configurations equivalent when they differ sharply on \emph{risk}, \emph{cost}, and \emph{decision
behavior}.

\emph{The full performance profile (Table 8) --- three facts Sharpe hides.} (a) The full agent halves the
index's drawdown at the same Sharpe --- the deliberate de-risking trade examined under RQ7 in §7.8.
(b) That risk discipline is robust to ablation --- every module-ablation config holds drawdown to
−3.2\ldots−4.7\% and Calmar 5.8--8.7 (cost tiers aside), with zero constraint-violations and style-drift
across all thirteen. This is consistent with risk discipline residing in the \emph{constrained
closed-loop} (caps + staged Risk-Officer/MAPC) rather than the optional modules. The attribution
is by elimination, since the constraint layer is not ablated alone. (c) The one module with a clean PnL footprint is
cost-awareness: removing it raises realized cost +10\% (the execution pair +18\%); the cost \emph{tier}
matters where modules do not (HIGH is 122× the LOW cost, vol 71\%, DD −76\%).

\begin{center}\begin{minipage}{0.86\linewidth}\small
\textbf{Table 8.}\textit{ Multi-metric ablation profile (means over the ablation campaign's repeated runs;
\emph{full} = the structured-gemini anchor, so values differ slightly from the same configuration's
campaign-seed means in Table 5; SPY shown for the risk-adjusted comparison). Combos marked ⁺.}
\end{minipage}\end{center}

{\def\LTcaptype{none} 
\begin{longtable}[]{@{}
  >{\raggedright\arraybackslash}p{(\linewidth - 18\tabcolsep) * \real{0.2568}}
  >{\raggedright\arraybackslash}p{(\linewidth - 18\tabcolsep) * \real{0.0811}}
  >{\raggedright\arraybackslash}p{(\linewidth - 18\tabcolsep) * \real{0.0946}}
  >{\raggedright\arraybackslash}p{(\linewidth - 18\tabcolsep) * \real{0.0811}}
  >{\raggedright\arraybackslash}p{(\linewidth - 18\tabcolsep) * \real{0.0541}}
  >{\raggedright\arraybackslash}p{(\linewidth - 18\tabcolsep) * \real{0.0811}}
  >{\raggedright\arraybackslash}p{(\linewidth - 18\tabcolsep) * \real{0.0811}}
  >{\raggedright\arraybackslash}p{(\linewidth - 18\tabcolsep) * \real{0.1216}}
  >{\raggedright\arraybackslash}p{(\linewidth - 18\tabcolsep) * \real{0.0946}}
  >{\raggedright\arraybackslash}p{(\linewidth - 18\tabcolsep) * \real{0.0541}}@{}}
\toprule\noalign{}
\begin{minipage}[b]{\linewidth}\raggedright
Config
\end{minipage} & \begin{minipage}[b]{\linewidth}\raggedright
Sharpe
\end{minipage} & \begin{minipage}[b]{\linewidth}\raggedright
Sortino
\end{minipage} & \begin{minipage}[b]{\linewidth}\raggedright
Calmar
\end{minipage} & \begin{minipage}[b]{\linewidth}\raggedright
Vol\%
\end{minipage} & \begin{minipage}[b]{\linewidth}\raggedright
MaxDD\%
\end{minipage} & \begin{minipage}[b]{\linewidth}\raggedright
Beat/\allowbreak{}8
\end{minipage} & \begin{minipage}[b]{\linewidth}\raggedright
IR vs SPY
\end{minipage} & \begin{minipage}[b]{\linewidth}\raggedright
Cost \$
\end{minipage} & \begin{minipage}[b]{\linewidth}\raggedright
Turn
\end{minipage} \\
\midrule\noalign{}
\endhead
\bottomrule\noalign{}
\endlastfoot
\emph{SPY (cap-wtd index)} & 2.29 & --- & 3.1 & 11.9 & \textbf{−8.9} & --- & --- & --- & --- \\
\textbf{full (all modules)} & 2.26 & 4.88 & 7.03 & 11.0 & \textbf{−4.1} & 5.7 & −0.04 & 3 353 & 0.15 \\
no-news-sentiment & 2.38 & 5.31 & 5.91 & 11.0 & −4.4 & 6.7 & −0.01 & 3 338 & 0.15 \\
no-advanced-quant & 2.42 & 5.82 & 8.19 & 10.5 & −3.3 & 6.3 & −0.03 & 3 788 & 0.17 \\
no-memory & 2.78 & 5.33 & 7.00 & 9.0 & −3.6 & 6.7 & −0.04 & 3 657 & 0.16 \\
analysis-blind ⁺ & 2.42 & 4.81 & 5.82 & 11.2 & −4.7 & 7.0 & −0.00 & 4 111 & 0.18 \\
bare-workflow ⁺ & 2.52 & 5.48 & 7.16 & 10.9 & −4.2 & 7.3 & −0.00 & 3 983 & 0.17 \\
cost-medium & 2.07 & 4.85 & 4.57 & 13.3 & −6.0 & 7.0 & +0.01 & 4 787 & 0.16 \\
cost-high & \textbf{−1.06} & −1.03 & −0.99 & \textbf{70.9} & \textbf{−75.9} & 0.0 & −0.33 & \textbf{411 154} & 0.14 \\
\end{longtable}
}

\emph{Behavioral diagnostics --- detaching modules degrades the }process\emph{, not the outcome (Fig. 10).} The
bare-workflow agent does half the analysis, weighs half the signals, scouts no candidates, performs
no reflection, and spends 40\% less reasoning effort --- yet proposes the \emph{largest} trades. Where the
full agent records a mean-variance optimization with explicit weights, exclusions, and a
cost-management lesson, the bare-workflow agent's \emph{entire} rationale is a one-line optimizer note (Appendix E). None of this shows in returns: concentration stays flat across all thirteen
configurations, so the rising presumptuousness surfaces only in \emph{proposed} moves the caps clip ---
the axes record what the constraints mask.

\begin{figure}[tbp]\centering
\includegraphics[width=0.8\linewidth,height=\textheight,keepaspectratio]{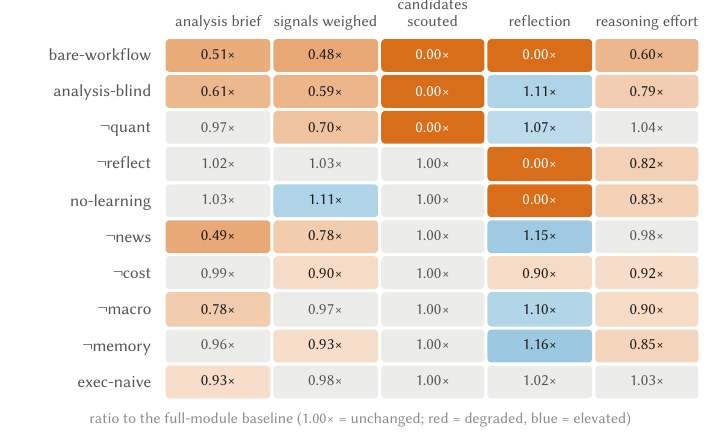}

\vspace{7pt}
\begin{minipage}{0.86\textwidth}\small \textbf{Figure 10.}\textit{ Decision-quality proxies per ablation as a fraction of the full-module baseline (1.0 =
unchanged; lower = more degraded). Stripping modules collapses analysis depth, signal breadth,
candidate exploration, self-scrutiny and reasoning effort --- bare-workflow is degraded on every
axis --- while returns are unaffected because the mandate's constraints contain the degraded decisions
(Table 8).}\end{minipage}
\end{figure}

\emph{Module feasibility for a professional agent-investing platform.} Synthesising the
returns/risk/cost/capability/behavioral evidence yields an actionable per-module verdict (Table 9),
not a returns rank.

\begin{center}\begin{minipage}{0.86\linewidth}\small
\textbf{Table 9.}\textit{ Per-module feasibility, grounded in the ablation evidence.}
\end{minipage}\end{center}

{\def\LTcaptype{none} 
\begin{longtable}[]{@{}
  >{\raggedright\arraybackslash}p{(\linewidth - 4\tabcolsep) * \real{0.2093}}
  >{\raggedright\arraybackslash}p{(\linewidth - 4\tabcolsep) * \real{0.4419}}
  >{\raggedright\arraybackslash}p{(\linewidth - 4\tabcolsep) * \real{0.3488}}@{}}
\toprule\noalign{}
\begin{minipage}[b]{\linewidth}\raggedright
Module
\end{minipage} & \begin{minipage}[b]{\linewidth}\raggedright
Measured effect of removal
\end{minipage} & \begin{minipage}[b]{\linewidth}\raggedright
Verdict (production platform)
\end{minipage} \\
\midrule\noalign{}
\endhead
\bottomrule\noalign{}
\endlastfoot
\emph{cost-modeling} & returns/\allowbreak{}risk-neutral; \textbf{+10\% realized friction}; HIGH tier −76\% (mis-cal) & \textbf{Keep} --- cheap,\allowbreak{} cost realism; \emph{calibrate the tier} (HIGH is broken) \\
\emph{systematic-execution} & neutral returns; turnover/\allowbreak{}cost ↑ when paired-off & \textbf{Keep} --- execution discipline \\
\emph{investment-target} + constraints & (not ablated alone) inferred source of the \textbf{−4\% vs −9\% drawdown edge}; contains overconfidence & \textbf{Keep --- essential}; the real risk-adjusted edge + the safeguard \\
\emph{memory} & returns-neutral; \textbf{zeroes lessons; strips Composure's strategy-consistency leg (D3,\allowbreak{} §6.4)} & \textbf{Keep} --- the only source of the strategy-consistency signal \\
\emph{reflection} & returns-neutral; \textbf{zeroes self-scrutiny/\allowbreak{}learning} & \textbf{Conditional} --- keep for auditability/\allowbreak{}improvement; deferrable for raw returns \\
\emph{advanced-quant} & returns-neutral (risk-parity fallback suffices); \textbf{zeroes candidate exploration,\allowbreak{} shallows analysis} & \textbf{Conditional} --- keep for breadth/\allowbreak{}optimization fidelity \\
\emph{news-sentiment},\allowbreak{} \emph{world-macro},\allowbreak{} \emph{asset-class-data} & returns-neutral; \textbf{feed Acuity (analysis-blind → −0.01) and analysis depth} (¬news halves the brief) & \textbf{Optional} in benign regimes; \textbf{keep for breadth/\allowbreak{}stress} \\
\end{longtable}
}

\paragraph{7.7.3 The Regime Does Much of the Work}\label{the-regime-does-much-of-the-work}

\emph{One benign regime masks both module value and model risk-taking.} Everything above is measured in
one benign, rising market: SPY +27\% over the campaign, a −8.9\% drawdown, calm VIX (median 17; above
25 on just 7\% of days). Three consequences follow, each a reason the ablation Δ are small.

First, the market --- not the modules --- drives the returns: in a trending bull almost any reasonable
long-only book posts a strong Sharpe and clears the defensive baselines, so removing a module
\emph{cannot} move returns much.

Second, the modules whose value is most regime-contingent are exactly the ones this period mutes.
The information stack (\emph{news-sentiment} / \emph{world-macro} / \emph{asset-class-data}) earns its keep at
\emph{turning points}, not riding a trend. The learning loop (\emph{memory} / \emph{reflection}) pays off \emph{across}
regimes, which a single stable trend barely exercises. The covariance-aware optimizer
(\emph{advanced-quant}) matters most when correlations spike in a sell-off. Their flat Δ \emph{here} is weak
evidence of irrelevance and strong evidence this regime cannot test them.

Third --- and most consequential --- a rising market masks risk-taking shortcomings the scorecard
catches but returns do not: the configurations that look best on raw return take the \emph{most} risk,
while the prudent de-risking the agents actually exhibit (half the index's drawdown, Table 8)
\emph{costs} return. A returns leaderboard in this regime rewards the recklessness a drawdown would
punish and penalizes the risk management that would protect it --- qwen3·structured's 5-round Sharpe
(§7.3) is the extreme case. The benign window thus limits the power of the \emph{ablations}
(its stress bucket spans only ≈ 4--5 of 26 rounds). It also makes the capability and risk-adjusted
axes structurally necessary: they stay diagnostic when the regime makes returns uninformative,
or worse, misleading.

\subsubsection{7.8 RQ7: Net of Costs, Agents Clear the Defensive Baselines but Not the Index Cleanly}\label{rq7-net-of-costs-agents-clear-the-defensive-baselines-but-not-the-index-cleanly}

Over the full bi-weekly year, the reliable structured configurations clear a median \textasciitilde6 of 8
passive baselines (deepseek 6.8, haiku 6.0, gemini 6.0; gpt-5-mini, at 4.6, is the below-median case;
Figure 11). But the beaten set is dominated by the suite's
defensive / diversified members (IEF bonds, risk-parity, 60/40, equal-weight); the cap-weighted
equity index remains the hardest bar. The autonomous configurations clear far fewer baselines:
qwen3-235b autonomous is lowest at 2.2/8 on seed means, and gemini autonomous clears 3.0/8.
Because the instrument scores
every configuration against the \emph{whole} suite, not a hand-picked benchmark, the supportable statement
is bounded: under realistic costs these agents reliably beat the defensive baselines but do not
uniformly clear the cap-weighted index, and autonomous mode markedly weakens even that.
(qwen3·structured's 8/8 is excluded as a 5-round artifact, §7.3.)

\textbf{The index gap is a de-risking trade, not a deficit.} ``Does not beat the cap-weighted index'' understates
what the risk-adjusted read shows in Table 8: against SPY the agents run lower volatility and roughly
\emph{half} the maximum drawdown (−4.1\% vs −8.9\%) at the same Sharpe and ≈2× the Calmar. The shortfall in
\emph{absolute} return is the price of deliberate diversification away from the index's mega-cap concentration ---
a trade that \emph{loses} a little in a concentrated bull run and \emph{wins} decisively in a drawdown. The
ablation diagnostics show that this risk discipline is intrinsic to the constrained construction and
persists even when the workflow is stripped to its bare form. The Composure axis (§7.6) measures
exactly this behavior within the instrument; a returns leaderboard, fixated on the −2.4\% vs SPY, would miss it.

\begin{figure}[tbp]\centering
\includegraphics[width=0.82\linewidth,height=\textheight,keepaspectratio]{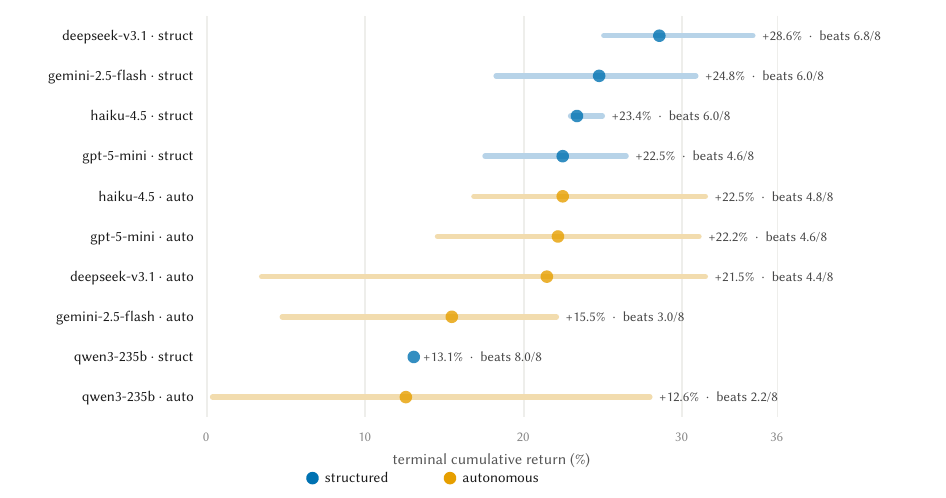}

\vspace{7pt}
\begin{minipage}{0.86\textwidth}\small \textbf{Figure 11.}\textit{ Terminal cumulative return by configuration over the 26 bi-weekly rounds
(2025-06 → 2026-06); whiskers span the range across repeated runs, and each label
gives the number of the 8 passive baselines the configuration beats. Most structured configs sit above
their autonomous counterparts on terminal return and clear the defensive baselines (≈6/8), while
autonomous configs trail (qwen3-235b autonomous the lowest at 2.2/8 on seed means).}\end{minipage}
\end{figure}

\subsubsection{7.9 Live-Track Decision-Level Validation (Completed Four-Week Campaign, 18 Valid Days)}\label{live-track-decision-level-validation-completed-four-week-campaign-18-valid-days}

The live track asks one question: \textbf{does the instrument produce coherent, interpretable capability
diagnostics on data the models never saw?} It does: the agreement↔judge coherence gap reproduces
at +0.23 (vs +0.30 backtest, below); as in backtest, no configuration dominates all five axes; and \textbf{every live
model's five-axis composite is higher in autonomous mode than in structured mode} (Table 10).

\textbf{Setup.} The four-week daily-rebalance campaign on a shared broker paper account is complete, with
\textbf{eighteen valid NYSE days} (2026-06-15 → 2026-07-10). The campaign ran 6 newer ``thinking'' models
(minimax-m3, deepseek-v4-pro, qwen3.7-plus, gpt-5.4-mini, gemini-3.5-flash, claude-sonnet-4.6)
in both structured and autonomous modes, yielding 12 concurrent virtual-ledger sessions, each a
\$1M slice at medium reasoning. 2026-06-19 (Juneteenth) and the July-3 market holiday are dropped via a published exclusion
manifest enforced by an exchange-calendar gate. Each ledger opens from an equal-weight core book at
the prior close, only the daily delta reaches the broker, and an end-of-day two-layer flatten
returns the account to flat. Because the live cohort is a \emph{newer generation} than the backtest
cohort and realized P\&L is a sequence of one-day deltas, live returns are indicative-only and
evaluation is at the \emph{decision level}.

\textbf{Campaign integrity ledger.} Three operational events are marked in the data rather than silently
absorbed. \emph{(i) Mid-campaign remediation boundary} (the 2026-07-02 session --- ledger round 14, the
13th valid trading day): a prompt-layer and failure-handling audit led to three changes.
The autonomous path gained output-schema coercion and fail-loud decision retries (a failed decision
previously degraded silently into a hold, §7.10); the Risk Officer's prompt gained the mandate's
actual constraint numbers; and the memory loop was closed with bounded lesson/history feedback. Rounds 1--13
ran on the earlier prompts, so the pooled eighteen-round scorecard is the primary read; the one
metric the boundary visibly moves --- the autonomous hold rate --- is split in §7.10, where the
boundary doubles as an intervention test. \emph{(ii) Broker-gateway outage} (2026-07-06): contract-id
resolution broke at EXECUTE for the six autonomous sessions \emph{after} their decisions were recorded;
EXECUTE alone was re-run the same day against the unchanged recorded targets --- decision-level
scoring is unaffected by construction. \emph{(iii) Per-model reasoning deviation} (final round):
qwen3.7-plus's structured PM entered a non-terminating reasoning/tool loop at medium effort (single
calls of 200--770 s, reproducible in isolation), echoing the qwen3-235b non-termination (§7.5) in a
second qwen generation. Its final round ran in both modes with reasoning disabled. All twelve
ledgers verify end-to-end (19 recorded rounds each --- 18 valid days plus the excluded Juneteenth
round; hash chains PASS).

\textbf{Decision-level scorecard (live).} Computed over the eighteen valid rounds, the five axes
separate the live cohort along the same dimensions as backtest (Table 10). claude-sonnet-4.6 leads
(autonomous APM-CS 83.6); deepseek-v4-pro autonomous carries the cohort's best judge leg (0.67).
The uniform
autonomous-over-structured split held at nine, fourteen, and eighteen rounds and persists on the
mode-comparable D1/D3/D5 sensitivity (Appendix F.4). The floor scores are particularly
informative: minimax-m3 on Reliability (parse failures in both modes, 9 and 7 of 18 rounds), and
gemini and minimax on Composure in both modes. Composure is near-bimodal. The sharpest horizon effect:
gemini-3.5-flash·structured fell from mid-pack at nine days to the cohort floor at fourteen,
staying there as its Composure collapsed, while deepseek's and gpt's autonomous Composure rose
through the final week --- the shortest-horizon axes keep moving even at four weeks. §7.10 dissects
the autonomous hold behavior behind minimax's reliability hit and gemini-3.5-flash's distinct hold
mechanism, both eliminated after the round-14 remediation.

\begin{center}\begin{minipage}{0.86\linewidth}\small
\textbf{Table 10.}\textit{ Live decision-level scorecard (18 valid NYSE days, 2026-06-15 → 07-10, the completed
campaign; Coherence = ½·signal--action agreement + ½·held-out GLM-5.2 judge, the live primary per
§6.4.1; D2 uses the two-leg noise composite; APM-CS = mean cohort percentile across the five
axes). Returns are indicative-only and omitted. Sorted by APM-CS; per-axis leader bold.}
\end{minipage}\end{center}

{\def\LTcaptype{none} 
\begin{longtable}[]{@{}
  >{\raggedright\arraybackslash}p{(\linewidth - 12\tabcolsep) * \real{0.3023}}
  >{\raggedright\arraybackslash}p{(\linewidth - 12\tabcolsep) * \real{0.1163}}
  >{\raggedright\arraybackslash}p{(\linewidth - 12\tabcolsep) * \real{0.1163}}
  >{\raggedright\arraybackslash}p{(\linewidth - 12\tabcolsep) * \real{0.1279}}
  >{\raggedright\arraybackslash}p{(\linewidth - 12\tabcolsep) * \real{0.1279}}
  >{\raggedright\arraybackslash}p{(\linewidth - 12\tabcolsep) * \real{0.1279}}
  >{\raggedright\arraybackslash}p{(\linewidth - 12\tabcolsep) * \real{0.0814}}@{}}
\toprule\noalign{}
\begin{minipage}[b]{\linewidth}\raggedright
Model · mode
\end{minipage} & \begin{minipage}[b]{\linewidth}\raggedright
D1 Coher.
\end{minipage} & \begin{minipage}[b]{\linewidth}\raggedright
D2 Acuity
\end{minipage} & \begin{minipage}[b]{\linewidth}\raggedright
D3 Compos.
\end{minipage} & \begin{minipage}[b]{\linewidth}\raggedright
D4 Discip.
\end{minipage} & \begin{minipage}[b]{\linewidth}\raggedright
D5 Reliab.
\end{minipage} & \begin{minipage}[b]{\linewidth}\raggedright
\textbf{APM-CS}
\end{minipage} \\
\midrule\noalign{}
\endhead
\bottomrule\noalign{}
\endlastfoot
claude-sonnet-4.6 · auto & 0.60 & \textbf{0.30} & 0.94 & 0.80 & \textbf{1.00} & \textbf{83.6} \\
qwen3.7-plus · auto & 0.59 & 0.26 & 0.94 & 0.75 & \textbf{1.00} & 69.1 \\
deepseek-v4-pro · auto & \textbf{0.62} & 0.14 & \textbf{0.95} & 0.62 & \textbf{1.00} & 63.6 \\
gpt-5.4-mini · auto & 0.56 & 0.27 & 0.69 & 0.83 & 0.99 & 63.6 \\
claude-sonnet-4.6 · stru & 0.52 & 0.14 & 0.93 & 0.95 & 0.82 & 52.7 \\
deepseek-v4-pro · stru & 0.50 & 0.09 & \textbf{0.95} & 0.68 & \textbf{1.00} & 50.9 \\
gemini-3.5-flash · auto & 0.51 & 0.29 & 0.44 & 0.65 & \textbf{1.00} & 43.6 \\
minimax-m3 · auto & 0.56 & 0.25 & 0.45 & 0.91 & 0.49 & 41.8 \\
qwen3.7-plus · stru & 0.41 & −0.10 & \textbf{0.95} & 0.95 & 0.58 & 40.0 \\
gpt-5.4-mini · stru & 0.48 & 0.10 & 0.49 & 0.78 & 0.97 & 36.4 \\
minimax-m3 · stru & 0.41 & 0.00 & 0.47 & \textbf{0.97} & 0.56 & 29.1 \\
gemini-3.5-flash · stru & 0.42 & 0.18 & 0.46 & 0.58 & 0.94 & 25.5 \\
\end{longtable}
}

\textbf{The cross-track invariant (headline live result).} The mechanical signal--action \emph{agreement} and
the held-out reasoning \emph{judge} --- the two components of Coherence --- diverge systematically on both
tracks: \textbf{+0.30 backtest} (agreement 0.58 / minimax-m3 judge 0.28), \textbf{+0.23 live} (agreement
0.63 / GLM-5.2 judge 0.40; Figure 12). Each track has a different primary (§6.4.1), so the cleanest
comparisons are \emph{within} judge. The structure is the same: every protocol-clean gap falls in a narrow
\textbf{+0.23\ldots+0.30 band} across two judges, two cohorts, two periods, and two cadences (one superseded
read of +0.34, under the pre-hold-out protocol of Appendix C, is excluded). Nor is
the divergence an artifact of weak models: newer live models lift both components in step
(deepseek's judge leg rises 0.37 → 0.67 from backtest-era v3.1 to live v4-pro), yet the
cohort-level \emph{difference} persists. Agents keep \emph{agreeing with signals mechanically} while failing
to \emph{cohere with their own stated reasoning}, on data they never saw.

\begin{figure}[tbp]\centering
\includegraphics[width=0.62\linewidth,height=\textheight,keepaspectratio]{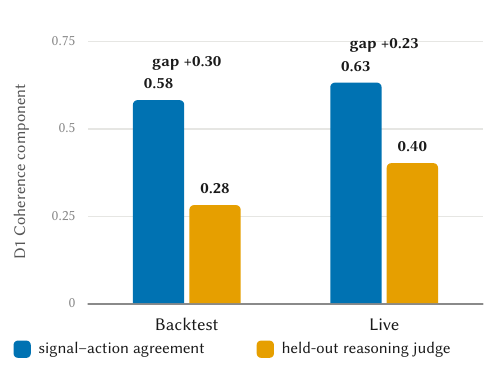}

\vspace{7pt}
\begin{minipage}{0.86\textwidth}\small \textbf{Figure 12.}\textit{ The agreement↔judge coherence gap: +0.30 (backtest, minimax-m3 judge), +0.23 (live,
GLM-5.2 judge, eighteen valid rounds); protocol-clean within-judge comparisons stay in the
+0.23\ldots+0.30 band.}\end{minipage}
\end{figure}

\textbf{What the gap measures, and why coherence scores low.} \emph{Agreement}, the easier half, checks only
the direction of each proposed trade against the sign of that name's informative composite --- a low
bar: a coin-flip clears 0.5, modest skill lifts to \textasciitilde0.6. The \emph{judge} asks whether the specific weights are a coherent \emph{consequence} of
the agent's own stated analysis --- and the cohort fails it: the year-long backtest campaign's
held-out judge mean is 2.12 of 5 (normalized 0.28); the newer live cohort reads higher under
GLM-5.2 (0.40) while still trailing its own agreement leg. The \textbf{two-leg divergence} is treated
descriptively --- the components are bounded to the same interval but differ in baseline and
difficulty, so their numerical difference is not a calibrated effect size; its stability is the
empirical result. The judge's cached rationales make the low scores checkable. Five patterns recur:
analysis--trade contradiction, equal-weight collapse ignoring the
differentiation the analysis emphasized, mismatch between the names in the thesis and in the book,
weights anti-correlated with the cited signals, and no thesis at all (verbatim, round-pinned
examples in Appendix E).

\textbf{Where the incoherence enters.} Part is the agent's narrative not matching even its own intent;
part is the \emph{machinery}. The judge scores the post-construction target, already reshaped by the
quant optimizer and the constrained projection. Equal-weight collapse (the optimizer flattening a
tactical tilt) and cap-hits (the projection clipping) are partly system effects. The instrument
records \emph{proposed} (pre-MAPC intent) and \emph{projected} (post-machinery) targets separately so the two
sources of incoherence can be distinguished (§3.7); the extreme tell is the bare-workflow agent's thesis-free one-line rationale
(§7.7.2). Read at the agent-in-the-system level, the gap measures a real reasoning--action
consistency deficit: a competent reader, handed the agent's analysis, would not derive the book it
produced.

\textbf{Scope: what the cross-track result does and does not establish.} The divergence persists on
genuinely post-cutoff live data, making a period-specific or leakage-only explanation less
plausible. But the live and backtest cohorts are different model \emph{versions}, so the result is
\textbf{instrument-consistency, not same-model generalization}. The per-track gaps are \emph{similar, not
identical} (+0.30 vs +0.23 under different primaries), consistent with one systematic deficit
measured at judge-dependent levels rather than a proof of equality. The pattern is consistent with,
but does not test, the declarative-to-procedural interpretation motivated by ACT* (§2.4). One
further caution: Composure and Acuity --- the shortest-baseline axes --- moved with each horizon
extension while Coherence, Reliability, and Discipline stayed stable. The short-horizon axes
therefore warrant proportionate caution in interpreting any short live record, our own earlier
reads included.

\textbf{Robustness to the judge.} The second judge rules out one judge's idiosyncrasy, addressing a
different failure mode than the cross-track test. The per-track primaries are calibrated on the backtest
rounds both have scored (protocol in §6.4.1). Over 633 shared rounds they agree within one rubric
point on 97\% of rounds (Spearman ρ = 0.65, Pearson r = 0.66; GLM runs \textasciitilde0.2 rubric-points more
lenient, mean 2.32 vs 2.12). The gap survives a full judge swap on the backtest --- +0.27 under
GLM vs +0.30 under minimax-m3: \textbf{the divergence is a property of the \emph{allocations}, not of one
particular judge.}

\subsubsection{7.10 Behavioral Interpretability: Findings}\label{behavioral-interpretability-findings}

The most instructive live behavior: an
autonomous-mode convergence to hold (no rebalance) with two distinct, model-specific mechanisms ---
both prevented by the structured scaffold.

\textbf{Two roads to the same hold (Fig. 13).} Across the eighteen valid live days, exactly two of six
models ever produce a null-strategy hold, and only in autonomous mode: gemini-3.5-flash holds 9 of
18 rounds and minimax-m3 8 of 18; the other four autonomous models --- and all six in structured
mode --- hold 0\%. The roads are distinct:

\begin{itemize}
\tightlist
\item
  \emph{gemini-3.5-flash --- tool-turn-budget exhaustion.} Its reflection carries a complete mean-variance
  thesis (ten high-momentum names, a risk-on regime read), yet the committed round records only a
  truncation marker. The budget is spent exploring, no final allocation is serialized, and an
  empty-weights fallback holds the book --- it \emph{reasoned} but did not \emph{deliver in budget}.
\item
  \emph{minimax-m3 --- decision-schema-adherence failure.} The round is rejected for schema-invalid
  output --- an empty value where the schema required a list --- and with no valid decision it falls
  back to hold.
\end{itemize}

\textbf{Intermittent, independent, recoverable --- not a one-way collapse.} The first live week looks like
``day-1 trade, then lock-up'' --- both models traded day 1 then held days 2--4 in lockstep, which a
short pilot would report as a deterministic day-2 collapse. The second week overturns that: both
models trade on several later days, so the holds are intermittent --- \emph{recoverable}, not absorbing.
From week 2 on, the holds are also independent: the two models hold on \emph{different} days, with no
synchrony (per-day dates in Appendix E). The diagnostically correct summary is a
\emph{per-model autonomous hold-rate} (gemini 0.50, minimax 0.44 over the completed campaign), reflecting
a stochastic, model-specific fragility exposed by the autonomous scaffold. It is \emph{not} a shared,
monotone pattern in which ``models stop trading after day 1.''

\textbf{The remediation boundary turns the diagnosis into an intervention test.} Both mechanisms were
root-caused mid-campaign from the audit trail. The round-14 remediation (§7.9) deployed matching
scaffold-level counter-measures: output-schema coercion for the list-type failure, and fail-loud
decision retries (with the previous failure named in the retry prompt) to replace the silent
empty-weights fallback. The change is sharp: \textbf{every one of the 17 hold rounds precedes the
boundary, and neither model holds in the six valid days from the boundary session through campaign
end (0 of 12 model-days)} --- with no change to the models themselves. Re-running a failed round had
reproduced the failure under the earlier scaffold. The holds' disappearance after the targeted,
prospectively deployed remediation is therefore mechanism-consistent intervention evidence that scaffold-level failure handling interacts with model-specific output fragility --- the holds were not strategic choice. It is not
a general efficacy estimate: a bundled scaffold change, two models, six post-boundary days, no
parallel control. Historical rounds stay as recorded (the live series is path-dependent), flagged
by the scorer and excluded from the consistency axis.

\begin{figure}[tbp]\centering
\includegraphics[width=0.58\linewidth,height=\textheight,keepaspectratio]{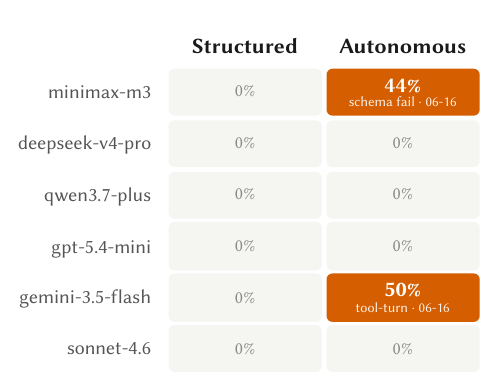}

\vspace{7pt}
\begin{minipage}{0.86\textwidth}\small \textbf{Figure 13.}\textit{ Live hold-round rate by model × mode over the eighteen valid days. Only two models
ever hold, only in autonomous mode --- gemini-3.5-flash (0.50, tool-turn truncation), minimax-m3
(0.44, schema-adherence failure) --- via \emph{different} mechanisms; the holds are intermittent,
recoverable, and confined to the pre-remediation segment (all 17 hold rounds precede the round-14
scaffold fix, after which both models trade every day, §7.10). Structured mode = 0\% for all.}\end{minipage}
\end{figure}

\textbf{Scaffold magnifies vs.~constrains (the structured--autonomous thread).} The same minimax-m3 in
\emph{structured} mode still logs parse failures yet keeps trading: the staged pipeline (Risk-Officer
review, MAPC reconciliation) contains a malformed sub-output and recovers. The single-shot
autonomous path lets the same failure collapse the round. Structure converts raw fragility into
graceful degradation, at the cost of the autonomy that, for the more capable models, yields the
cohort's \emph{best} decisions (Table 10). For the fragile pair, the composite score and failure
containment tell different stories. minimax-m3's autonomous APM-CS still exceeds its structured
one (41.8 vs 29.1) because its structured committee legs are weaker still. The scaffold's value
shows in \emph{which} failures occur, not in the composite. Whether structure or autonomy is the right deployment is
model-dependent, not settled (§8.3).

\textbf{Competence, read fairly.} The trail is not only a failure catalogue. claude-sonnet-4.6's day-1
rationale is crisp and executable: it reads a risk-on regime (VIX \textasciitilde18), flags a negative momentum
loading in its factor model, rotates to correct it, and exits its worst-momentum,
highest-volatility names --- a defensible thesis the allocation then follows and the scorecard
rewards (Coherence 0.60, APM-CS 83.6 --- the live leader).

Every quote above is pinned to a hash-verified \emph{round\_id} (anecdote bank, Appendix E); each behavioral lens of §6.5 maps back to a capability axis --- the qualitative read
\emph{explains, in the model's own words and actions, how} the quantitative deficits arise. We report
the most diagnostic slice, the hold-mechanism split (Figure 13) and the anecdote bank, and leave
fuller per-model behavioral tabulation to future work.

\subsubsection{7.11 Summary}\label{summary}

\textbf{The validation arc closes from backtest, through ablation, to live execution --- on chain-verified
data at every step.} Read against §7.1's questions: the scorecard discriminates decision quality
where Sharpe conflates it and exposes the nominal Sharpe winner as a reliability artifact (RQ1--RQ2,
§7.3--§7.4). The audit trail localizes three model-specific reliability mechanisms in backtest and two
autonomous hold mechanisms in live execution. A targeted mid-campaign scaffold remediation eliminates
the latter and doubles as an intervention test (RQ3, §7.5, §7.10). The scaffold reshapes the
capability profile by model-specific amounts, with the live cohort uniformly favoring autonomy on the
composite (RQ4, §7.6, §7.9). Module value cannot be ranked from returns even with repeated runs, but
registers on the capability and behavioral axes (RQ5, §7.7). The cost-tier grid validates LOW and
exposes HIGH as mis-calibrated --- the instrument auditing itself (RQ6, §7.7). Net of costs, the
agents clear the defensive baselines but not the cap-weighted index, a de-risking trade rather than
absent skill (RQ7, §7.8). No configuration dominates all five axes, so by the instrument's own
standard the cohort is not cleanly rankable. The coherence gap, reproducing on genuinely unseen data at similar magnitude, is the
strongest single exhibit (§7.9).

\subsection{8. Discussion}\label{discussion}

We draw the inferences the campaign supports: what diagnosis-before-ranking buys (§8.1), what the axes say beyond finance (§8.2), what the scaffold contrast implies for deploying LLM investors (§8.3), and lessons for benchmark builders (§8.4). §8.5 bounds every claim, §8.6 names the next priorities, §8.7 closes with ethics and broader impact.

\subsubsection{8.1 Diagnosis Before Ranking}\label{diagnosis-before-ranking}

§7.11 collects the findings; here we draw the inference. A leaderboard would report a spurious number and stop; the scorecard returns a \emph{profile} per configuration --- where each model's process breaks, and under which scaffold. \textbf{The same evidence explains why we decline to rank: no configuration dominates all five axes, and a credible cross-regime ranking would need an order of magnitude more models, regimes, and repeated runs} --- out of scope by design.

\subsubsection{8.2 Capability Axes as General Agent Diagnostics}\label{capability-axes-as-general-agent-diagnostics}

\textbf{The axis→construct mapping.} Each axis is a finance instantiation of a general capability. Coherence: \emph{systems-level reasoning vs.~greedy local moves}, the agentic analogue of chain-of-thought \emph{unfaithfulness} (Turpin et al.~2023) --- the judge/agreement divergence (§7.9) shows a model naming the right consideration without letting it govern the output. Acuity: \emph{salience attention under polluted context} (the negative-Acuity autonomous run reads as distractibility). Composure: \emph{robustness vs.~reactive overcorrection} (gemini-autonomous's vol-driven churn, §7.6). Discipline: \emph{systematic alignment to a standing mandate vs.~turn-by-turn instruction-following} --- self-regulation without a hard guardrail. Reliability: \emph{protocol/format adherence and operational robustness} --- valid form, every time.

\textbf{The ACT* reading of the coherence gap.} In Anderson's ACT* terms (Anderson 1983), the deficit lies in the transition from declarative to procedural knowledge. The agent states the correct consideration (``VIX is elevated, reduce momentum exposure'') but lacks the compiled \emph{procedural} rule converting it into specific weights. ACT* predicts this failure for text-acquired knowledge never compiled through practice. The gap recurs at similar magnitude across backtest and live (+0.30 / +0.23; protocol-clean within-judge band +0.23\ldots+0.30, §7.9). Our most parsimonious interpretation --- not a tested causal claim --- is that this is consistent with the deficit rather than noise: declarative store and missing compilation both transfer to unseen data. The reflect-and-learn loop (ALE and three-tier memory, §3.6) could in principle supply the missing practice. The ablation grid (§7.7) poses the question of whether in-context experience begins compiling the rules; a dedicated analysis of Coherence under ablation is left to the broader-run pass (§8.6).

\textbf{Parallels in the benchmark literature.} Three findings rhyme with results from very different instruments --- mild evidence they are properties of the \emph{agents}, not our setup. The coherence gap echoes Profit Mirage (Li et al.~2025c): apparent alpha collapsing once leakage is removed, apparent skill failing to cohere with stated reasoning. The capability ≠ outcome result (§7.3) is the portfolio instance of procedure-aware evaluation's ``corrupt success'' (Cao et al.~2026): success through an unsound process. The repeated-run noise band (§7.7) concretizes the field's statistical-power cautions: a single-axis, single-run score is not just incomplete but \emph{unstable}.

\textbf{A portable lens.} The same five questions apply to any agentic system operating a complex, cost-bearing, partially observable process, and the same audit trail supports the behavioral companion read of §6.5/§7.10 (intent-vs-clip gaps, committee negotiations, tool-foraging order, lesson reuse; anecdote bank in Appendix E).

\subsubsection{8.3 Scaffolding × Capability, and the Cost of Intelligence}\label{scaffolding-capability-and-the-cost-of-intelligence}

The structured↔autonomous contrast is the campaign's clearest interaction. Structure lifts Composure robustly and Discipline partly by construction (D4 is mode-confounded, §6.4), with the largest descriptive composite contrast at deepseek (APM-CS +38). But it also \emph{demands} a decision schema that stresses smaller models differently (gemini's genuine parse-fails; qwen3's tool-loop non-termination, §7.5). Structure thus \emph{helps and hurts the same model on different axes} --- deepseek's reliability reverses with mode. In the efficiency ledger, gemini·structured is the heaviest reasoner (\textasciitilde280k reasoning tokens/run despite thinking pinned off). Its compute is consumed by the retry loop around schema failures: the ``cost of intelligence'' is dominated by \emph{failed}, not productive, reasoning (Figure 14).

\begin{figure}[tbp]\centering
\includegraphics[width=0.9\linewidth,keepaspectratio]{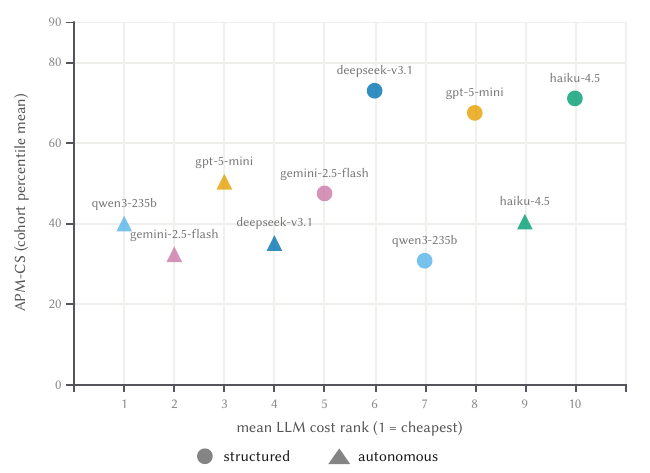}

\vspace{7pt}
\begin{minipage}{0.86\textwidth}\small \textbf{Figure 14.}\textit{ Capability (APM-CS) vs.~average LLM cost rank (1 = cheapest); circles = structured, triangles = autonomous. deepseek·structured is highest-capability (72.9) among cheaper configurations; haiku·structured comparable (71.0) at several times the cost; gemini·structured, the heaviest reasoner, only mid-capability --- compute consumed by the parse-fail retry loop. Capability does not track spend.}\end{minipage}
\end{figure}

\textbf{Which operating mode --- structured, autonomous, or hybrid?} Structure is a \emph{containment} layer, converting raw fragility into graceful degradation (minimax-m3 keeps trading in structured mode but collapses to holds in autonomous mode, §7.10). Autonomy destabilizes models that cannot self-govern (gemini-3.5-flash and minimax-m3 collapse to intermittent holds). Yet every live model's composite scores higher in autonomous mode (Table 10). The gains are largest and cleanest for the capable self-governors (claude-sonnet-4.6 83.6, qwen3.7-plus 69.1, both at full Reliability), while gemini's and minimax's smaller edges depend on the rounds their hold-collapses spare. \textbf{The right mode is model-dependent, and the interesting frontier is a hybrid} --- autonomous operation with a lightweight governance layer (schema validation, mandate reconciliation, a held-out coherence check) catching exactly the failures structure would have prevented. The round-14 remediation is a scoped existence proof, though the intervention was uncontrolled (§7.10). A thin governance layer eliminated both models' observed hold-collapses without visibly costing autonomy's composite advantage. Whether an LLM should act as a \emph{disciplined expert-human proxy} (structured) or be trusted to \emph{exceed human heuristics} (autonomous) is thus a function of self-governance capability, which the scorecard measures directly.

\subsubsection{8.4 Benchmark Hygiene: Output Budgets, Cost Realism, Auditability}\label{benchmark-hygiene-output-budgets-cost-realism-auditability}

Two warnings generalize beyond CLQT. First, \textbf{output budgets}: a reasoning model under a tight output cap can look like an incapable portfolio manager when merely \emph{truncated}. The fix is simple (budget generously; billing is on generated tokens), and the telemetry mandatory (cap-hits, truncation, parse-fail). Otherwise, a budget artifact is misattributed to capability. qwen3's telemetry first surfaced and then \emph{ruled out} the budget explanation, steering diagnosis to the true tool-loop cause (§7.5). Second, \textbf{cost realism must be audited, not assumed}: the cost-tier ablation surfaced the instrument's own measurement defect. The HIGH (Almgren--Chriss) tier charges an implausible \textasciitilde+527 bps/round; it is mis-calibrated rather than ``pessimistic''. We therefore retain LOW (validated per §3.4/§5.2), leaving HIGH-tier recalibration to future work. The recompute-verifiable chain (§4.4) is the auditability half of the same discipline, though, as scoped in §4.4, without externally anchored non-repudiation.

Four checks should precede any closed-loop trading-agent claim: (i) point-in-time enforcement stated and verified, not assumed; (ii) reliability telemetry (completion, parse-fail, truncation) reported beside returns; (iii) every ablation or model delta read against a repeated-run noise band; and (iv) any LLM judge held out of the evaluated cohort, with rubric and caching disclosed. Each is cheap.

\subsubsection{8.5 Limitations}\label{limitations}

We read this single-period campaign through the standard threats-to-validity frame.

\begin{itemize}
\item
  \textbf{What returns can identify.} The returns axis is underpowered for module attribution at this scale: no single-module or cluster knockout's run range clears the ±0.42 repeated-run band (§7.7). No module-importance claim can therefore rest on returns; module value is confined to the capability axes (e.g., analysis-blind drives Acuity to ≈−0.01). Scope conditions: the HIGH cost tier is mis-calibrated (\textasciitilde527 bps/round; LOW retained); Acuity and Discipline are mode-confounded, read only within mode; the ablation grid runs on one anchor model (gemini·structured), so module effects may be model-dependent.
\item
  \textbf{Construct validity of the axes.} The five axes operationalize latent capabilities; the revision history of Discipline, Composure, and the Acuity noise set (Appendix F) shows the read is sensitive to construct definition. Validity is argued empirically, not assumed. The evidence includes rank-stability across axis revisions (per-revision Kendall τ = 0.75--0.86, Spearman ρ = 0.88--0.94; τ = 0.58 with all revisions compounded --- Appendix F.1), near-orthogonality (mean \textbar r\textbar{} = 0.23), and robustness to leave-one-out and judge-reweighting (Appendix F). One revision overturned a prior sub-finding: memory ``breaking'' Composure was an artifact of the earlier construction. Stated, not proven: the informative composite (momentum + earnings surprise + sentiment) is the normative yardstick for the agreement leg and Acuity \emph{by construction}; its alpha content in this window is not separately demonstrated. The ±0.42 noise band is estimated on the anchor configuration and assumed to transfer across variants.
\item
  \textbf{Judge independence and validity.} A held-out, non-cohort judge mitigates but cannot fully eliminate LLM-as-a-judge self-preference. A second, different-family judge corroborates the headline gap on the backtest (GLM-5.2, ρ = 0.65 over 633 shared rounds; gap +0.27 vs minimax-m3's +0.30). Each track's primary judge is out-of-cohort by enforced protocol (§6.4.1; Appendix C). Part of what the judge scores is the post-construction target reshaped by the optimizer and projection; the proposed/projected split attributes the share due to this machinery (§7.9). A blinded human-anchoring study (Appendix F.5) anchors the judge's ordering to a human rater at inter-judge-agreement level (ρ = 0.60 vs 0.65); two auxiliary frontier raters corroborate this on the same blinded sheet. The remaining limitation --- a single author-rater rather than an independent expert panel --- is a §8.6 priority.
\item
  \textbf{External validity: one regime, modeled backtest costs.} The window is a single, predominantly rising market (SPY +27\%, max DD −8.9\%, VIX median 17); backtest costs are modeled, not executed (the live track supplies real fills on the daily deltas). Returns-based statements are bounded to one regime, and the bound cuts both ways: the benign window \emph{masks} modules whose value concentrates at turning points and \emph{flatters} high-exposure risk-taking while penalizing the agents' prudent de-risking (§7.7). Capability and risk-adjusted axes are comparatively regime-robust; absolute-return conclusions are regime-conditional.
\item
  \textbf{Live evidence: consistency, not transfer.} The live track shows the instrument measures the same decision property on genuinely unseen data, not that any model generalizes --- the cohorts are different model versions, and live returns are indicative-only by design (daily flatten; overnight exposure virtual and modeled). The four-week horizon keeps Acuity and Composure the shortest-baseline axes, and each live-horizon extension revised a short-horizon read (§7.9--§7.10) --- a caution against over-reading any short live record, ours included. Of three operational events disclosed in §7.9, two bound the read; the broker-gateway outage is decision-level-neutral by construction. The mid-campaign prompt/failure-handling remediation splits the campaign at round 14. Pooled results are primary; the one boundary-sensitive metric (autonomous hold rate) is reported separately before and after the boundary, which also serves as an intervention test (§7.10). In addition, one model's final round ran at reduced reasoning effort after a non-terminating reasoning loop. Together, these two events leave the panel not perfectly homogeneous in prompt or effort conditions.
\item
  \textbf{Cohort and configuration scope.} The panel is small --- five backtest and six live models, one standing mandate, one reasoning-effort level per track --- so capability-axis statements are bounded to this cohort and configuration, and APM-CS is additionally cohort-relative (§6.4). qwen3-235b's training cutoff sits within a documented ≤1-month grace of the backtest start (§7.2), a contamination-adjacent caveat the post-cutoff live track lacks. Built-in contamination probes (canary documents, future-shift probes) went unexercised in this benign campaign (§4.3); integrity rests on the point-in-time gate, cutoff filtering, and the verified chain.
\end{itemize}

None of these undercut the contribution --- the instrument and its read-out --- but all bound what the \emph{numbers} may claim.

\subsubsection{8.6 Future Work}\label{future-work}

\begin{itemize}
\tightlist
\item
  \textbf{Scale and statistical power.} Broaden the model panel and repeated runs so every axis carries confidence intervals and rank-stability rather than point estimates and run-ranges.
\item
  \textbf{Regime robustness.} Stratify rounds by market regime and report how far each capability degrades from calm into stress --- the load-bearing extension before returns-based conclusions generalize beyond this single benign, rising market.
\item
  \textbf{Generalization of the profile.} Vary the standing mandate and reasoning-effort budget to test whether a model's capability fingerprint is a stable property or an artifact of the single mandate and effort level used here.
\item
  \textbf{Active stress and instrument hardening.} Adversarial misleading-data and canary probes to stress Acuity and Composure directly; a larger independent human-rater panel (extending the blinded author anchoring of Appendix F.5); a broader held-out judge panel; a recalibrated HIGH cost tier.
\end{itemize}

Finer-grained per-sector and per-name attribution is a natural next layer, localizing not just \emph{where} a process fails but on \emph{which} exposures.

\subsubsection{8.7 Ethics and Broader Impact}\label{ethics-and-broader-impact}

CLQT evaluates agents on historical data and broker \emph{paper} trading only; no real capital is deployed, and nothing here claims any evaluated configuration is fit for deployment --- the findings argue the opposite. The dual-use question is whether a trading-agent benchmark is a recipe for deploying LLM traders. Our results argue against this reading: a cohort-wide coherence failure under an independent judge, reliability deficits invisible to returns, and scaffold-dependent decision collapses are the evidence a responsible pre-deployment gate should demand. We frame the instrument as that gate. Correlated decisions from similar foundation models raise legitimate herding and stability concerns; an instrument measuring overreaction (Composure) and mandate adherence (Discipline) is a tool for \emph{studying} that risk, not amplifying it. The recompute-verifiable audit trail (§4.4) supplies a decision-level integrity record that can support regulatory review of AI in finance. All market, news, and macro data come from commercial and public APIs under their terms of service; no personal data is collected or processed.

\subsection{9. Conclusion}\label{conclusion}

Returns-only leaderboards mismeasure LLM trading agents --- and this paper shows the failure is identifiable and fixable. In a year of chain-verified evidence, the nominal Sharpe ``winner'' was a 5-of-26-round reliability artifact, the capability leader sat mid-pack on returns, and no configuration dominated all five axes: by the instrument's own standard, ranking would be premature. CLQT is the instrument that makes this visible --- a closed-loop, cost-aware, temporally gated benchmark in which every reported number can be recomputed from a hash-verifiable audit trail. It combines a five-axis capability scorecard (APM-CS), whose axes are finance instantiations of general agent capabilities, with a held-out, two-judge Coherence protocol. It treats process scaffolding as a measured experimental variable through to live broker execution.

The finding the field can build on is the \emph{stating-versus-doing} gap: agents agree with market signals while failing to follow their own stated analysis. The +0.30/+0.23 agreement--judge divergence reproduces across two cohorts, two horizons, and two held-out judges on genuinely unseen data (instrument-consistency across model generations, not same-model transfer). The capability and behavioral axes reveal differences in module value that returns cannot distinguish. A targeted mid-campaign scaffold remediation eliminated both autonomous hold mechanisms (§7.10) --- evidence that the diagnosis is actionable, not merely descriptive. A credible ranking of LLM portfolio managers still needs more models, regimes, and repeated runs; CLQT supplies the audit trail, the axes, and the capability map to make those rankings scientifically interpretable. Diagnosis before ranking is not a retreat from ambition --- it is how benchmarks for autonomous agents should be built.

\subsection{References}\label{references}
\begingroup\small

Almgren R, Chriss N (2001) Optimal execution of portfolio transactions. \emph{Journal of Risk}, 3(2), 5--39.

Anderson J R (1983) \emph{The Architecture of Cognition.} Harvard University Press. \emph{(ACT*; declarative vs.~procedural knowledge, compilation through practice.)}

Anthropic (2024) Model Context Protocol (MCP): An Open Standard for Connecting AI Assistants to Tools and Data. https://modelcontextprotocol.io

Boyd J R (1987) \emph{A Discourse on Winning and Losing.} Air University. \emph{(The OODA decision cycle.)}

Cai X, Guo M, Liu J, Han J, Guo B, Long Y, Chen Y, Wu B, Metaxas D N, Li R (2026) OpenPM: Auditable Point-in-Time Evaluation for LLM Portfolio-Management Agents. arXiv:2608.09988.

Cao H, Driouich I, Thomas E (2026) Beyond Task Completion: Revealing Corrupt Success in LLM Agents through Procedure-Aware Evaluation. arXiv:2603.03116.

Chan A, et al (2026) LATTICE: Evaluating Decision Support Utility of Crypto Agents. arXiv:2604.26235.

Chen Y, et al (2025) StockBench: Can LLM Agents Trade Stocks Profitably In Real-world Markets? arXiv:2510.02209.

De Bondt WFM, Thaler R (1985) Does the Stock Market Overreact? \emph{The Journal of Finance}, 40(3), 793--805.

Endsley M R (1995) Toward a Theory of Situation Awareness in Dynamic Systems. \emph{Human Factors}, 37(1), 32--64.

Fan T, et al (2025) AI-Trader: Benchmarking Autonomous Agents in Real-Time Financial Markets. arXiv:2512.10971.

Grądzki P, Wójcik P, Lessmann S (2025) Algorithmic crypto trading using information-driven bars, triple barrier labeling and deep learning. Financ Innov 11:136. doi:10.1186/s40854-025-00866-w.

Hong S, et al (2024) MetaGPT: Meta Programming for a Multi-Agent Collaborative Framework. \emph{ICLR}. arXiv:2308.00352.

Kim S, Kwon H, Yoon J, Kim H (2026) Dynamic factor-informed reinforcement learning for enhancing portfolio optimization. Financ Innov 12:3. doi:10.1186/s40854-025-00803-x.

Kolb D A (1984) \emph{Experiential Learning: Experience as the Source of Learning and Development.} Prentice-Hall.

Kou G, Li Y, Wang H, Wang X (2026) Human--AI hybrid finance: from AI tools to decision systems. Financ Innov 12:125. doi:10.1186/s40854-026-00941-w.

Li C, et al (2026) NextFund: A Unified Performance Tracking Platform for Agentic Portfolio Management. arXiv:2607.11141.

Li C, Shi Y, Luo Y, Tang N (2025a) Will LLMs be Professional at Fund Investment? DeepFund: A Live Arena Perspective. arXiv:2503.18313.

Li H, et al (2024) InvestorBench: A Benchmark for Financial Decision-Making Tasks with LLM-based Agent. ACL 2025. arXiv:2412.18174.

Li WW, Kim H, Cucuringu M, Ma T (2025b) Can LLM-based Financial Investing Strategies Outperform the Market in Long Run? (FINSABER). arXiv:2505.07078.

Li X, et al (2025c) Profit Mirage: Revisiting Information Leakage in LLM-based Financial Agents. arXiv:2510.07920.

Liang P, et al (2022) Holistic Evaluation of Language Models (HELM). \emph{TMLR} (2023). arXiv:2211.09110.

Liu X, et al (2024) AgentBench: Evaluating LLMs as Agents. \emph{ICLR}. arXiv:2308.03688.

Nguyen P, Pham T (2026) Toward Reliable Evaluation of LLM-Based Financial Multi-Agent Systems: Taxonomy, Coordination Primacy, and Cost Awareness. arXiv:2603.27539.

Packer C, et al (2023) MemGPT: Towards LLMs as Operating Systems. arXiv:2310.08560.

Papadakis C, et al (2025) StockSim: A Dual-Mode Order-Level Simulator for Evaluating Multi-Agent LLMs in Financial Markets. arXiv:2507.09255.

Park JS, et al (2023) Generative Agents: Interactive Simulacra of Human Behavior. \emph{UIST}. arXiv:2304.03442.

Peng X, et al (2026) Herculean: An Agentic Benchmark for Financial Intelligence. arXiv:2605.14355.

Qian L, et al (2025) When Agents Trade: Live Multi-Market Trading Benchmark for LLM Agents. \emph{The Web Conference (WWW) 2026}. arXiv:2510.11695.

Qin Y, et al (2024) ToolLLM: Facilitating LLMs to Master 16000+ Real-World APIs. \emph{ICLR}. arXiv:2307.16789.

Shinn N, et al (2023) Reflexion: Language Agents with Verbal Reinforcement Learning. \emph{NeurIPS}. arXiv:2303.11366.

Sumers T, Yao S, Narasimhan K, Griffiths T (2024) Cognitive Architectures for Language Agents (CoALA). \emph{TMLR}. arXiv:2309.02427.

Turpin M, Michael J, Perez E, Bowman S R (2023) Language Models Don't Always Say What They Think: Unfaithful Explanations in Chain-of-Thought Prompting. \emph{NeurIPS}. arXiv:2305.04388.

Xia Y, et al (2026) Agentic Trading: When LLM Agents Meet Financial Markets. arXiv:2605.19337.

Xiao Y, Sun E, Luo D, Wang W (2024) TradingAgents: Multi-Agents LLM Financial Trading Framework. arXiv:2412.20138.

Yang K, Deng R, Wei Y, Wang S (2025) The power of ChatGPT in processing text: evidence from analysis and prediction in the exchange rate markets. Financ Innov 11:118. doi:10.1186/s40854-025-00789-6.

Yao J, Zheng Z (2026) Beyond Agent Architecture: Execution Assumptions and Reproducibility in LLM-Based Trading Systems. arXiv:2606.08285.

Yao S, et al (2023) ReAct: Synergizing Reasoning and Acting in Language Models. \emph{ICLR}. arXiv:2210.03629.

Yao S, et al (2024) τ-bench: A Benchmark for Tool-Agent-User Interaction in Real-World Domains. arXiv:2406.12045.

Ye Y, et al (2026) The Alpha Illusion: Reported Alpha from LLM Trading Agents Should Not Be Treated as Deployment Evidence. arXiv:2605.16895.

Yu H, Li F, You J (2025) LiveTradeBench: Seeking Real-World Alpha with Large Language Models. arXiv:2511.03628.

Yu Y, et al (2023) FinMem: A Performance-Enhanced LLM Trading Agent with Layered Memory and Character Design. arXiv:2311.13743.

Yuan X, et al (2026) TraderBench: How Robust Are AI Agents in Adversarial Capital Markets? arXiv:2603.00285.

Zhang K, et al (2026a) OpenFinGym: A Verifiable Multi-Task Gym Environment for Evaluating Quant Agents. arXiv:2606.26350.

Zhang W, et al (2025) FinWorld: An All-in-One Open-Source Platform for End-to-End Financial AI Research and Deployment. arXiv:2508.02292.

Zhang W, et al (2026b) AlphaForgeBench: Benchmarking End-to-End Trading Strategy Design with Large Language Models. arXiv:2602.18481.

Zhao Y, Chen S, Su N (2026) PortBench: A Correlation-Aware, Full-Pipeline Benchmark for LLM-Driven Portfolio Management. arXiv:2605.27887.

Zheng L, et al (2023) Judging LLM-as-a-Judge with MT-Bench and Chatbot Arena. \emph{NeurIPS Datasets \& Benchmarks}. arXiv:2306.05685.

Zhu T, et al (2026) From Knowing to Doing: A Memory-Controlled Benchmark for LLM Trading Agents on Stock Markets. arXiv:2605.28359.

\endgroup
\clearpage
\subsection{Appendix A: Cost Tier Parameterizations}\label{appendix-a-cost-tier-parameterizations}

{\def\LTcaptype{none} 
\begin{longtable}[]{@{}
  >{\raggedright\arraybackslash}p{(\linewidth - 10\tabcolsep) * \real{0.1529}}
  >{\raggedright\arraybackslash}p{(\linewidth - 10\tabcolsep) * \real{0.1412}}
  >{\raggedright\arraybackslash}p{(\linewidth - 10\tabcolsep) * \real{0.1882}}
  >{\raggedright\arraybackslash}p{(\linewidth - 10\tabcolsep) * \real{0.1647}}
  >{\raggedright\arraybackslash}p{(\linewidth - 10\tabcolsep) * \real{0.1882}}
  >{\raggedright\arraybackslash}p{(\linewidth - 10\tabcolsep) * \real{0.1647}}@{}}
\toprule\noalign{}
\begin{minipage}[b]{\linewidth}\raggedright
Tier
\end{minipage} & \begin{minipage}[b]{\linewidth}\raggedright
Spread (bps)
\end{minipage} & \begin{minipage}[b]{\linewidth}\raggedright
Commission (bps)
\end{minipage} & \begin{minipage}[b]{\linewidth}\raggedright
Slippage (bps)
\end{minipage} & \begin{minipage}[b]{\linewidth}\raggedright
Borrow (ann bps)
\end{minipage} & \begin{minipage}[b]{\linewidth}\raggedright
Impact Model
\end{minipage} \\
\midrule\noalign{}
\endhead
\bottomrule\noalign{}
\endlastfoot
zero & 0 & 0 & 0 & 0 & none \\
low \emph{(default)} & 2.0 & 0.5 & 1.0 & 30 & square\_\allowbreak{}root \\
medium & 3.0 & 1.0 & 2.0 & 50 & square\_\allowbreak{}root \\
high & 5.0 & 2.0 & 5.0 & 100 & almgren\_\allowbreak{}chriss \\
\end{longtable}
}

\begin{center}\begin{minipage}{0.86\linewidth}\small
\textbf{Table A1.}\textit{ Cost tier parameterizations. Live broker paper fills (§5.2) put the flat commission at \textasciitilde\$1 USD per fill --- below the modeled 0.5 bps at the campaign's typical \$200k+ position sizes, so the LOW tier's commission leg is, if anything, conservative. Caveat: the HIGH (Almgren--Chriss) tier as parameterized produces \textasciitilde+527 bps/round in the campaign (§7.7) and must be recalibrated before use.}
\end{minipage}\end{center}

\subsection{Appendix B: SCOUT Signal Computation and MCP Tool Inventory}\label{appendix-b-scout-signal-computation-and-mcp-tool-inventory}

Momentum (\(m_{12,1}\)): \((P_{t-21}/P_{t-252}) - 1\) on adjusted closes (1-month skip-lag avoids short-term reversal). Earnings surprise (\(\epsilon_{\text{eps}}\)): \((\text{EPS}_{\text{act}} - \text{EPS}_{\text{est}})/|\text{EPS}_{\text{est}}|\), clamped to \([-5,+5]\); None/excluded when no estimate. Macro correlation (\(\rho^{\text{mkt}}\)): 60-day rolling Pearson correlation of daily log returns vs SPY. Degenerate-window guard: the correlation is set to None and its weight redistributed when the window has \textless10 observations or near-zero variance.

All quantitative analysis (§3.1) is performed through 19 tools defined to the MCP standard and grouped into four namespaces. The benchmark invokes them \emph{in-process} against the underlying quant/data libraries. We use the MCP tool-definition interface and schemas, not a networked server transport. Every call is hashed, timed, and logged, so tool \emph{use} is itself auditable.

{\def\LTcaptype{none} 
\begin{longtable}[]{@{}
  >{\raggedright\arraybackslash}p{(\linewidth - 2\tabcolsep) * \real{0.2326}}
  >{\raggedright\arraybackslash}p{(\linewidth - 2\tabcolsep) * \real{0.7674}}@{}}
\toprule\noalign{}
\begin{minipage}[b]{\linewidth}\raggedright
Namespace
\end{minipage} & \begin{minipage}[b]{\linewidth}\raggedright
Tools
\end{minipage} \\
\midrule\noalign{}
\endhead
\bottomrule\noalign{}
\endlastfoot
\emph{clqt-market-data} & \emph{fetch\_\allowbreak{}market\_\allowbreak{}bars},\allowbreak{} \emph{fetch\_\allowbreak{}market\_\allowbreak{}snapshot},\allowbreak{} \emph{get\_\allowbreak{}universe},\allowbreak{} \emph{fetch\_\allowbreak{}fundamentals} \\
\emph{clqt-quant-tools} & \emph{compute\_\allowbreak{}signals},\allowbreak{} \emph{compute\_\allowbreak{}risk\_\allowbreak{}report},\allowbreak{} \emph{optimize\_\allowbreak{}portfolio},\allowbreak{} \emph{estimate\_\allowbreak{}transaction\_\allowbreak{}costs},\allowbreak{} \emph{compute\_\allowbreak{}portfolio\_\allowbreak{}analytics},\allowbreak{} \emph{compute\_\allowbreak{}covariance\_\allowbreak{}matrix},\allowbreak{} \emph{compute\_\allowbreak{}factor\_\allowbreak{}attribution},\allowbreak{} \emph{compute\_\allowbreak{}regime\_\allowbreak{}performance},\allowbreak{} \emph{compute\_\allowbreak{}alpha\_\allowbreak{}beta} \\
\emph{clqt-news-sentiment} & \emph{fetch\_\allowbreak{}news},\allowbreak{} \emph{fetch\_\allowbreak{}sentiment},\allowbreak{} \emph{fetch\_\allowbreak{}macro\_\allowbreak{}indicators} \\
\emph{clqt-execution} & \emph{simulate\_\allowbreak{}execution},\allowbreak{} \emph{execute\_\allowbreak{}orders},\allowbreak{} \emph{get\_\allowbreak{}execution\_\allowbreak{}report} \\
\end{longtable}
}

\begin{center}\begin{minipage}{0.86\linewidth}\small
\textbf{Table B1.}\textit{ Tool inventory --- 19 MCP-defined tools across four namespaces, invoked in-process.}
\end{minipage}\end{center}

\subsection{Appendix C: Model Registry --- IDs, Snapshots, and Training Cutoffs}\label{appendix-c-model-registry-ids-snapshots-and-training-cutoffs}

{\def\LTcaptype{none} 
\begin{longtable}[]{@{}
  >{\raggedright\arraybackslash}p{(\linewidth - 6\tabcolsep) * \real{0.5349}}
  >{\raggedright\arraybackslash}p{(\linewidth - 6\tabcolsep) * \real{0.1163}}
  >{\raggedright\arraybackslash}p{(\linewidth - 6\tabcolsep) * \real{0.1977}}
  >{\raggedright\arraybackslash}p{(\linewidth - 6\tabcolsep) * \real{0.1512}}@{}}
\toprule\noalign{}
\begin{minipage}[b]{\linewidth}\raggedright
Model (OpenRouter id)
\end{minipage} & \begin{minipage}[b]{\linewidth}\raggedright
Provider
\end{minipage} & \begin{minipage}[b]{\linewidth}\raggedright
Knowledge cutoff
\end{minipage} & \begin{minipage}[b]{\linewidth}\raggedright
Group
\end{minipage} \\
\midrule\noalign{}
\endhead
\bottomrule\noalign{}
\endlastfoot
deepseek/\allowbreak{}deepseek-chat-v3.1 & DeepSeek & Mar 2025 & Backtest \\
google/\allowbreak{}gemini-2.5-flash-lite & Google & Jan 2025 & Backtest \\
openai/\allowbreak{}gpt-5-mini & OpenAI & May 2024 & Backtest \\
anthropic/\allowbreak{}claude-haiku-4.5 & Anthropic & Feb 2025 ‡ & Backtest \\
qwen/\allowbreak{}qwen3-235b-a22b-2507 & Alibaba & Jun 2025 & Backtest \\
deepseek/\allowbreak{}deepseek-v4-pro & DeepSeek & not disclosed † & Live \\
google/\allowbreak{}gemini-3.5-flash & Google & Jan 2025 & Live \\
openai/\allowbreak{}gpt-5.4-mini & OpenAI & Aug 2025 & Live \\
anthropic/\allowbreak{}claude-sonnet-4.6 & Anthropic & Aug 2025 ‡ & Live \\
qwen/\allowbreak{}qwen3.7-plus & Alibaba & not disclosed † & Live \\
minimax/\allowbreak{}minimax-m3 & MiniMax & not disclosed † & Live · Judge \\
z-ai/\allowbreak{}glm-5.2 & Z.ai & late 2025 § & Judge \\
\end{longtable}
}

\begin{center}\begin{minipage}{0.86\linewidth}\small
\textbf{Table C1.}\textit{ Model registry for the backtest cohort, the live cohort (§7.9), and the held-out Coherence
judges (§6.4.1), all accessed via \textbf{OpenRouter} under the \texttt{provider/model} ids shown (dated suffixes pin
the snapshot). \emph{Reasoning:} backtest per §7.2 (gpt-5-mini effort \emph{low}, the rest non-thinking/off); live at
\emph{medium} (§7.9; disclosed deviation --- qwen3.7-plus's final round ran with reasoning off after a
non-terminating reasoning loop at medium). \emph{Cutoffs} are provider-stated where published; \textbf{‡} = Anthropic's \emph{reliable} cutoff
(Claude training extends to \textasciitilde Jan 2026, still pre-live); \textbf{§} = aggregator-reported approximate; \textbf{†} = no
cutoff disclosed. The contamination argument needs no tabulated date: training precedes each model's
2026 release, and all releases are well before the 2026-06-15 live start. Integrity is enforced by
\emph{executing on post-cutoff data} (cutoff gate + exchange-calendar exclusion). Backtest cutoffs precede the 2025-06-16
start (qwen3-235b's Jun 2025 within the ≤1-month grace, §7.2). \emph{Judge assignment (v1.1):} minimax-m3 is the
primary judge for the backtest cohort only (it is not in that cohort); the live panel --- which contains
minimax-m3 --- is judged by GLM-5.2, non-cohort on both tracks. The assignment is enforced by the
scorecard tooling, which refuses a judge matching any scored run. (In v1.0, minimax self-judged its own
live rounds and scored itself among the lowest, ≈0.13--0.25. No self-preference inflation was observed,
but the enforced hold-out removes the concern by construction, §8.5.)}
\end{minipage}\end{center}

\subsection{Appendix D: Audit-Trail Field Map for the Scorecard}\label{appendix-d-audit-trail-field-map-for-the-scorecard}

Every metric in this paper is recoverable from the persisted per-run store without new instrumentation.

{\def\LTcaptype{none} 
\begin{longtable}[]{@{}
  >{\raggedright\arraybackslash}p{(\linewidth - 2\tabcolsep) * \real{0.6512}}
  >{\raggedright\arraybackslash}p{(\linewidth - 2\tabcolsep) * \real{0.3488}}@{}}
\toprule\noalign{}
\begin{minipage}[b]{\linewidth}\raggedright
Source field(s)
\end{minipage} & \begin{minipage}[b]{\linewidth}\raggedright
Axis /\allowbreak{} metric powered
\end{minipage} \\
\midrule\noalign{}
\endhead
\bottomrule\noalign{}
\endlastfoot
\emph{portfolio\_\allowbreak{}after.nav/\allowbreak{}weights/\allowbreak{}leverage/\allowbreak{}cash} & returns,\allowbreak{} vol,\allowbreak{} Sharpe/\allowbreak{}Sortino/\allowbreak{}Calmar,\allowbreak{} MaxDD \\
\emph{execution\_\allowbreak{}report.\{turnover,\allowbreak{}total\_\allowbreak{}cost,\allowbreak{}commission,\allowbreak{}slippage,\allowbreak{}market\_\allowbreak{}impact,\allowbreak{}spread\}} & cost decomposition; Composure (turnover) \\
\emph{mapc\_\allowbreak{}delta\_\allowbreak{}proposed} + \emph{signals} & \textbf{Coherence} (agreement),\allowbreak{} \textbf{Acuity} (informative−noise Spearman) \\
\emph{research\_\allowbreak{}brief} /\allowbreak{} \emph{target\_\allowbreak{}allocation.rationale} + \emph{target\_\allowbreak{}weights} & \textbf{Coherence} (judge prompt) \\
\emph{consistency\_\allowbreak{}score.\{composite\_\allowbreak{}score,\allowbreak{}style\_\allowbreak{}drift\}} + \emph{realized\_\allowbreak{}vol} signal & \textbf{Composure} (drift,\allowbreak{} overreaction slope) \\
\emph{target\_\allowbreak{}allocation.constraints\_\allowbreak{}satisfied} (pre-MAPC),\allowbreak{} \emph{mapc\_\allowbreak{}adjustments\_\allowbreak{}clipped},\allowbreak{} turnover vs cap & \textbf{Discipline} \\
\emph{parse\_\allowbreak{}failures},\allowbreak{} \emph{decision\_\allowbreak{}attempts},\allowbreak{} \emph{agent\_\allowbreak{}costs{[}{]}.out\_\allowbreak{}tokens} (cap-hits),\allowbreak{} round count & \textbf{Reliability} + truncation telemetry \\
\emph{agent\_\allowbreak{}costs{[}{]}} (in/\allowbreak{}out/\allowbreak{}cache/\allowbreak{}reasoning tokens,\allowbreak{} elapsed\_\allowbreak{}s,\allowbreak{} cost\_\allowbreak{}usd) & efficiency; \textbf{Behavioral} (effort allocation per role) \\
\emph{observation\_\allowbreak{}hash}/\allowbreak{}\emph{action\_\allowbreak{}hash}/\allowbreak{}\emph{prev\_\allowbreak{}round\_\allowbreak{}hash}/\allowbreak{}\emph{signed} + \emph{IntegrityChecker} & integrity (§4.4) \\
\emph{benchmark\_\allowbreak{}metrics.benchmark\_\allowbreak{}comparisons{[}{]}} & benchmark-relative,\allowbreak{} ``beat /\allowbreak{}8'' \\
\emph{research\_\allowbreak{}brief},\allowbreak{} \emph{analyst\_\allowbreak{}brief},\allowbreak{} \emph{target\_\allowbreak{}allocation.rationale},\allowbreak{} \emph{risk\_\allowbreak{}officer\_\allowbreak{}notes},\allowbreak{} \emph{reflection\_\allowbreak{}notes} & \textbf{Behavioral} (stated-reasoning corpus,\allowbreak{} §6.5/\allowbreak{}§7.10) \\
\emph{lessons\_\allowbreak{}generated{[}{]}} (observation/\allowbreak{}implication/\allowbreak{}type/\allowbreak{}event\_\allowbreak{}types/\allowbreak{}regime),\allowbreak{} \emph{memory\_\allowbreak{}provenance},\allowbreak{} \emph{times\_\allowbreak{}referenced} & \textbf{Behavioral} (metacognition /\allowbreak{} learning trajectories) \\
\emph{mapc\_\allowbreak{}delta\_\allowbreak{}proposed} vs \emph{mapc\_\allowbreak{}delta\_\allowbreak{}projected},\allowbreak{} \emph{mapc\_\allowbreak{}adjustments\_\allowbreak{}clipped} & \textbf{Behavioral} (intent-vs-guardrails); Discipline \\
\emph{risk\_\allowbreak{}approved},\allowbreak{} \emph{accepted\_\allowbreak{}rejection},\allowbreak{} \emph{no\_\allowbreak{}trade\_\allowbreak{}accepted} & \textbf{Behavioral} (inter-agent /\allowbreak{} committee dynamics) \\
\emph{tool\_\allowbreak{}calls\_\allowbreak{}log{[}{]}} (tool,\allowbreak{} order,\allowbreak{} turns),\allowbreak{} \emph{candidate\_\allowbreak{}signals},\allowbreak{} \emph{universe\_\allowbreak{}changes} & \textbf{Behavioral} (information-foraging fingerprint) \\
\emph{decision\_\allowbreak{}attempts},\allowbreak{} \emph{parse\_\allowbreak{}failures},\allowbreak{} \emph{constraint\_\allowbreak{}violations} & \textbf{Behavioral} (struggle signatures); Reliability \\
\end{longtable}
}

\begin{center}\begin{minipage}{0.86\linewidth}\small
\textbf{Table D1.}\textit{ Audit field → metric map. Tooling: \emph{build\_paper\_tables.py} (body tables, D1--D5 scorecard, and held-out judge outputs), \emph{run\_ablation.py} (variant generation + launch), \emph{show\_audit.py --verify} (chain verification).}
\end{minipage}\end{center}

\subsection{Appendix E: Behavioral Anecdotes (round-id-pinned)}\label{appendix-e-behavioral-anecdotes-round-id-pinned}

Representative behaviors from the decision trail, each resolvable to a hash-verified round (the full bank
is available from the authors on request).

Module-detach collapses the decision process (the behavioral diagnostic in §7.7). In one round
(2025-11-03), both agents use the same anchor model. The \emph{full} agent weighs \textasciitilde128 signals, scouts \textasciitilde35
candidates, and commits a mean-variance-optimal book with explicit weights, an exclusion list, binding
constraints, and a cost-management lesson. The \emph{bare-workflow} agent weighs \textasciitilde56 signals and scouts none.
Its entire rationale is a one-line note accepting a default risk-parity book with no constraints, no
adjustments, and an empty reflection. Yet the caps keep the realized book safe, so returns are unaffected
(Table 8).

Two roads to a hold (autonomous mode). \emph{gemini-3.5-flash} (2026-06-16) holds a full mean-variance thesis
in its reflection, but its committed round records only a truncation marker --- it exhausted its tool-turn
budget exploring and never serialized a final allocation. \emph{minimax-m3} (2026-06-16) emits schema-invalid
output (an empty value where a list was required); the round is rejected and falls back to hold. The
week-2 record shows the two failures are independent and recoverable: gemini trades 06-23/06-24 while
minimax holds 06-24; gemini holds 06-25/06-26 while minimax trades 06-23/06-25/06-26 --- no synchrony
(§7.10).

Five patterns behind the low Coherence-judge scores (§7.9), verbatim from the cached judge rationales:
(i) direct contradiction --- ``analysis says TSLA should be \emph{reduced} but its weight change is +1.7\%'';
(ii) equal-weight collapse --- ``near-equal-weight 7.1--7.3\% across all names \ldots{} CAT +3.74 vs LIN −3.59
receive essentially identical weights, incoherent with a signal-driven mean-variance analysis'';
(iii) name mismatch --- the thesis names AVGO/GS/MS/NEM while the book holds PG/AMT/NEE/AMZN;
(iv) weights anti-correlated with the cited signals --- the most bearish name overweighted at 12\% while
the strongest bullish name gets 0.2\%; (v) no thesis at all --- an autonomous→MAPC placeholder with ``no
thesis for the specific name selection.''

Competent, signal-grounded decisions. \emph{claude-sonnet-4.6} (2026-06-15) reads a risk-on regime, flags a
negative momentum loading in its factor model, and rotates to correct it, exiting its worst-momentum and
highest-volatility names. \emph{deepseek-v4-pro} (2026-06-17) rotates from deteriorating mega-cap tech into
cyclicals and financials, citing a large-cap sentiment collapse against improving trajectories elsewhere.
\emph{gpt-5.4-mini} (2026-06-17) tilts to the strongest-momentum names, keeps an overweight where the news
trend improved, and trims negative-momentum or deteriorating-sentiment ones.

These illustrate the two hold mechanisms (§7.10) alongside the signal-grounded reasoning the scorecard
rewards (claude-sonnet-4.6 Coherence 0.60, APM-CS 83.6).

\subsection{Appendix F: Capability-Axis Robustness (construct validity)}\label{appendix-f-capability-axis-robustness-construct-validity}

This version hardens two axes (§6.4) and validates that the change does not rewrite the conclusions.
Discipline (D4) is now bounded, \(\tfrac12\text{self\_constrained}+\tfrac12(1-\text{clip\_dependence})\).
The earlier draft scored only the pre-MAPC self-constraint rate plus a turnover-cap leg that was a no-op
at campaign turnover (\textasciitilde0.15). Composure (D3) is now the mean of its legs rather than their product.
The product let one weak leg zero the axis and, because the consistency leg comes from the memory
module, made Composure \emph{uncomputable} under the memory ablation. The mean falls back to the
memory-free leg. All numbers come
from \emph{scripts/analyze\_axis\_robustness.py} over the per-run scorecard.

F.1 --- Rank stability (v1 → v2, layered). On the current five-seed judged scorecard, we isolate each
revision and compare config-mean APM-CS orderings. The D3/D4 revision alone gives Kendall τ = 0.86 /
Spearman ρ = 0.94 (its original v0.8.1 measurement on the three-seed cohort was τ = 0.78 / ρ = 0.90,
p \textless{} 0.001). The v1.1 Acuity noise-set revision alone gives τ = 0.75 / ρ = 0.88 (F.1b).
Compounding all three axis revisions at once gives τ = 0.58 / ρ = 0.78. The main claims are unchanged under every
variant --- deepseek·structured leads APM-CS and qwen3·structured is the Reliability artifact under \emph{both}
constructions. The movement is where the fixes aim: in the configurations that v1 Discipline over-credited (a perfect self-constraint rate without the clip-dependence penalty), and in the mid-pack, which D2 re-ranks once attention to the horizon-mismatched RSI distractor is penalized. The D3/D4 revision overturned exactly
one prior sub-finding --- that removing memory ``breaks'' Composure --- an artifact of the old uncomputable
construction; under v2, memory removal merely drops the consistency leg (§7.7).

F.1b --- Acuity noise-set revision (v1.1). Widening the D2 noise leg from realized-vol alone to the
two-distractor composite mean(z(realized\_vol), z(RSI\(_{14}\))) (§6.4) re-orders the D2 axis moderately
(config-mean D2 ordering: Kendall τ = 0.56, Spearman ρ = 0.78 across the ten configurations).
It nevertheless leaves the APM-CS composite ordering essentially intact (τ = 0.75, ρ = 0.88 on the judged five-seed
scorecard; D1 held fixed between variants, so the movement isolates the noise-set change). The single
substantive re-read is haiku·structured, whose Acuity rises from ≈0.2 to 0.44 --- its proposals avoid
the RSI distractor that the single-leg test did not probe (measured on the single ≥3-tilt round its
sparse-proposal style yields, §7.3-F4). The single-leg variant is exported as
\emph{D2\_acuity\_v1} alongside the headline axis, mirroring the D3/D4 v1↔v2 treatment.

F.2 --- Orthogonality (non-redundancy). Across the ten off-diagonal axis pairs the mean \textbar Pearson r\textbar{} is
0.23 (max 0.55, Coherence↔Discipline; Figure F1), so the five axes are near-orthogonal. The strongest pair (Coherence vs.~Discipline, r ≈ −0.55)
is interpretable: configurations that self-constrain hardest are not the ones whose allocations best
follow their stated reasoning.

\begin{center}
\includegraphics[width=0.52\linewidth,height=\textheight,keepaspectratio]{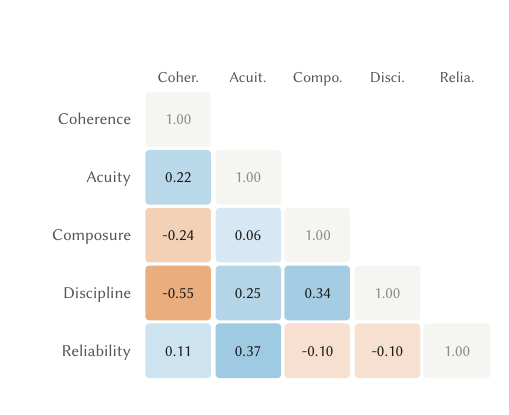}

\vspace{5pt}
\begin{minipage}{0.86\textwidth}\small \textbf{Figure F1.}\textit{ Inter-axis Pearson correlation across the campaign cohort (D1--D5). Off-diagonal magnitudes are
small (mean \textbar r\textbar{} = 0.23), evidence the five axes are non-redundant.}\end{minipage}
\end{center}

F.3 --- Weight sensitivity. Re-aggregating APM-CS under leave-one-axis-out (five variants) and under
reweighting the two Coherence components (agreement vs.~judge from 0.3/0.7 to 0.7/0.3) leaves the configuration
ordering stable: Kendall τ ≥ 0.71 against the equal-weight baseline in every case (the Coherence-reweighting
variants stay at τ ≥ 0.89).

F.4 --- Cross-mode sensitivity on the mode-comparable axes. Because Acuity (D2) and Discipline (D4)
are mode-confounded (§6.4), the cross-mode APM-CS contrasts in the body are descriptive profile
summaries. As a sensitivity, Table F1 restricts the comparison to the three axes measured identically
in both modes --- Coherence (D1), Composure (D3), and Reliability (D5). It averages their raw values
(all bounded to {[}0,1{]}; equal weights; computed directly from the published per-axis values in
Tables 5 and 10). This is a robustness read, not an official composite. On it, the backtest picture
is model-dependent rather than uniformly pro-structure: structured is favored only for deepseek
(+0.14), haiku and gpt-5-mini are within ±0.02, and gemini and qwen3 favor autonomy (qwen3's
structured leg is the five-round showcase seed). The live cohort favors autonomy for all six pairs
(margins +0.02 to +0.20). The contrast with the full composite --- which favors the four reliable
structured backtest configurations --- indicates that the composite's structured advantage on the
backtest rides substantially on the mode-confounded axes, consistent with the scoped reading in
§7.3-F2 and §7.6.

\begin{center}\begin{minipage}{0.86\linewidth}\small
\textbf{Table F1.}\textit{ Cross-mode sensitivity on the mode-comparable axes: per-configuration means of the raw
D1/D3/D5 values (from Tables 5 and 10) and their three-axis average; Δ = structured − autonomous.
✦ qwen3-235b·structured is the five-round showcase seed (§7.3).}
\end{minipage}\end{center}

{\def\LTcaptype{none} 
\begin{longtable}[]{@{}
  >{\raggedright\arraybackslash}p{(\linewidth - 10\tabcolsep) * \real{0.3488}}
  >{\raggedright\arraybackslash}p{(\linewidth - 10\tabcolsep) * \real{0.1395}}
  >{\raggedright\arraybackslash}p{(\linewidth - 10\tabcolsep) * \real{0.1395}}
  >{\raggedright\arraybackslash}p{(\linewidth - 10\tabcolsep) * \real{0.1395}}
  >{\raggedright\arraybackslash}p{(\linewidth - 10\tabcolsep) * \real{0.1395}}
  >{\raggedright\arraybackslash}p{(\linewidth - 10\tabcolsep) * \real{0.0930}}@{}}
\toprule\noalign{}
\begin{minipage}[b]{\linewidth}\raggedright
Cohort · model
\end{minipage} & \begin{minipage}[b]{\linewidth}\raggedright
D1 s/\allowbreak{}a
\end{minipage} & \begin{minipage}[b]{\linewidth}\raggedright
D3 s/\allowbreak{}a
\end{minipage} & \begin{minipage}[b]{\linewidth}\raggedright
D5 s/\allowbreak{}a
\end{minipage} & \begin{minipage}[b]{\linewidth}\raggedright
mean s/\allowbreak{}a
\end{minipage} & \begin{minipage}[b]{\linewidth}\raggedright
Δ (s−a)
\end{minipage} \\
\midrule\noalign{}
\endhead
\bottomrule\noalign{}
\endlastfoot
BT · deepseek-v3.1 & 0.45 /\allowbreak{} 0.42 & 0.93 /\allowbreak{} 0.90 & 0.99 /\allowbreak{} 0.62 & 0.79 /\allowbreak{} 0.65 & +0.14 \\
BT · claude-haiku-4.5 & 0.23 /\allowbreak{} 0.44 & 0.95 /\allowbreak{} 0.81 & 1.00 /\allowbreak{} 0.98 & 0.73 /\allowbreak{} 0.74 & −0.02 \\
BT · gpt-5-mini & 0.43 /\allowbreak{} 0.47 & 0.92 /\allowbreak{} 0.92 & 1.00 /\allowbreak{} 0.99 & 0.78 /\allowbreak{} 0.79 & −0.01 \\
BT · gemini-2.5-flash-lite & 0.38 /\allowbreak{} 0.47 & 0.93 /\allowbreak{} 0.77 & 0.61 /\allowbreak{} 0.95 & 0.64 /\allowbreak{} 0.73 & −0.09 \\
BT · qwen3-235b ✦ & 0.33 /\allowbreak{} 0.42 & 0.95 /\allowbreak{} 0.91 & 0.19 /\allowbreak{} 0.99 & 0.49 /\allowbreak{} 0.77 & −0.28 \\
Live · claude-sonnet-4.6 & 0.52 /\allowbreak{} 0.60 & 0.93 /\allowbreak{} 0.94 & 0.82 /\allowbreak{} 1.00 & 0.76 /\allowbreak{} 0.85 & −0.09 \\
Live · deepseek-v4-pro & 0.50 /\allowbreak{} 0.62 & 0.95 /\allowbreak{} 0.95 & 1.00 /\allowbreak{} 1.00 & 0.82 /\allowbreak{} 0.86 & −0.04 \\
Live · qwen3.7-plus & 0.41 /\allowbreak{} 0.59 & 0.95 /\allowbreak{} 0.94 & 0.58 /\allowbreak{} 1.00 & 0.65 /\allowbreak{} 0.84 & −0.20 \\
Live · gpt-5.4-mini & 0.48 /\allowbreak{} 0.56 & 0.49 /\allowbreak{} 0.69 & 0.97 /\allowbreak{} 0.99 & 0.65 /\allowbreak{} 0.75 & −0.10 \\
Live · gemini-3.5-flash & 0.42 /\allowbreak{} 0.51 & 0.46 /\allowbreak{} 0.44 & 0.94 /\allowbreak{} 1.00 & 0.61 /\allowbreak{} 0.65 & −0.04 \\
Live · minimax-m3 & 0.41 /\allowbreak{} 0.56 & 0.47 /\allowbreak{} 0.45 & 0.56 /\allowbreak{} 0.49 & 0.48 /\allowbreak{} 0.50 & −0.02 \\
\end{longtable}
}

F.5 --- Human anchoring of the Coherence judge (blinded). To anchor the judge component against
human judgment, one author blind-rated 50 sampled rounds (26 backtest, 24 live; seed-fixed
stratified sampling) on the judges' own rubric and inputs, with model identities and judge scores
withheld during rating. Agreement with the per-track held-out judges is Spearman ρ = 0.60
(p \textless{} 1e-5; 74\% within one rubric point; backtest 0.63, live 0.56) --- comparable to the two judges'
agreement with each other on their 633 shared rounds (ρ = 0.65, §6.4.1). As a convergence
check, two frontier models outside both cohorts and both judge roles (Claude Fable 5 and
GPT-5.6-Sol; three ratings per round, per-round means) rated the same blinded sheet.
Their correlations are ρ = 0.79 and 0.76 against the held-out judges and 0.79 and 0.71 against
the human rater, respectively. Their agreement with each other is ρ = 0.88. Rating \emph{levels} differ systematically --- the held-out judges are the strictest (sample mean
2.72 of 5) and the five raters span roughly one rubric point --- while \emph{orderings} concord across
all ten rater pairs. This is exactly why the two-leg coherence divergence is treated as
descriptive rather than a calibrated effect size (§7.9): the anchoring validates the judge's
ordering, not any particular level. Two scope notes. First, the human rater is an author (blinding
removes the model-favoritism channel, but an independent expert panel remains future work, §8.6). Second, the auxiliary frontier raters serve this rater-agreement analysis only --- they never score models, and one shares a vendor family with a live-cohort model, which the held-out protocol of
§6.4.1 excludes from any judging role.

\subsection{Appendix G: Reproducibility and Artifacts}\label{appendix-g-reproducibility-and-artifacts}

Reproducibility. Every number in this paper recomputes from the persisted per-run audit stores: the
scorecard builder derives the campaign and live scorecards (Tables 5 and 10) and the D1--D5 axes from
the DecisionRound records (field map in Appendix D), the ablation tooling derives the ablation tables
(Tables 6--8) from the same stores, the behavioral analyzer derives the hold-mechanism split and the
anecdote bank, and the figure scripts regenerate all data figures from those outputs --- no reported
number is hand-maintained (Table 9's per-module verdicts synthesize that evidence). Judge scores are cached by \emph{round\_id} per judge model (minimax-m3 for backtest,
GLM-5.2 for live), so recomputation reuses the exact adjudications reported here rather than
re-sampling them. Campaign configuration is declarative (TOML): the backtest ran 26 bi-weekly rounds
per seed (five seeds per structurally-reliable configuration, §7.2) and the live track ran the
four-week daily campaign of §7.9; model identities and snapshots are pinned in Table C1 (Appendix C; all accessed
via OpenRouter), reasoning budgets and output caps are stated in §7.2/§7.9, and per-round LLM token
and dollar costs persist in the audit trail (\emph{agent\_costs}). Data providers are named in Table 2;
all access is point-in-time-gated (§4.1) with the gathered snapshots persisted per round. Hash-chain
verification (§4.4) re-derives every round's hashes, so a third party holding the stores can confirm
the recorded decisions are the ones the engine produced. The audit databases, scorecard and analysis
tooling, and figure-generation scripts are available from the corresponding author on reasonable request.

\end{document}